\documentclass[conference]{IEEEtran}
\usepackage{times}

\usepackage[numbers]{natbib}
\usepackage{amsmath,amssymb}
\usepackage{bm}
\usepackage{colortbl}
\usepackage{multicol}
\usepackage{listings}
\usepackage{xcolor}
\usepackage{graphicx}
\usepackage{subcaption}
\usepackage{svg}
\usepackage{booktabs}
\usepackage{makecell}

\usepackage[bookmarks=true]{hyperref}
\usepackage{xr-hyper}
\usepackage{bbding}

\usepackage{multirow}
\usepackage{tabularx}
\usepackage{arydshln}
\usepackage{ragged2e}
\usepackage{makecell}
\usepackage{xspace}
\usepackage{array}
\usepackage{colortbl}
\usepackage{float}
\usepackage{algorithm}
\usepackage{algorithmicx}
\usepackage{algpseudocode}
\usepackage{bm}
\usepackage[T1]{fontenc}
\usepackage[scaled=1.0]{inconsolata} 

\definecolor{kwpurple}{RGB}{140,0,200}
\definecolor{cmteal}{RGB}{0,140,140}

\definecolor{lightgray}{rgb}{0.95, 0.95, 0.95}

\newcolumntype{Y}{>{\RaggedRight\arraybackslash}p{3.8cm}}
\newcolumntype{V}{!{\color{black!45}\vrule width 0.4pt}}

\hypersetup{
  pdftitle={GuidedVLA: Specifying Task-Relevant Factors via Plug-and-Play Action Attention Specialization},
  pdfauthor={Xiaosong Jia, Bowen Yang, Zuhao Ge, Xian Nie, Yuchen Zhou, Cunxin Fan, et al.},
  pdfsubject={Vision-Language-Action Models},
  pdfkeywords={Vision-Language-Action, Robot Learning, Attention Specialization, Factor Guidance}
}

\newcommand{\alias}{GuidedVLA\xspace}
\newcommand{\mainref}[1]{\ref{#1}}
\newcommand{\appref}[1]{\ref{#1}}

\begin{document}

\makeatletter
\let\@oldmaketitle\@maketitle
\renewcommand{\@maketitle}{\@oldmaketitle
  \begin{center}
  \captionsetup{type=figure}
  \setcounter{figure}{0}
  \includegraphics[trim=0.4ex 0 0 0, clip, width=1.0\textwidth]{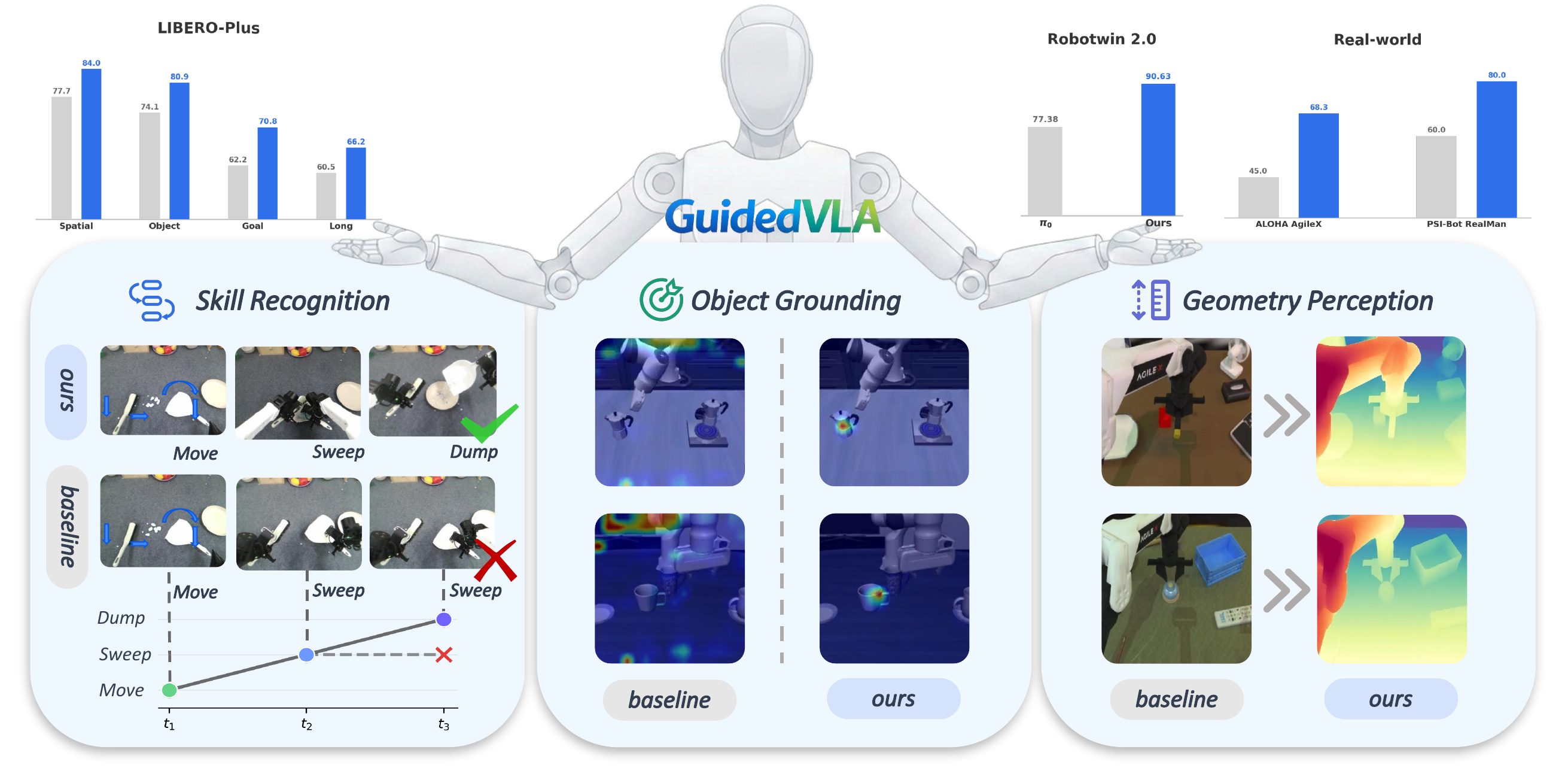}
    \caption{We present \textbf{\alias}, a VLA paradigm in which the action decoder is explicitly guided to capture task-relevant information such as object grounding, spatial geometry, and temporal skill logic. Across simulation and real-robot experiments, GuidedVLA significantly improves success rates in both in-domain and out-of-domain settings, demonstrating the effectiveness of specifying action-decoder attention heads with explicit guidance.}
    \label{fig:teaser}
    \vspace{-7mm}
  \end{center}
}
\makeatother

\title{\alias: Specifying Task-Relevant Factors via Plug-and-Play Action Attention Specialization}

\author{Xiaosong Jia$^{*\dagger,1,2}$, Bowen Yang$^{*,3}$, Zuhao Ge$^{*,1,2}$, Xian Nie$^{*,3}$, Yuchen Zhou$^{*,1,2}$, Cunxin Fan$^{*\dagger,3}$, \\ Yufeng Li$^{3}$, Yilin Chai$^{3}$,
Chao Jing$^{1,2}$, Zijian Liang$^{3}$, Qingwen Bu$^{4}$, \\ Haidong Cao$^{1,2}$, Chao Wu$^{1,2}$, Qifeng Li$^{3}$, Zhenjie Yang$^{3}$, Chenhe Zhang$^{1,2}$,\\
Hongyang Li$^{4}$, Zuxuan Wu\textsuperscript{\Envelope}$^{1,2}$, Junchi Yan\textsuperscript{\Envelope}$^{3}$, Yu-Gang Jiang\textsuperscript{\Envelope}$^{1,2}$ \\
$^{1}$Institute of Trustworthy Embodied AI (TEAI), Fudan University \\
$^{2}$Shanghai Key Laboratory of Multimodal Embodied AI\\
$^{3}$ Shanghai Jiao Tong University \\
$^{4}$ OpenDriveLab, The University of Hong Kong \\

* Core Contributors \quad\quad $^{\dagger}$ Project Lead \quad\quad \textsuperscript{\Envelope} Correspondence Authors \\
\normalsize{
\color{magenta}\url{https://guidedvla.github.io/project_page/}
    }

}

\maketitle

\begin{abstract}

Vision-Language-Action (VLA) models aim for general robot learning by aligning action as a modality within powerful Vision-Language Models (VLMs). Existing VLAs rely on end-to-end supervision to implicitly enable the action decoding process to learn task-relevant features. However, without explicit guidance, these models often overfit to spurious correlations, such as visual shortcuts or environmental noise, limiting their generalization.
In this paper, we introduce \alias, a framework designed to manually guide the action generation to focus on task-relevant factors. Our core insight is to treat the action decoder not as a monolithic learner, but as an assembly of functional components. Individual attention heads are supervised by manually defined auxiliary signals to capture distinct factors.
As an initial study, we instantiate this paradigm with three specialized heads: object grounding, spatial geometry, and temporal skill logic. Across simulation and real-robot experiments, \alias improves success rates in both in-domain and out-of-domain settings compared to strong VLA baselines. Finally, we show that the quality of these specialized factors correlates positively with task performance and that our mechanism yields decoupled, high-quality features. Our results suggest that explicitly guiding action-decoder learning is a promising direction for building more robust and general VLA models.
\end{abstract}

\IEEEpeerreviewmaketitle

\section{Introduction}

Vision-Language-Action (VLA) models~\cite{zitkovich2023rt2,kim2025openvla,black2024pi_0} represent a significant step toward generalist robot policies by integrating action as a specialized modality within the rich feature space of Vision-Language Models (VLMs). By leveraging the massive pre-training of VLMs~\cite{team2025gemini,touvron2023llama,beyer2024paligemma,li2023blip}, these models can inherit high-level semantic knowledge and reasoning capabilities essential for complex tasks. However, current VLA training typically relies on end-to-end supervision where the action decoder is expected to implicitly learn task-relevant factors from demonstration data~\cite{brohan2022rt1}. According to pioneering studies in the field of computer vision and imitation learning, end-to-end learning without explicit guidance may lead to shortcut learning~\cite{geirhos2020shortcut, geirhos2018imagenet} or causal confusion~\cite{de2019causal}.

In practice, we observe that the action decoder of VLA often latches onto spurious correlations, such as background textures or incidental camera artifacts, as shown in Fig.~\ref{fig:teaser}. While some cross-attention heads in the action decoder occasionally attend to relevant regions, this behavior is highly stochastic and varies across different heads and scenarios. This randomness suggests that although VLM backbones provide a robust feature stream, \textbf{the action decoder does not learn a stable, causal understanding of the task, but instead relies on a shifting set of features} and thus struggles to generalize.

Because end-to-end supervision alone makes the decision-making process of VLA opaque and inconsistent, we propose \textbf{\alias}, a framework that transforms the action decoder from a monolithic learner into an assembly of functionally specialized components. Instead of allowing the cross-attention heads to develop roles implicitly, we manually specify the information each head should capture by supervising them with distinct, task-relevant auxiliary signals.

While this paradigm is designed to be general and extensible, in this work, we instantiate it by supervising three primary factors, as in Fig.~\ref{fig:teaser}: (i) \textbf{object grounding}, ensuring action tokens attend to task-relevant regions; (ii) \textbf{skill recognition}, enabling action tokens to identify the intended sub-skill or phase of a multi-step behavior; and (iii) \textbf{geometry perception}, allowing action tokens to leverage 3D spatial information. Our probing experiments reveal that current VLAs are brittle across all three factors, and that the proposed guidance mechanism effectively resolves these deficiencies.

Across multiple simulation benchmarks and real-world experiments, \alias achieves a significant performance boost for $\pi_0$~\cite{black2024pi_0}, surpassing other recent feature-training methods in the field~\cite{zhang2025dreamvla,shen2025expertise}. Furthermore, we provide a quantitative evaluation showing a strong positive correlation between factor understanding and overall success rates. Finally, we validate that partitioning factors into specialized attention heads produces better-decoupled features than a mixture approach where all heads are jointly supervised.

In summary, we make the following contributions:
\begin{itemize}
\item We propose \alias, a general paradigm for VLA that mitigates overfit risk by specifying task-relevant factors through functional attention specialization.
\item We instantiate this framework by designing three specialized heads: object grounding, skill recognition, and geometry perception and demonstrate through probing that such explicit guidance resolves the inherent brittleness and stochasticity of unguided VLA decoders.
\item We provide extensive evaluations across multiple simulation benchmarks and real-robot tasks, showing that \alias significantly improves state-of-the-art baselines in both in-domain and out-of-distribution scenarios.
\item We offer quantitative insights into head specialization, validating that our approach yields high-quality features that correlate positively with task performance.
\end{itemize}
\vspace{-1mm}

\begin{figure*}[!htbp]
    \centering
    \includegraphics[width=1\linewidth]{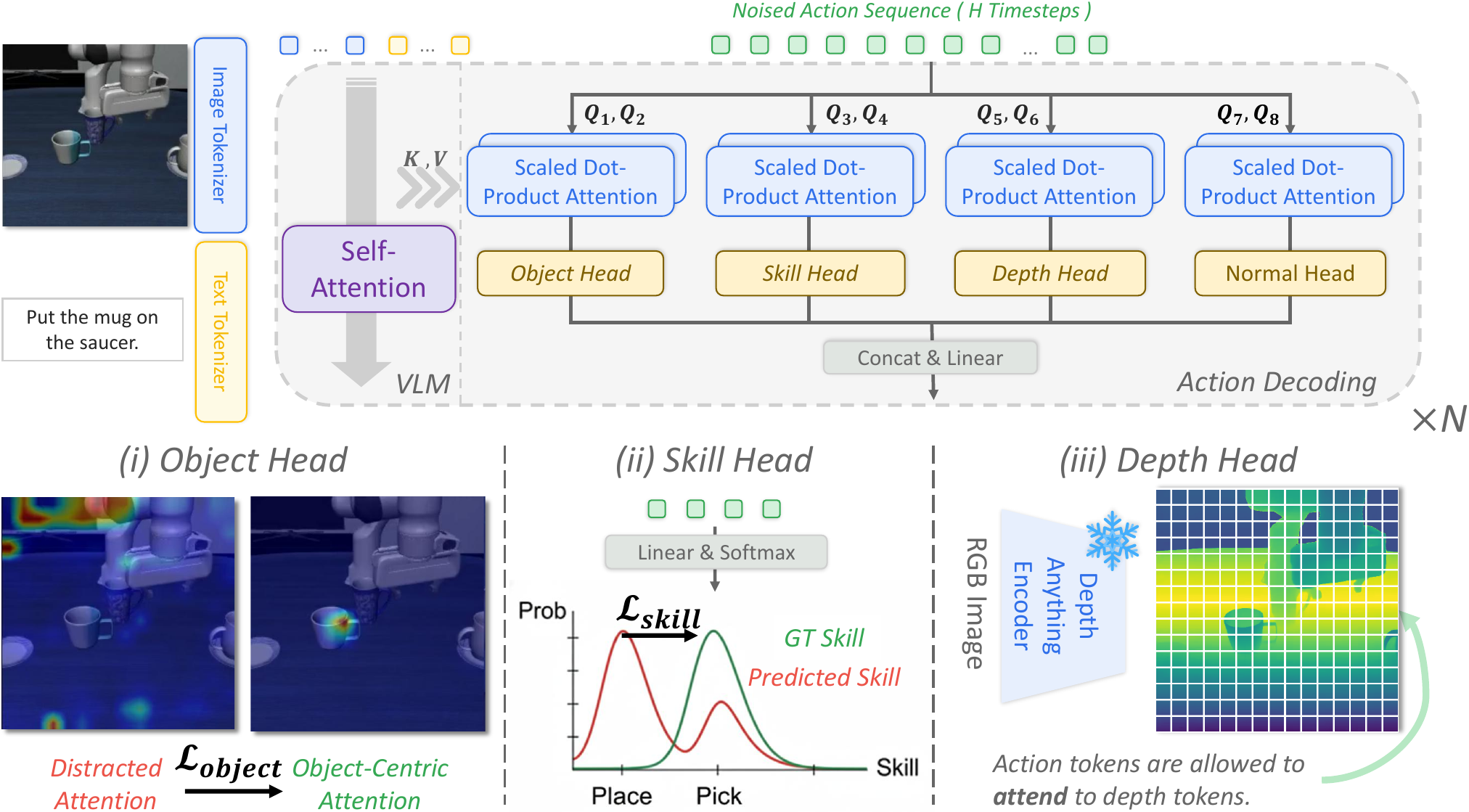}
    \caption{\textbf{Architecture of \alias.} We introduce explicit, structured guidance into the multi-head attention layers of the VLA action decoder. Instead of relying on implicitly entangled representations, we repurpose dedicated attention heads to specialize in distinct task-relevant factors:
\textbf{(i) Object Head} supervises  its attention maps to explicitly ground task-relevant objects and suppress distractors via $\mathcal{L}_{object}$;
\textbf{(ii) Skill Head} aligns internal feature representations with temporal skill phases (e.g., Pick $\rightarrow$ Place) through auxiliary classification $\mathcal{L}_{skill}$;
\textbf{(iii) Depth Head} injects geometric cues via  cross attention only to features from a depth encoder.
These guidance forces the policy to explicitly aware spatial, temporal, and geometric structures.}
\label{fig:method}
\end{figure*}

\section{Related Work}

\noindent\textbf{Vision-Language-Action (VLA) Models}: Vision-Language-Action (VLA) models aim to map visual observations and language instructions to low-level robot actions by combining pretrained vision-language models with large-scale robot demonstrations~\cite{ahn2022saycan,brohan2022rt1,zitkovich2023rt2,driess2023palme,ghosh2024octo,kim2025openvla,bjorck2025gr00t,zheng2024tracevla,yang2025drivemoe}.
One important research direction focuses on scaling embodied data through multi-source datasets \cite{o2024openx,walke2023bridgedata,khazatsky2024droid,dasari2019robonet,mandlekar2018roboturk,jiang2025galaxea, h2r}, standardized multi-task benchmarks \cite{mees2022calvin,liu2023libero,fei2025libero,james2020rlbench,yu2020meta}, and evaluations under distribution shift \cite{fei2025libero,nair2022r3m,ebert2023bridgedata}. Another line of work improves training and inference recipes, including multimodal prompting \cite{jiang2022vima,fan2025interleave}, parameter-efficient adaptation \cite{kim2025openvlaoft,wen2025dexvla,clare,hu2022lora}, and inference-time acceleration~\cite{wen2025tinyvla,black2025real,asyncvla,ma2025running,zhao2025vla}. In parallel, prior work strengthens the action pathway through alternative action parameterizations and learning objectives, including diffusion- or flow-based generation \cite{chi2025diffusion,black2024pi_0,black2025pi_05,liu2025rdt1b,cheng2025moe,liang2025discrete,chen2025unified,wen2025dvla}, action chunking for temporal abstraction~\cite{zhao2023act}, and discrete or compressed action tokenizers to better match control bandwidth~\cite{pertsch2025fast,wang2025vqvla}.

\noindent\textbf{Auxiliary Tasks for Robotics Models}
Structured intermediate representations improve policy robustness under distribution shift. Object-centric methods factor manipulation around task-relevant entities such as object poses, keypoints, and relations \cite{huang2025rekep,haldar2025point,lin2025cpgen,hsu2025spot,wu2025afforddp, posavla,coavla,li2025language,pan2025omnimanip,chapin2026storm}.
Skill-based representations support long-horizon reasoning by decomposing tasks into reusable subgoals \cite{ahn2022saycan,liang2024skilldiffuser,ma2024hdp,skill-aware,skill-video,mosvla,ahn2022can}.
Geometry-aware policies that operate on 3D representations, including point clouds and 3D scene tokens, achieve strong viewpoint and instance generalization \cite{ze2024dp3,wang2024diffuseractor,lin2024manicm,zhen20243dvla,qu2025spatialvla,aimbot, point-vla,geopredict,stereovla,peafowl, geovla, depthvla,spatialforcing,li2025bridgevla,lee2025tracegen,huang2026pointworld,ni2025vo,zhi20253dflowaction,bhat20253d}.
Our work complements these approaches by supervising object-centric structure, skills, and geometry into separable internal pathways within the action decoder of VLA.

\section{Method}

\subsection{Problem Setup and Motivation}

\begin{algorithm}[htbp]
\small
\caption{Decoupled Attention with Guided Heads}
\label{alg:pg_heads_control_losses}
\begingroup
\newcommand{\HeadComment}[1]{\hfill\makebox[0.24\linewidth][l]{$\triangleright$~#1}}
\begin{algorithmic}[1]
\Require Action hidden states $X_L$ at layer $L$
\Require Head sets: $\mathcal{H}_{o}$ (object), $\mathcal{H}_{s}$ (skill), $\mathcal{H}_{d}$ (depth)
\Require Joint cache $(K, V)$; Depth cache $(K^d, V^d)$
\Ensure Fused attention output $A_L$
\State $Q \leftarrow \mathrm{Proj_Q}(X_L)$
\Statex
\Statex Stage 1: Factor-Specific Attention
\State $P_o, A_o \leftarrow \mathrm{Attn}(Q[\mathcal{H}_{o}], K, V)$ \HeadComment{Object Head}
\State $A_s \leftarrow \mathrm{Attn}(Q[\mathcal{H}_{s}], K, V)$ \HeadComment{Skill Head}
\State $A_d \leftarrow \mathrm{Attn}(Q[\mathcal{H}_{d}], K^d, V^d)$ \HeadComment{Depth Head}
\Statex
\Statex Stage 2: Per-Head Supervision
\State Apply $\mathcal{L}_{object}$ (Eq.~\ref{eq:grounding_loss}) to $P_o$
\State Apply $\mathcal{L}_{skill}$ (Eq.~\ref{eq:skill_loss}) to $A_s$
\Statex
\Statex Stage 3: ControlNet-style Residual Fusion
\State $A^{\text{specified}}_L \leftarrow \mathrm{Proj_O}\bigl(\mathrm{Concat}(A[:])\bigr)$
\State $A_L \leftarrow \mathrm{ZeroConv}(A^{\text{specified}}_L) + A_L^{\text{main}}$ \Comment{Merge with Main Branch}
\end{algorithmic}
\endgroup
\end{algorithm}

Recent representative Vision-Language-Action (VLA) models~\cite{black2024pi_0,kim2025openvla,zitkovich2023rt2} extend Vision-Language Models (VLMs) by introducing action tokens $\bm{a}$ alongside vision tokens $\bm{v}$ and language tokens $\bm{l}$.
The action generation process is trained to regress robot trajectories by denoising $\bm{a}$ conditioned on $(\bm{v}, \bm{l})$.

Although VLMs provide rich semantic features in $(\bm{v}, \bm{l})$, the action tokens $\bm{a}$ do not inherently learn to extract task-critical information.
As in Figure~\ref{fig:teaser}, action token attention often diffuses over irrelevant background regions. This motivates our central designs: \textbf{\emph{How can we guide action decoding to extract task-relevant information?}}

\vspace{-0.5em}
\subsection{What to Guide: Three Task-Relevant Factors}
\label{sec:factors}

We identify three factors correlated with robotics tasks, based on our preliminary probing experiments in Sec.~\ref{sec:factor_quality_analysis}:

\begin{enumerate}
  \item \textbf{Object Grounding}: whether action tokens can attend to the correct task-relevant regions (e.g., affordance).
  \item \textbf{Skill Recognition}: whether action tokens correctly identify the current sub-skill or temporal phase within a complex robotics task (e.g., long-horizon).
  \item \textbf{Geometry Perception}: whether action tokens can utilize 3D spatial information when performing fine-grained tasks (e.g., click bell).
\end{enumerate}

These factors are complementary: grounding localizes the target, skill recognition defines the behavior, and geometry provides the spatial constraints for execution.
Together, they constitute a comprehensive semantic interface between high-level VLM representations and low-level control.

\subsection{How to Guide: Attention Head Specialization}
\label{sec:head_specialization}

To capture decoupled task-relevant information, the Multi-Head Attention (MHA)~\cite{vaswani2017attention} adopted in the action decoder offers a natural solution with minimal structure modification: we explicitly assign specific heads to capture certain factors by \textbf{applying different supervision signals on different heads}.

\begin{figure}[t]
    \centering
    \includegraphics[width=\columnwidth,height=0.25\textheight,keepaspectratio]{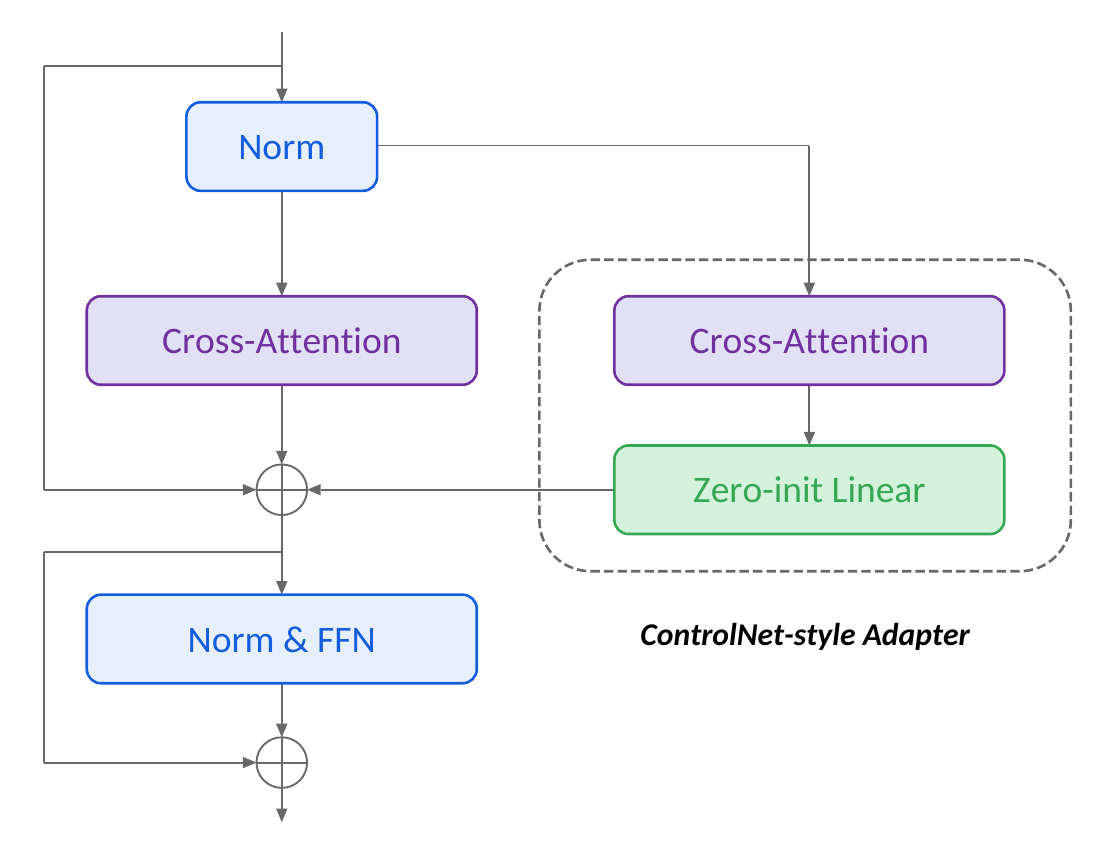}
    \vspace{-1em}
    \caption{\textbf{ControlNet-style residual adapter for plug-and-play head specialization.} The pretrained main attention branch is kept as the behavior-preserving path, while a factor-specific attention branch is fused through a zero-initialized projection. The adapter copies weights from the base policy and gradually injects task-relevant biases during training.}
    \label{fig:controlnet_style_adapter}
    \vspace{-2em}
\end{figure}

Because large-scale pretrained backbones already exist~\cite{kim2025openvla,black2024pi_0}, we can equip them with specified heads while preserving pretrained capabilities, thanks to the natural decoupling characteristics of multi-head attention. Specifically, for those pretrained backbones, we propose a \textbf{ControlNet-style}~\cite{zhang2023adding} residual adapter strategy, as illustrated in Fig.~\ref{fig:controlnet_style_adapter}. To add a supervised head $\mathrm{Attn}_{\text{specified}}$, we introduce a zero-initialized projection $\mathrm{ZeroConv}$ before fusing with the main branch attention features $\mathrm{Attn}_{\text{main}}$:
\begin{equation}
\setlength{\abovedisplayskip}{0pt}
\setlength{\belowdisplayskip}{0pt}
\setlength{\abovedisplayshortskip}{0pt}
\setlength{\belowdisplayshortskip}{0pt}
    \mathrm{Attn}(\mathbf{x}) = \mathrm{Attn}_{\text{main}}(\mathbf{x}) + \mathrm{ZeroConv}\left(\mathrm{Attn}_{\text{specified}}(\mathbf{x})\right).
\end{equation}
Since $\mathrm{ZeroConv}$ is initialized to zero, the control branch initially contributes no signal. This ensures the model retains its pre-trained behavior at the start of training, while gradually learning to inject factor-specific biases as optimization proceeds.

\begin{table*}[htbp]
\centering
\caption{\textbf{LIBERO-Plus Benchmark Results.} The proposed model achieves the highest average success rate, with a significant boost compared to its base model $\pi_0$. Notably, \textbf{single-head ablations reveal task-specific alignment}: the object head is strongest among single-head variants on the \textit{Object} and \textit{Long} suites, the skill head gives the best single-head result on the \textit{Goal} suite, and the depth head performs best on the \textit{Spatial} suite.}
\vspace{-2mm}
\label{tab:libero_plus_leaderboard}
\small
\setlength{\tabcolsep}{5pt}
\setlength{\aboverulesep}{0pt}
\setlength{\belowrulesep}{0pt}
\renewcommand{\arraystretch}{1.25}
\scalebox{0.93}{
\begin{tabular}{lVcccccccVccccVc}
\toprule

\multicolumn{1}{lV}{}
& \multicolumn{7}{cV}{\textbf{Perturbation Dimensions}}
& \multicolumn{4}{cV}{\textbf{Task Suites}}
& \\

\cmidrule(lr){2-8}\cmidrule(lr){9-12}

\textbf{Model}
& \textbf{Camera}
& \textbf{Robot}
& \textbf{Language}
& \textbf{Light}
& \textbf{Background}
& \textbf{Noise}
& \textbf{Layout}
& \textbf{Spatial}
& \textbf{Object}
& \textbf{Goal}
& \textbf{Long}
& \textbf{Total} \\
\midrule

OpenVLA~\cite{kim2025openvla}                 & 0.8  & 3.5  & 23.0 & 8.1  & 34.8 & 15.2 & 28.5 & 19.4 & 14.0 & 15.1 & 14.3 & 15.6 \\
OpenVLA-OFT~\cite{kim2025openvlaoft}         & 56.4 & 31.9 & 79.5 & 88.7 & 93.3 & 75.8 & 74.2 & 84.0 & 66.5 & 63.0 & 66.4 & 69.6 \\
NORA~\cite{hung2025nora}                      & 2.2  & 37.0 & 65.1 & 45.7 & 58.6 & 12.8 & 62.1 & 47.6 & 34.4 & 38.8 & 36.3 & 39.0 \\
WorldVLA~\cite{worldvla2025}                  & 0.1  & 27.9 & 41.6 & 43.7 & 17.1 & 10.9 & 38.0 & 32.5 & 28.6 & 31.8 & 8.2  & 25.0 \\
UniVLA~\cite{univla2025}                      & 1.8  & 46.2 & 69.6 & 69.0 & 81.0 & 21.2 & 31.9 & 55.5 & 36.7 & 40.7 & 39.9 & 43.9 \\
$\pi_0$-Fast~\cite{pertsch2025fast}           & 65.1 & 21.6 & 61.0 & 73.2 & 73.2 & 74.4 & 68.8 & 74.4 & 72.7 & 57.5 & 43.4 & 61.6 \\
RIPT-VLA~\cite{tan2025interactive}            & 55.2 & 31.2 & 77.6 & 88.4 & 91.6 & 73.5 & 74.2 & 85.8 & 64.3 & 58.0 & 67.5 & 68.4 \\
DreamVLA~\cite{zhang2025dreamvla}             & 65.0 & 40.9 & 63.5 & 85.7 & 82.7 & 85.0 & 74.0 & 79.7 & 79.0 & 61.7 & 59.8 & 69.9 \\
AdaMoE~\cite{shen2025expertise}               & 53.8 & 17.5 & 20.6 & 73.7 & 73.8 & 58.6 & 65.8 & 51.0 & 57.9 & 53.3 & 38.1 & 50.1 \\
Spatial Forcing~\cite{spatialforcing}         & 20.1 & 13.4 & 40.9 & 29.1 & 33.4 & 25.7 & 39.3 & 52.9 & 31.0 & 28.2 & 5.4  & 29.1 \\
VLA-Adapter~\cite{wang2026vlaadapter}         & 36.2 & 37.9 & 74.6 & 70.6 & 76.1 & 58.0 & 69.7 & 85.0 & 46.3 & 56.0 & 50.4 & 59.1 \\
\midrule
$\pi_0$~\cite{black2024pi_0}       & 62.3 & 39.8 & 63.1 & 86.0 & 82.8 & 82.4 & 69.6 & 77.7 & 74.1 & 61.4 & 60.1 & 68.2 \\
w/ object head       & \underline{71.7} & \underline{45.8} & \underline{63.5} & \underline{92.4} & \underline{86.9} & 85.1 & \underline{77.4} & 80.6 & \textbf{82.5} & 67.1 & \underline{64.0} & \underline{73.4} \\
w/ skill head        & 70.0 & 45.0 & 61.7 & 90.2 & 83.0 & \textbf{88.4} & 76.3 & 79.8 & 78.9 & \underline{68.9} & 62.7 & 72.5 \\
w/ depth head        & 68.1 & 43.9 & \textbf{65.8} & 90.7 & 83.4 & \underline{85.6} & 72.8 & \underline{81.4} & 79.0 & 65.4 & 61.8 & 71.7 \\
w/ all heads \textbf{(Ours)}      & \textbf{73.7} & \textbf{51.4} & 62.6 & \textbf{94.6} & \textbf{89.0} & 85.2 & \textbf{79.9} & \textbf{84.0} & \underline{80.9} & \textbf{70.8} & \textbf{66.2} & \textbf{75.4} \\

\bottomrule
\end{tabular}}
\end{table*}

We now describe the specific objectives and mechanisms for the three specific factors, as in Alg.~\ref{alg:pg_heads_control_losses}.

\subsubsection{\textbf{Object Head (Visual Grounding)}}
Intuitively, action decoding benefits from attending to semantically meaningful regions, such as the object to be grasped and the target destination. To enforce this, we guide a subset of heads $\mathcal{H}_{\text{obj}}$ to concentrate their attention mass on ground-truth object-region masks.
Given attention probabilities $\bm{P}$ from action queries to all keys, we use mean-head aggregation and first average the selected object heads:
\begin{equation}
\setlength{\abovedisplayskip}{0pt}
\setlength{\belowdisplayskip}{0pt}
\setlength{\abovedisplayshortskip}{0pt}
\setlength{\belowdisplayshortskip}{0pt}
    \bar{P}_{b,t,k}=\frac{1}{|\mathcal{H}_{\mathrm{obj}}|}\sum_{h\in\mathcal{H}_{\mathrm{obj}}}P_{b,h,t,k}.
\end{equation}
Let $M_{b,k}\in[0,1]$ denote the object-region target on the full key axis, with non-object image patches and non-image tokens assigned zero weight. The object mass for action query $t$ is
\begin{equation}
\setlength{\abovedisplayskip}{0pt}
\setlength{\belowdisplayskip}{0pt}
\setlength{\abovedisplayshortskip}{0pt}
\setlength{\belowdisplayshortskip}{0pt}
    m_{b,t}=\sum_k \bar{P}_{b,t,k}M_{b,k}.
\end{equation}
We minimize the negative log object mass over samples whose target object is visible:
\begin{equation}
\setlength{\abovedisplayskip}{0pt}
\setlength{\belowdisplayskip}{0pt}
\setlength{\abovedisplayshortskip}{0pt}
\setlength{\belowdisplayshortskip}{0pt}
    \mathcal{L}_{\text{object}} =
    -\frac{1}{\sum_b v_b|\mathcal{T}_{a}|}
    \sum_b v_b \sum_{t\in\mathcal{T}_{a}}
    \log\left(\max(m_{b,t},\epsilon)\right),
    \label{eq:grounding_loss}
\end{equation}
where $v_b$ indicates that at least one labeled object patch is available for sample $b$, $\mathcal{T}_{a}$ is the set of action queries, and $\epsilon$ is a small numerical constant. To construct $M$, we use foundation models like grounding SAM~\cite{ren2024grounded} to annotate the object to be grasped or the target destination as interested areas and assign zero weight to all other key positions, while allowing the model to decide how to distribute attention within the interest area. The implementation details are provided in Appendix~\appref{sec:head_adapter_impl}, and the stage-aware mask construction is described in Appendix~\appref{sec:object_mask_construction}.

\subsubsection{\textbf{Skill Head (Temporal Logic Intent)}}
Skills capture high-level, temporally extended semantics that modulate the model’s action behaviors in long-horizon tasks. To encode this, we designate a subset of heads $\mathcal{H}_{\text{skill}}$ to specialize in intent recognition. We pool the selected skill-head output features over guided layers, heads, and action queries:
\begin{equation}
\setlength{\abovedisplayskip}{0pt}
\setlength{\belowdisplayskip}{0pt}
\setlength{\abovedisplayshortskip}{0pt}
\setlength{\belowdisplayshortskip}{0pt}
  \bar{\mathbf{f}}_b =
  \frac{1}{|\mathcal{L}_g||\mathcal{H}_{\mathrm{skill}}||\mathcal{T}_a|}
  \sum_{\ell\in\mathcal{L}_g}\sum_{h\in\mathcal{H}_{\mathrm{skill}}}\sum_{t\in\mathcal{T}_a}
  \mathbf{f}_{b,\ell,h,t}.
\end{equation}
We then project the pooled feature to a skill probability distribution and apply a KL-divergence loss against the ground-truth soft label $\mathbf{y}$, which represents the skill distribution over a future horizon:
\begin{equation}
  \mathbf{\hat{p}}_b = \mathrm{softmax}(\bm{W}\bar{\mathbf{f}}_b + \mathbf{b})
\end{equation}
\begin{equation}
\setlength{\abovedisplayskip}{0pt}
\setlength{\belowdisplayskip}{0pt}
\setlength{\abovedisplayshortskip}{0pt}
\setlength{\belowdisplayshortskip}{0pt}
  \mathcal{L}_{\text{skill}} =
  \frac{1}{B}\sum_{b=1}^{B}\sum_{k} y_{b,k}
  \left(\log y_{b,k} - \log \hat{p}_{b,k}\right)
  \label{eq:skill_loss}
\end{equation}
Regarding the annotation of skill types at each time-step, we combine foundation models and manual correction. For LIBERO, the skill distribution covers three effective task-level skill classes plus one null/background class for unannotated or transition frames. The construction and implementation details of the skill-label are provided in Appendix~\appref{sec:skill_label_construction} and Appendix~\appref{sec:head_adapter_impl}.

\subsubsection{\textbf{Depth Head (3D Structure)}}
Since standard vision encoders (e.g., SigLIP) in VLA~\cite{black2024pi_0} are trained with 2D supervision and lack explicit 3D awareness, we design specialized depth heads. Instead of a loss, we use a structural constraint: we extract features from a frozen depth encoder (e.g., DA3~\cite{depthanything3}) on the primary camera view, $F_{\text{Depth}}$, and project them into depth-aware keys and values, $K_{\text{Depth}}$ and $V_{\text{Depth}}$.
The query still comes from the action decoder; we constrain the action-query heads in $\mathcal{H}_{\text{depth}}$ to attend only to these depth-derived keys and values:
\begin{equation}
\setlength{\abovedisplayskip}{0pt}
\setlength{\belowdisplayskip}{0pt}
\setlength{\abovedisplayshortskip}{0pt}
\setlength{\belowdisplayshortskip}{0pt}
    \mathcal{H}_{\text{depth}}:
    \mathrm{softmax}\!\left(
    \frac{Q_{\text{act}}[\mathcal{H}_{\text{depth}}](K_{\text{Depth}})^\top}{\sqrt{d_h}}
    \right)V_{\text{Depth}}.
\end{equation}
This design forces specific heads to specialize in 3D geometry processing. The details of the implementation and the design ablation experiments are provided in Appendix~\appref{sec:head_adapter_impl} and Appendix~\appref{sec:depth-head-ablation}.

In summary, we adopt a mixed loss:
\begin{equation}
\setlength{\abovedisplayskip}{0pt}
\setlength{\belowdisplayskip}{0pt}
\setlength{\abovedisplayshortskip}{0pt}
\setlength{\belowdisplayshortskip}{0pt}
  \mathcal{L} = \mathcal{L}_{\text{FM}} + \lambda_{\text{object}}\mathcal{L}_{\text{object}} + \lambda_{\text{skill}}\mathcal{L}_{\text{skill}}.
  \label{eq:total_loss}
\end{equation}
where $\mathcal{L}_{\text{FM}}$ is the flow matching loss, and $\lambda_{\text{object}}$ and $\lambda_{\text{skill}}$ are the coefficients for the auxiliary objectives that supervise distinct subsets of attention heads. For geometric perception, we inject depth keys and values for depth heads rather than using a loss term. The remaining unsupervised heads are left free to capture purely data-driven patterns, preserving the model's flexibility and expressivity, as in Fig.~\ref{fig:method}.

\subsection{Guidance Dataset Construction}
\label{sec:guidance_dataset_main_paper}

\begin{figure}[htbp]
    \centering
    \includegraphics[width=1.0\linewidth]{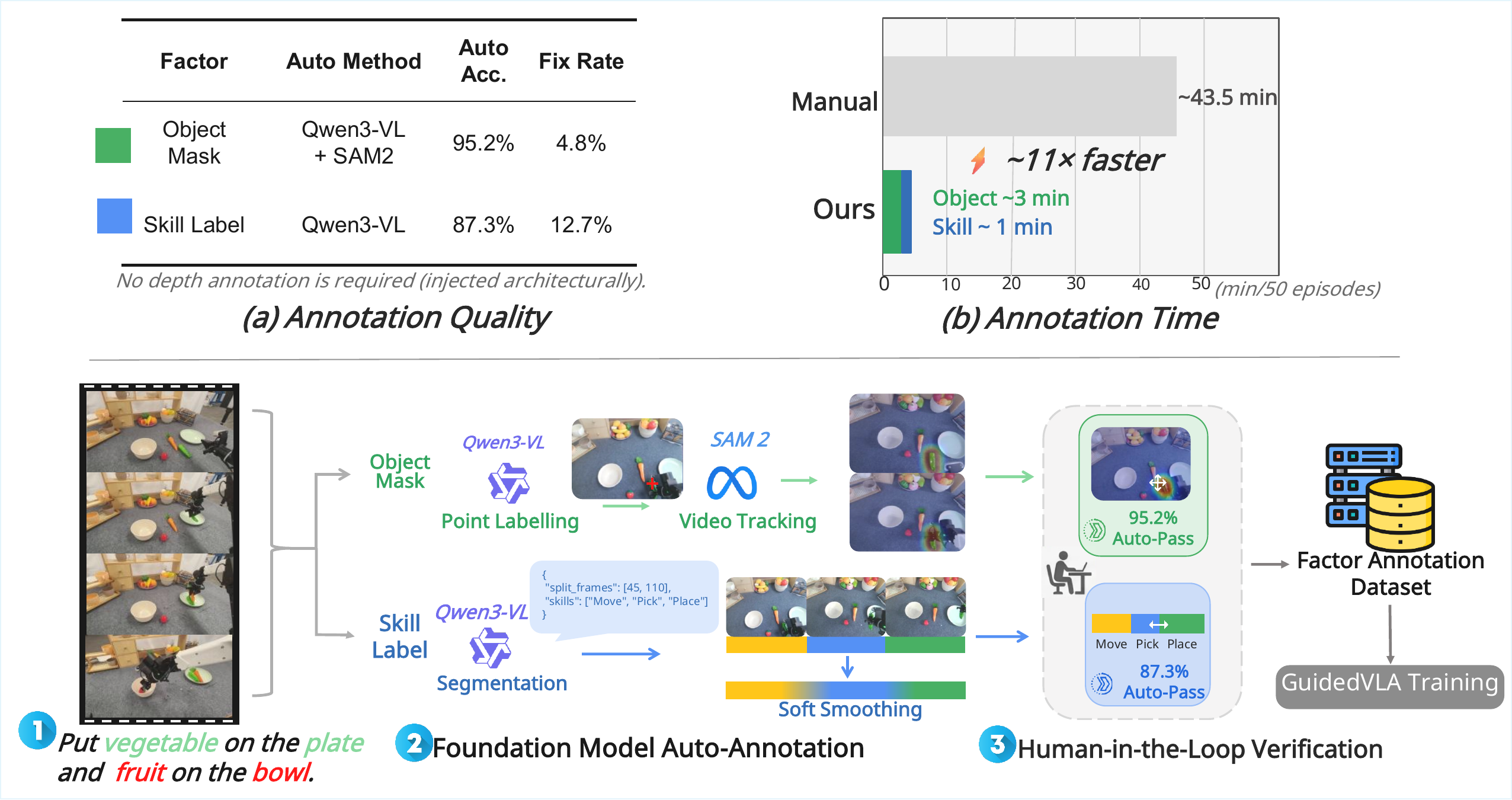}
    \caption{\textbf{Automatic factor annotation pipeline.} Object masks are initialized by Qwen3-VL point prompts and propagated by SAM2, skill labels are generated by Qwen3-VL from stage descriptions and a predefined skill list, and depth guidance uses frozen depth features without requiring depth labels. The pipeline substantially reduces human annotation time while preserving a human verification step for supervision quality.}
    \label{fig:scale_pipeline}
    \vspace{-5em}
\end{figure}

To reduce the annotation burden of factor guidance, we build a highly automatic factor annotation pipeline, as shown in Fig.~\ref{fig:scale_pipeline}. For object grounding, Qwen3-VL~\cite{bai2025qwen3vltechnicalreport} first identifies the task-relevant object from the stage description and proposes foreground point prompts; SAM2~\cite{ravi2025sam} then propagates the corresponding masks through the video segment, followed by human verification. For skill recognition, Qwen3-VL assigns stage-level skill labels from a predefined skill list, which are then converted into soft targets in Eq.~\ref{eq:skill_loss}. Depth guidance does not require manual depth annotation because the depth head directly consumes features from a frozen pretrained depth encoder. In our annotation method, 92\% of the episodes require no human correction; annotating 50 episodes takes about 4 minutes with our pipeline, compared to around 43.5 minutes under manual annotation. The implementation details are in Appendix~\appref{sec:dataset}.

\section{Experiments}

\subsection{Simulation Experiments}

\noindent\textbf{LIBERO-Plus}~\cite{fei2025libero} is a robustness-oriented benchmark built upon LIBERO~\cite{liu2023libero}. It is designed to \textbf{evaluate generalist manipulation policies under distribution shifts}. It introduces perturbations along seven dimensions: camera viewpoint, robot initial state, language variation, lighting condition, background texture, sensor noise, and object layout to expose failure modes under generalization scenario beyond in-domain evaluation. We compare with state-of-the-art baselines in Table~\ref{tab:libero_plus_leaderboard}.

\begin{figure*}[htbp]
    \centering
    \includegraphics[width=1.0\linewidth, trim=2.5mm 0 0 0, clip]{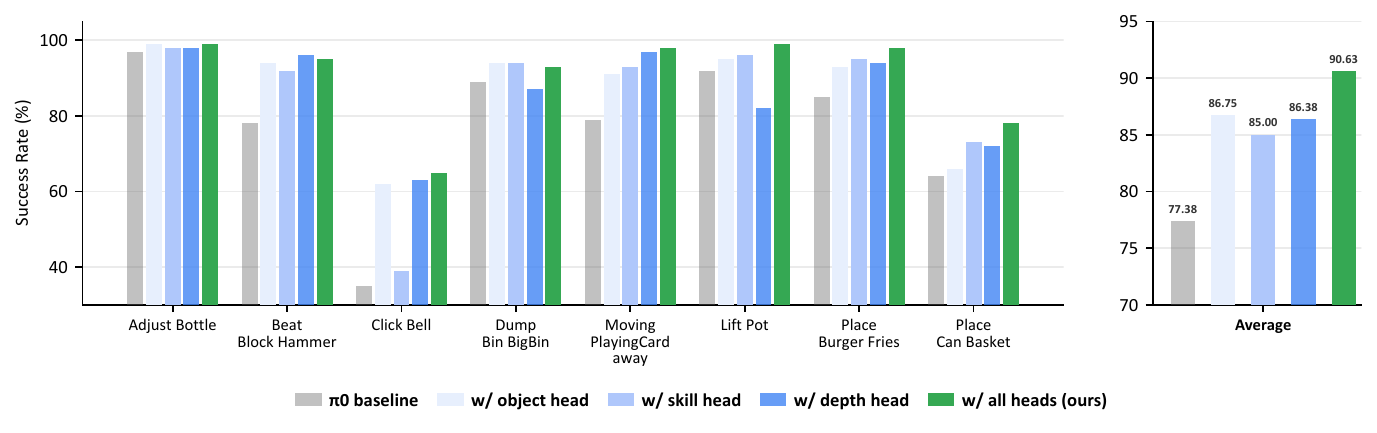}
    \vspace{-6mm}
    \caption{\textbf{RoboTwin 2.0 Benchmark Performance.} Success rates across 8 manipulation tasks comparing the $\pi_0$ baseline, single-head experts, and our full model. While specific heads excel at aligned tasks (e.g., depth head for geometry-heavy Beat Hammer Block), the full model (purple) integrates these capabilities to achieve the best overall average performance ($90.63\%$).}
    \vspace{-2mm}
    \label{fig:robotwin_results}
\end{figure*}

\noindent\textbf{RoboTwin 2.0}~\cite{chen2025robotwin20scalabledata} offers a multi-task evaluation platform across diverse robot embodiments, and leverages extensive scene/object randomization to scale data and enable out-of-distribution testing. As in Fig.~\ref{fig:robotwin_results}, we evaluate on eight representative tasks \textbf{under randomized, unseen settings (out-of-domain task instructions, environments, and object placements)} using the Agilex Piper dual-arm setup.

\subsection{Real-World Experiments}
We conduct real-world experiments on two dual-arm platforms to evaluate both in-domain action generation and cross-platform generalization against baselines.

\begin{figure*}[!ht]
    \centering
    \includegraphics[width=1\linewidth]{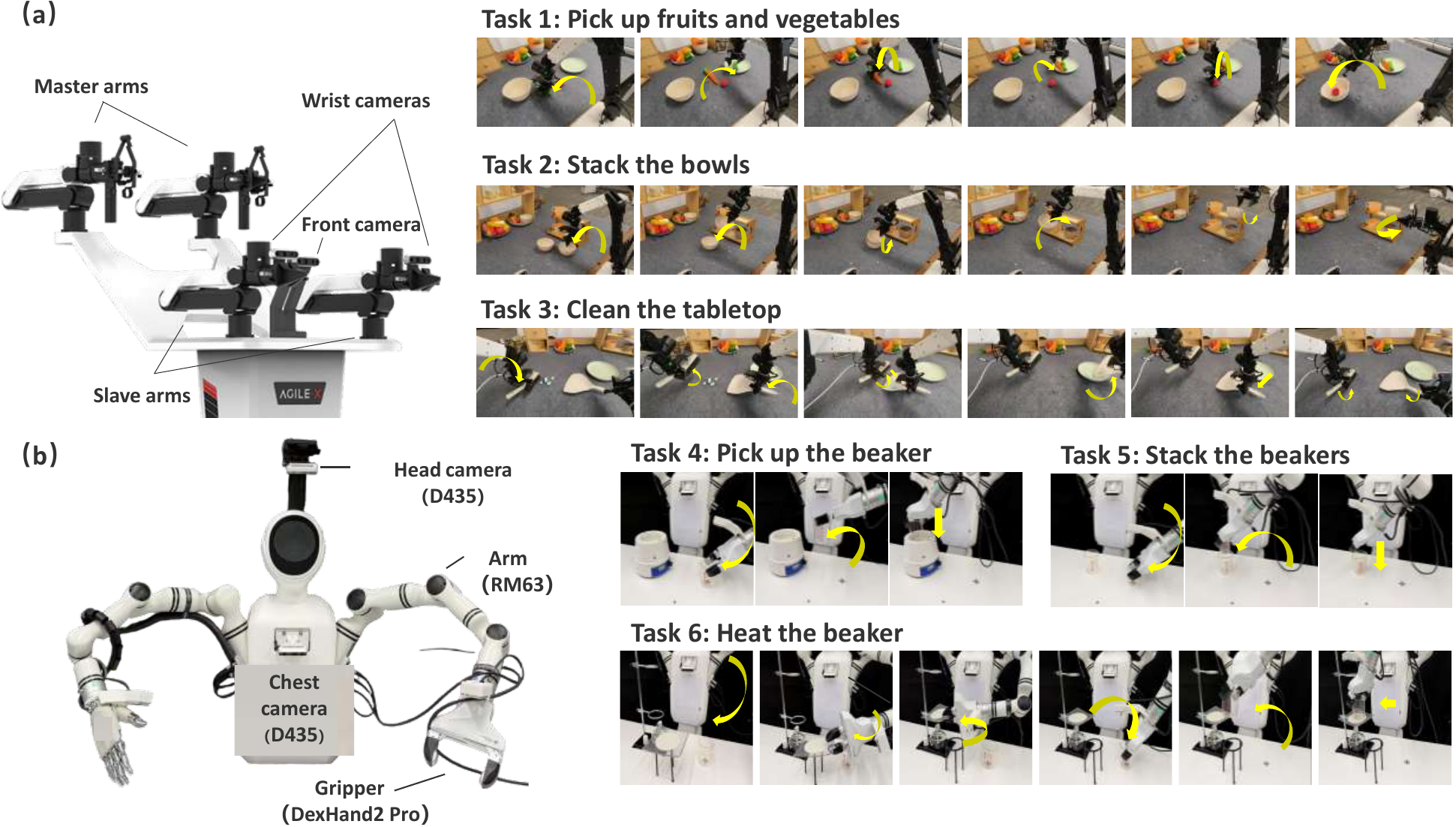}
    \caption{\textbf{Real-world Robot Platforms and Evaluation Tasks.} (a) ALOHA AgileX dual-arm mobile manipulator with left/right wrist Orbbec Dabai cameras and a third-person Orbbec Dabai camera; we evaluate three household tasks: pick up fruits and vegetables, stack the bowls, clean the tabletop. (b) PSI-Bot equipped with RealMan RM63 arm(s) and DexHand2 Pro hands, with head/chest RealSense D435 cameras; we evaluate three lab tasks: pick up beaker, stack beakers, and heat beaker. }
    \label{fig:real_setting}
\end{figure*}

\noindent\textbf{Platforms}: Platform A is an ALOHA AgileX dual-arm system, equipped with two Intel Orbbec Dabai wrist cameras (one per arm) and an additional Intel Orbbec Dabai third-person camera.
Platform B is a PSI-Bot dual-arm platform, using Intel RealSense D435 cameras for visual observations. Figure~\ref{fig:real_setting} summarizes the hardware setups and qualitative task rollouts.

\noindent\textbf{Tasks}: On ALOHA AgileX, we design three household tasks: (1) \emph{pick up fruits and vegetables}: classify and place pepper/carrot on plate, strawberry in bowl, (2) \emph{stack the bowls}: assemble two bowls and place on rack, and (3) \emph{clean the tabletop}: sweep trash with broom/dustpan, pour into tray. On PSI-Bot RealMan, we design three chemistry-lab manipulation tasks: (4) \emph{place beaker in heating mantle}: grasp a beaker and insert it into the heating mantle, (5) \emph{stack beakers}: nest small beakers inside a large one, and (6) \emph{heat beaker}: place an asbestos mesh on the iron stand and then place the beaker on the mesh. These laboratory tasks focus on the manipulation challenges posed by transparent, rigid objects and tight geometric constraints. They do not evaluate complete safety-critical chemical procedures; in particular, \emph{heat beaker} requires placing the beaker onto the heating setup rather than controlling the heating process itself. Detailed success criteria are provided in Appendix~\appref{sec:real_tasks}.

\noindent\textbf{Evaluation protocol}: For each task and model, we perform 20 trials. A trial is successful if the entire task is completed.

\noindent\textbf{Generalization evaluation}: We evaluate three generalization settings on both platforms: \emph{in-domain generalization}, \emph{scene generalization}, and \emph{lighting generalization}. Here, \emph{in-domain generalization} focuses on object position variations within the training distribution, while preserving task semantics and scene layout. \emph{Scene generalization} introduces distracting objects into the workspace, testing robustness to clutter and semantic interference. \emph{Lighting generalization} varies illumination intensity and color temperature, assessing sensitivity to perceptual shifts. Results are summarized in Table~\ref{tab:combined_real_world}, with detailed setting definitions in Appendix~\appref{sec:real_generalization}.

\begin{table*}[htbp]
\centering
\caption{\textbf{Cross-Platform Real-World Generalization.}  Success rates ($N=20$) across three generalization settings on ALOHA and PSI-Bot platforms. Our method consistently outperforms the baseline, achieving performance gains across all settings (up to $52.7\%$) and demonstrating robustness under challenging out-of-domain conditions. Task 1--6 correspond to: (1) pick up fruits and vegetables, (2) stack the bowls, (3) clean the tabletop, (4) place beaker in heating mantle, (5) stack beakers, and (6) heat beaker. In-domain generalization includes variations in object positions within the training distribution.}
\label{tab:combined_real_world}
\small
\setlength{\tabcolsep}{4pt}
\renewcommand{\arraystretch}{1.2}

\scalebox{0.92}{
\begin{tabular}{lVlVcccVcccVc} 
\toprule
\textbf{Generalization} & \textbf{Method} & \multicolumn{3}{c|}{\textbf{ALOHA AgileX}} & \multicolumn{3}{c|}{\textbf{PSI-Bot RealMan}} & \textbf{Average} \\
\cmidrule{3-5} \cmidrule{6-8} \cmidrule{9-9}
\textbf{Setting} & & Task 1 & Task 2 & Task 3 & Task 4 & Task 5 & Task 6 & (\%) \\
\midrule

\multirow{2}{*}{In-Domain}
& Base Policy & 10/20 & 11/20 & 9/20 & 12/20 & 12/20 & 13/20 & 55.8 \\
& \textbf{Ours} & \textbf{14/20} & \textbf{15/20} & \textbf{14/20} & \textbf{16/20} & \textbf{17/20} & \textbf{15/20} & \textbf{75.8}  \\
\midrule

\multirow{2}{*}{Scene}
& Base Policy & 7/20 & 8/20 & 6/20 & 12/20 & 11/20 & 9/20 & 44.2 \\
& \textbf{Ours} & \textbf{13/20} & \textbf{12/20} & \textbf{11/20} & \textbf{15/20} & \textbf{16/20} & \textbf{14/20} & \textbf{67.5} \\
\midrule

\multirow{2}{*}{Lighting}
& Base Policy & 11/20 & 9/20 & 10/20 & 14/20 & 12/20 & 13/20 & 57.5 \\
& \textbf{Ours} & \textbf{13/20} & \textbf{16/20} & \textbf{15/20} & \textbf{17/20} & \textbf{18/20} & \textbf{16/20} & \textbf{79.2} \\
\bottomrule
\end{tabular}}
\vspace{-4mm}
\end{table*}


\section{Analysis}

In this section, we aim to answer the following questions:
\begin{enumerate}
  \item Do VLAs under-utilize vision-language representations in action decoding process, and can explicit factor guidance close this gap? (Section~\ref{sec:factor_quality_analysis})
  \item Does our proposed \alias improve baseline performance under both in-distribution and out-of-distribution evaluations? (Section~\ref{sec:factor_task_contribution})
  \item Which factors (object, skill, geometry) matter most for which task types? (Section~\ref{sec:factor_task_contribution})
  \item Does our attention head specialization indeed lead to learning decoupled features? (Section~\ref{sec:decoupled_feature_learning})
  \item How different architectural choices for guidance influence performance? (Section~\ref{sec:design_choices_each_head})
\end{enumerate}

\subsection{Task-suite Analysis and Cross-benchmark Generalization}
\label{sec:factor_task_contribution}

We analyze how each factor contributes to different task suites and use representative results from simulation and real-world evaluations to explain \emph{why} each specified head helps.

\vspace{2pt}
\noindent\textbf{Object Head: Visual Generalization.}
Tasks involving clutter or distractors necessitate a precise understanding of object instance identities. On the LIBERO-Plus \textit{Object} suite, which stresses object-level distinctions, the object head yields the strongest single-head result (82.5\%, +8.4\% over $\pi_0$, Table~\ref{tab:libero_plus_leaderboard}; full results in Appendix~\appref{sec:complete_results}). This aligns with the intuition that explicit object-centric representations mitigate grounding failures, allowing the policy to filter out irrelevant visual cues that confuse the baseline.

\vspace{2pt}
\noindent\textbf{Skill Head: Temporal Coherence.}
Long-horizon manipulation requires maintaining ``stage awareness'' to transition correctly between sub-skills. On LIBERO-Plus, the skill head gives the best single-head result on the \textit{Goal} suite (68.9\%) and remains above $\pi_0$ on the \textit{Long} suite (62.7\% vs. 60.1\%, Table~\ref{tab:libero_plus_leaderboard}). Similarly, for the \textit{Lift Pot} task (RoboTwin 2.0) involving a strict sequence of grasping, stabilizing, and lifting, this head achieves the best single-head success rate (96\%, Fig.~\ref{fig:robotwin_results}; full results in Appendix~\appref{sec:complete_results}). These results validate that explicit skill recognition provides the temporal scaffolding needed to prevent premature termination or mode collapse during phase transitions.

\vspace{2pt}
\noindent\textbf{Depth Head: Geometric Precision.}
Tasks reliant on precise 3D localization, such as pressing or insertion, necessitate accurate depth estimation which 2D backbone alone often fails to provide. On RoboTwin 2.0, the \textit{Click Bell} task requires precise Z-axis control to trigger the mechanism without collision; the depth head drastically improves performance from 35\% to 63\% (Fig.~\ref{fig:robotwin_results}). We observe a similar trend in \textit{Beat Hammer Block} (78\% $\rightarrow$ 96\%), where height alignment is critical. These gains confirm that explicit geometric cues compensate for the lack of 3D observability in standard VLA inputs.

\vspace{2pt}
\noindent\textbf{Full Model: Synergetic Generalization.}
The full model integrates these complementary strengths---visual grounding, temporal coherence, and geometric precision---to achieve robust generalization across diverse domains. It raises the average success on RoboTwin 2.0 from 77.38\% to 90.63\% (Fig.~\ref{fig:robotwin_results}) and demonstrates superior robustness in the real world (Table~\ref{tab:combined_real_world}; head-wise real-robot diagnostics in Appendix~\appref{sec:real_head_diagnostics}). Crucially, while single heads excel in their respective niches, the full model is the only variant that reliably generalizes across \emph{all} dimensions of variability (scene, lighting, and position), highlighting that these factors are not redundant but mutually reinforcing.

\subsection{Sensitivity Analysis: Does Factor Quality Matter?}
\label{sec:factor_quality_analysis}

\begin{figure*}[!ht]
    \includegraphics[width=\linewidth]{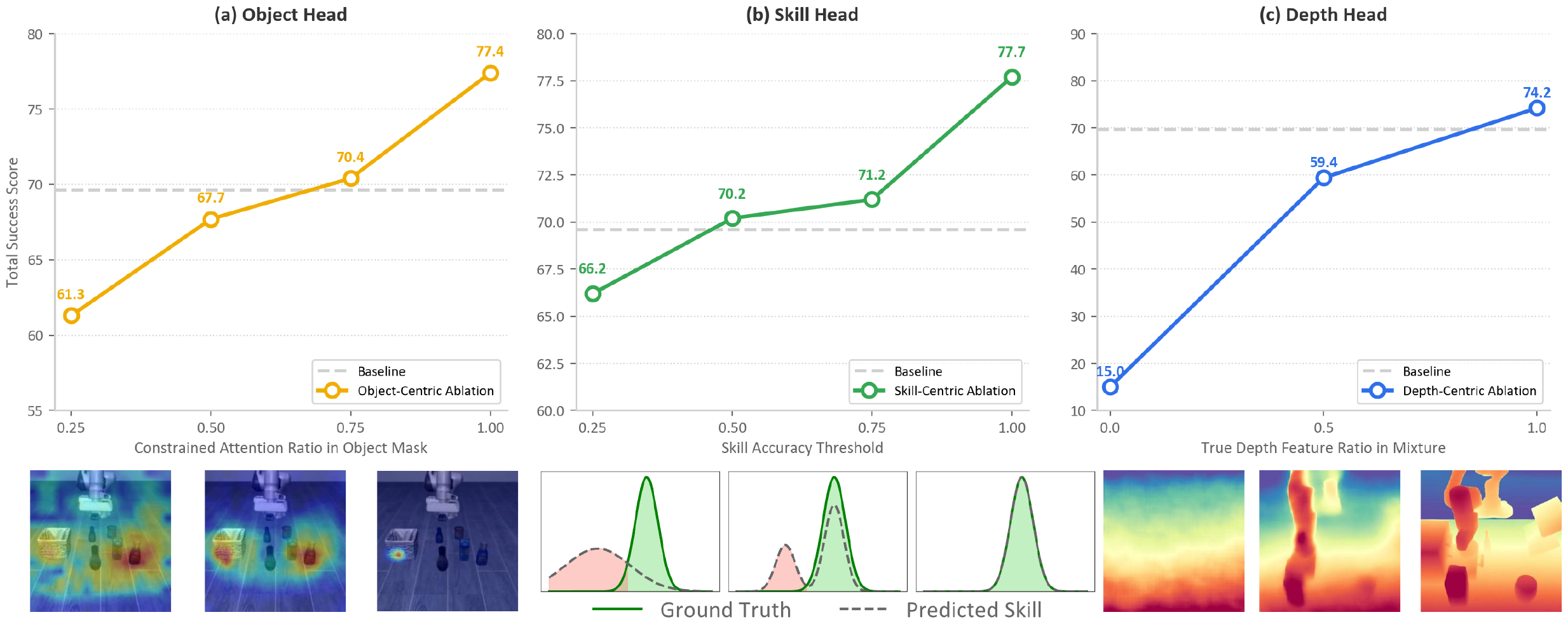}
    \caption{\textbf{Higher Factor Quality Leads to Better Task Performance.} \textbf{Top:} Quantitative analysis on the LIBERO-Plus layout perturbation track shows that improving the quality of each specialized head consistently boosts success rates. \textbf{(a) Object Head:} as the proportion of attention focused on task-relevant object regions increases, success rises from 61.3\% to 77.4\%, highlighting the importance of precise object-centric attention. \textbf{(b) Skill Head:} higher skill-recognition accuracy, measured by a linear probe, correlates with improved performance (66.2\% to 77.7\%), indicating that better temporal understanding enhances control. \textbf{(c) Depth Head:} increasing the ratio of true depth features (versus noise) dramatically improves both qualitative depth estimation and quantitative success (15.0\% to 74.2\%), confirming that explicit 3D cues are critical for robust manipulation. \textbf{Bottom:} Qualitative visualizations show how changes along the x-axis metrics are reflected in the corresponding feature representations.}
    \vspace{-6mm}
    \label{fig:factor_quality_analysis}
\end{figure*}

We move from a binary ``with/without'' comparison to a quantitative question: does better factor quality lead to higher success? We vary each factor's strength in controlled ablations and measure continuous proxies aligned with the intended semantics, as summarized in Figure~\ref{fig:factor_quality_analysis}.

\noindent\textbf{Object Grounding.} We measure the fraction of attention mass falling inside the object/gripper mask, using the same supervision target as Eq.~\ref{eq:grounding_loss}. As the mask-aligned attention ratio increases from 0.25 to 1.0, success rises from 61.3\% to 74.6\% (Figure~\ref{fig:factor_quality_analysis} (a)). In contrast, $\pi_0$ exhibits low intrinsic object focus (26.5\%), indicating that stronger spatial grounding directly correlates with performance. Under a stricter localization diagnostic, \alias increases object-region attention mass from 8.1\% to 84.0\% and argmax-hit accuracy from 2.2\% to 84.7\% over $\pi_0$.

\noindent\textbf{Skill Recognition.} We use a linear probe to predict task skill labels from action features, using probe accuracy as the quality metric. Raising the skill accuracy threshold from 0.25 to 1.0 increases success from 66.2\% to 72.9\% (Figure~\ref{fig:factor_quality_analysis} (b)). The baseline $\pi_0$ remains low at 48.4\%, confirming improved temporal representations translate into better performance.

\noindent\textbf{Geometry Perception.} We modulate the true depth feature ratio in the geometry stream as a proxy for geometric signal strength. Increasing this ratio from 0 to 1.0 yields a large success gain (15.6\% $\rightarrow$ 76.7\%, Figure~\ref{fig:factor_quality_analysis} (c)), demonstrating that richer geometric cues substantially improve task outcomes.

\begin{figure}
    \centering
    \includegraphics[width=1.0\linewidth]{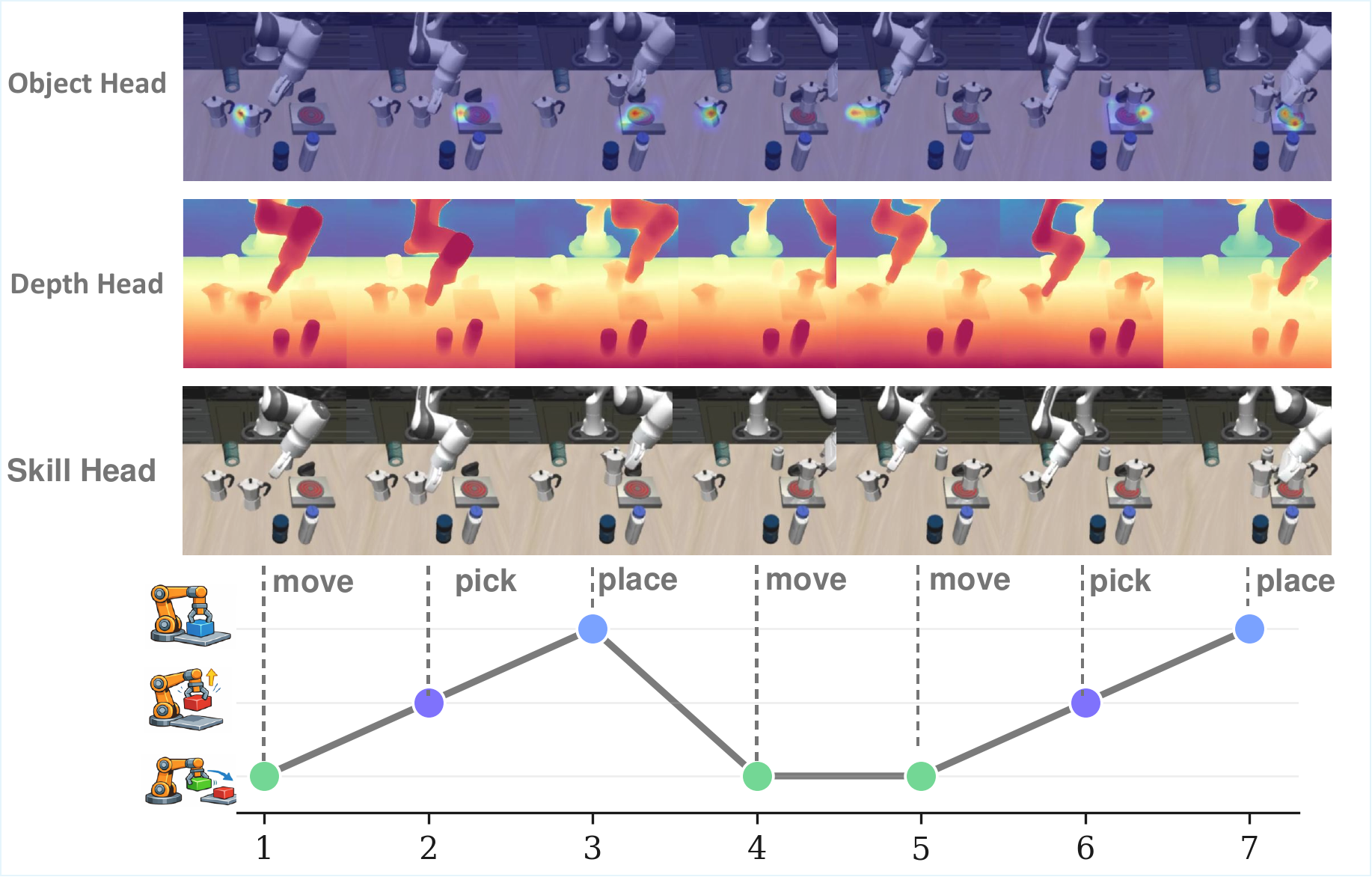}
    \caption{\textbf{Visualization of Learned Representations in \alias.} From top to bottom: \textbf{(i)} Object attention focuses on the manipulation target (e.g., pot handle); \textbf{(ii)} Depth features encode explicit 3D structure; \textbf{(iii)} Skill predictions track the temporal progress of task phases. This confirms that each head specializes in its designated semantic factor as intended.}
    \vspace{-5mm}\label{fig:qualitative_visualization_3heads}
\end{figure}

Figure~\ref{fig:factor_quality_analysis} summarizes these trends, showing that success increases monotonically with each factor's quality, not merely its presence.
Figure~\ref{fig:qualitative_visualization_3heads} provides a qualitative analysis of factor features over time in a task.
Details of these metrics and ablations are provided in Appendix~\appref{sec:factor_quality_ablation_impl}.

\subsection{Specialization Enables Decoupled Feature Learning}
\label{sec:decoupled_feature_learning}

\begin{figure}[htbp]
    \centering
    \includegraphics[width=\columnwidth]{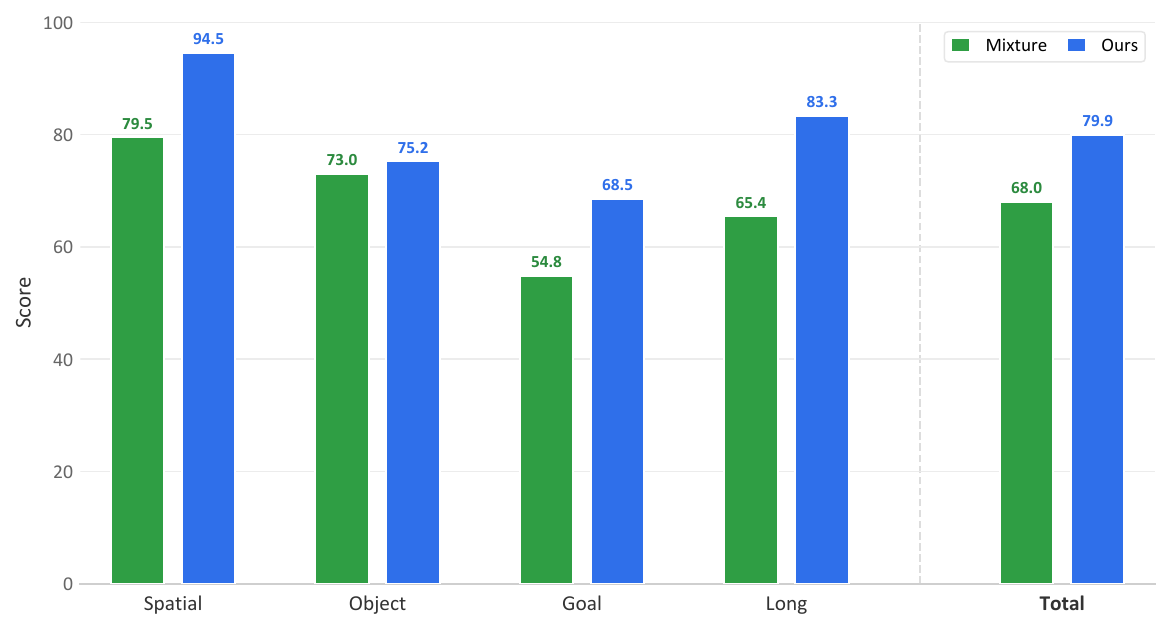}
    \caption{\textbf{Comparison of \alias against Mixture Alternative.} Attention head specialization explicitly outperforms learning all objectives in a mixture.}
    \vspace{-2em}
    \label{fig:ours_vs_mixture}
\end{figure}

We have shown that each factor (object grounding, geometry, and skill) correlates with task success. A natural next question is: \emph{can we guide multiple factors by training all heads with all factor objectives?} Our answer is \textbf{NO}: a naive mixed training protocol consistently underperforms (Figure~\ref{fig:ours_vs_mixture}).
The gain is not just extra supervision or capacity: under matched control settings, \alias consistently outperforms the shared-head mixture alternative and other non-factorized controls; additional architecture ablations are provided in Appendix~\appref{sec:overall_architecture_ablation}.

\begin{figure}[htbp]
  \centering
  \begin{subfigure}[b]{0.48\columnwidth}
    \includegraphics[width=\linewidth, trim=4mm 4mm 4mm 4mm, clip]{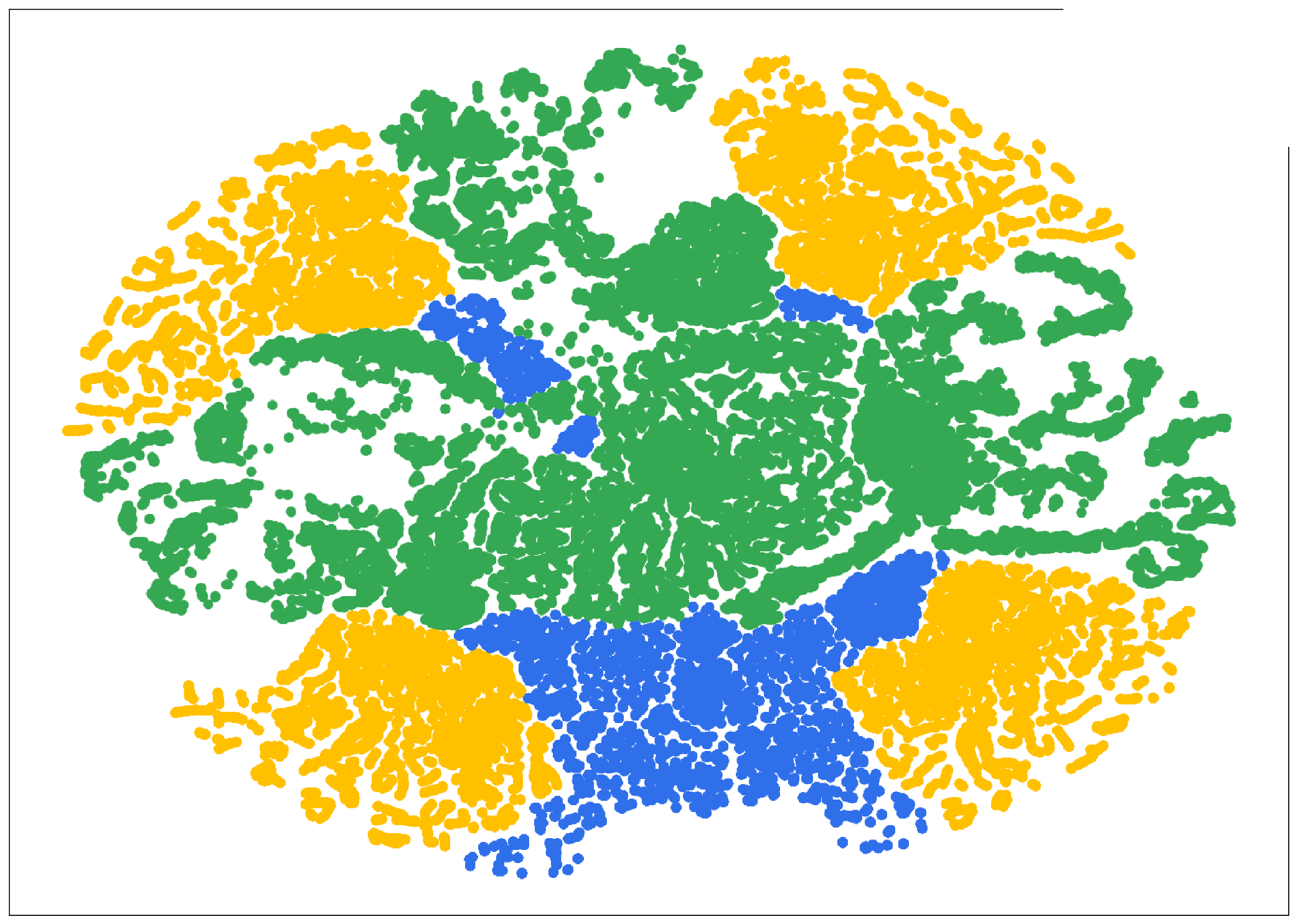}
    \caption{Ours}
    \label{fig:tsne_attention_output_decoupled}
  \end{subfigure}
  \begin{subfigure}[b]{0.48\columnwidth}
    \includegraphics[width=\linewidth, trim=4mm 4mm 4mm 4mm, clip]{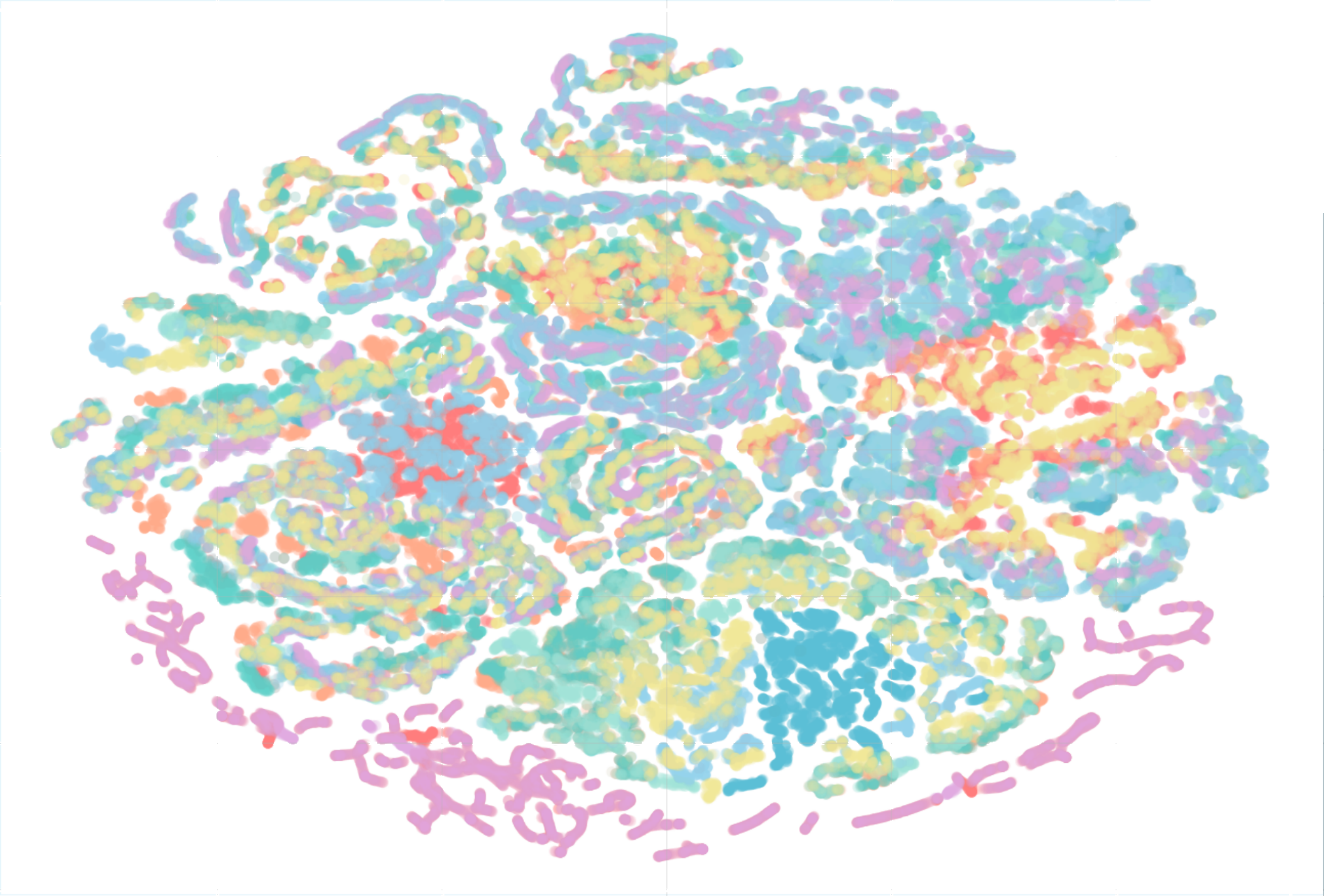}
    \caption{Mixture}
    \label{fig:tsne_attention_output_mixture}
  \end{subfigure}
  \hfill
  \caption{\textbf{t-SNE Visualization of Attention Outputs.} (a) Specialized attention heads (object: yellow, depth: blue, skill: green) form well-separated clusters, demonstrating factor disentanglement and minimal interference. (b) The mixture alternative shows overlapping clusters (different colors representing different heads), indicating entangled representations.}
  \vspace{-6mm}
  \label{fig:tsne_attention_output}
\end{figure}

When object grounding, geometry, and skill objectives are all supervised through all attention heads, their features become entangled, as in Fig.~\ref{fig:tsne_attention_output}. This coupling means that information from different factors is mixed, making it difficult to capture each factor clearly, thus leading to degraded performance.


\subsection{Comparison to Other Factor Guidance Approaches}
\label{sec:compare_other_approaches}

There exist several representative paradigms for introducing task-relevant factors into VLA models. DreamVLA~\cite{zhang2025dreamvla} uses extra VLM queries to predict dynamic regions, depth maps, and semantic knowledge. VLA-Adapter~\cite{wang2026vlaadapter} injects useful vision-language conditions into action prediction through Bridge Attention. Spatial Forcing~\cite{spatialforcing} provides implicit spatial guidance by aligning VLA visual embeddings with 3D foundation-model features. AdaMoE~\cite{shen2025expertise} learns task-adaptive action experts with a mixture-of-experts architecture.

As shown in Table~\ref{tab:libero_plus_leaderboard}, our method outperforms DreamVLA (69.9\%), VLA-Adapter (59.1\%), AdaMoE (50.1\%), and Spatial Forcing (29.1\%) in overall average success rate (75.4\%), with strong gains across most perturbation dimensions and task suites, especially in challenging settings such as camera, robot, and layout perturbations.  We attribute these gains to our explicit attention head specialization, which enables the model to disentangle and robustly capture object grounding, skill recognition, and geometric cues.

\subsection{Ablation of Design Choices}
\label{sec:design_choices_each_head}
In this section, we discuss the design choices for each specified head and the plug-and-play ControlNet-style residual adapter. We systematically conduct experiments on these choices in RoboTwin 2.0, with detailed results in Appendix~\appref{sec:head_adapter_design_ablation}.

\noindent\textbf{Object Head}. Suppressing attention outside object regions outperforms enforcing a fixed spatial prior (e.g., Gaussian) inside object regions. This allows the model to flexibly learn which object parts are most relevant at each task stage.

\noindent\textbf{Skill Head}. Soft targets outperform one-hot labels, better handling ambiguous or mixed-intent segments and thus leading to more stable training and better performance.

\noindent\textbf{Depth Head}. Adopting a lightweight downsampling adapter outperforms directly using original, non-downsampled depth features, since the number of depth tokens can be large and make the learning process difficult.

\noindent\textbf{Extra Branch}. A zero-initialized control branch introduces auxiliary signals gradually and does not disrupt the base model at the start of training, outperforming fusion without this branch.

\section{Conclusion}
We present \alias, a method that makes the VLA action decoding process more robust by explicitly specifying task-relevant factors through attention head specialization. By assigning dedicated heads to object grounding, temporal skill logic, and geometric cues, \alias transforms the action decoder from an entangled black box into a set of semantically decoupled pathways. Across simulation benchmarks and real-robot evaluations, this design delivers consistent gains in both in-domain performance and robustness under distribution shift. Further analyses show that (i) higher-quality factor signals correlate with higher task success, and (ii) allocating a dedicated head per factor produces clearly decoupled features. Together, these results point toward training VLAs that are both more interpretable and more generalizable.

\textbf{Limitations and Future Work.} Our method relies on predefined factors, and automating factor discovery remains an open challenge, especially for continuous tasks where automatic skill labeling is difficult. Promising directions include automatic skill discovery~\cite{zhu2022bottomup,wan2024lotus} and the use of continuous progress signals as latent skill targets~\cite{chen2026topreward}.

\section{Acknowledgment}
This work is supported by the National Natural Science Foundation of China (Grant No. 62521004),  the Science and Technology Commission of Shanghai Municipality (No. 24511103100) and the New Cornerstone Science Foundation through the XPLORER PRIZE. This work is also in part supported by Scientific Research Innovation Capability Support Project for Young Faculty (U40) of the Ministry of Education of China, SRICSPYF-ZY2025019.




\bibliographystyle{plainnat}
\bibliography{references}

\clearpage
\appendices
\appendix



\subsection{Contributions}

\noindent\textbf{Project Leaders:} Xiaosong Jia, Cunxin Fan.

\noindent\textbf{Technical Roadmap \& Methodology:} Xiaosong Jia, Cunxin Fan, Qingwen Bu, Bowen Yang, Xian Nie, Zuhao Ge, Hongyang Li, Haidong Cao, Chao Wu, Qifeng Li, Zhenjie Yang, Chenhe Zhang, Zuxuan Wu, Junchi Yan, Yu-Gang Jiang.

\noindent\textbf{Implementation \& Iteration:}
\begin{itemize}
    \setlength{\itemsep}{2pt}
    \item \textit{Object Head:} Bowen Yang, Xian Nie, Cunxin Fan, Xiaosong Jia.
    \item \textit{Depth Head:} Xian Nie, Yilin Chai, Zijian Liang, Bowen Yang, Cunxin Fan, Xiaosong Jia.
    \item \textit{Skill Head:} Zuhao Ge, Xian Nie, Bowen Yang, Cunxin Fan, Xiaosong Jia.
    \item \textit{Head Merge:} Bowen Yang, Xian Nie, Zuhao Ge, Cunxin Fan, Xiaosong Jia.
    \item \textit{Codebase \& Infra:} Bowen Yang, Cunxin Fan, Xiaosong Jia.
\end{itemize}

\noindent\textbf{Simulation Experiments:} Bowen Yang, Xian Nie, Zuhao Ge.

\noindent\textbf{Real-Robot Experiments:} Zuhao Ge, Yuchen Zhou, Yufeng Li, Chao Jing.

\noindent\textbf{Writing \& Illustration:} Xiaosong Jia, Cunxin Fan, Bowen Yang, Xian Nie, Chao Jing, Zuhao Ge, Yuchen Zhou, Qingwen Bu.

\subsection{Complete Results for All Datasets}
\label{sec:complete_results}

For Real robot experiments, we provide the complete results across all 6 tasks in Table~\ref{tab:real_headwise}.

\begin{table*}[htbp]
\centering
\caption{\textbf{Head-wise real-robot results on factor-aligned tasks.}
We report success rates for the base $\pi_0$ policy, three single-head diagnostic variants (Object-only / Depth-only / Skill-only), and the full GuidedVLA under three distribution shifts: positional (in-domain), scene, and lighting generalization.
To isolate factor contributions, each single-head variant is evaluated \emph{only} on its aligned tasks:
Object$\rightarrow$T1/T4, Depth$\rightarrow$T2/T5, Skill$\rightarrow$T3/T6; other entries are not evaluated and shown as ``--''.}
\label{tab:real_headwise}
\vspace{-6pt}
\setlength{\tabcolsep}{6pt}
\renewcommand{\arraystretch}{1.15}
\begin{tabular}{l l c c c c c c c}
\toprule
\multirow{2}{*}{Setting} & \multirow{2}{*}{Method} &
\multicolumn{3}{c}{ALOHA AgileX} & \multicolumn{3}{c}{PSI-Bot} & \multirow{2}{*}{Avg.} \\
\cmidrule(lr){3-5}\cmidrule(lr){6-8}
& & \textbf{T1 (Object)} & \textbf{T2 (Depth)} & \textbf{T3 (Skill)} &
\textbf{T4 (Object)} & \textbf{T5 (Depth)} & \textbf{T6 (Skill)} & \\
\midrule

\multirow{5}{*}{In-Domain}
& $\pi_0$ (Base)          & 10/20 & 11/20 &  9/20 & 12/20 & 12/20 & 13/20 & 55.8 \\
& Object-only             & 11/20  & --    & --    & 14/20  & --    & --    & 62.5 \\
& Depth-only              & --    & 13/20  & --    & --    & 14/20  & --    & 67.5 \\
& Skill-only              & --    & --    & 12/20  & --    & --    & 13/20  & 62.5 \\
& GuidedVLA (Full)        & 14/20 & 15/20 & 14/20 & 16/20 & 17/20 & 15/20 & 75.8 \\
\midrule

\multirow{5}{*}{Scene}
& $\pi_0$ (Base)          &  7/20 &  8/20 &  6/20 & 12/20 & 11/20 &  9/20 & 44.2 \\
& Object-only             & 10/20  & --    & --    & 12/20  & --    & --    & 55 \\
& Depth-only              & --    & 11/20  & --    & --    & 13/20  & --    & 60 \\
& Skill-only              & --    & --    & 9/20  & --    & --    & 12/20  & 52.5 \\
& GuidedVLA (Full)        & 13/20 & 12/20 & 11/20 & 15/20 & 16/20 & 14/20 & 67.5 \\
\midrule

\multirow{5}{*}{Lighting}
& $\pi_0$ (Base)          & 11/20 &  9/20 & 10/20 & 14/20 & 12/20 & 13/20 & 57.5 \\
& Object-only             & 11/20  & --    & --    & 15/20  & --    & --    & 65 \\
& Depth-only              & --    & 13/20  & --    & --    & 15/20  & --    & 70 \\
& Skill-only              & --    & --    & 12/20  & --    & --    & 14/20  & 65 \\
& GuidedVLA (Full)        & 13/20 & 16/20 & 15/20 & 17/20 & 18/20 & 16/20 & 79.2 \\
\bottomrule
\end{tabular}
\vspace{-10pt}
\end{table*}

For LIBERO-plus~\cite{fei2025libero} benchmark, we provide the complete results in Table~\ref{tab:libero_plus_full_results}.

\begin{table*}[htbp]
\centering
\small
\setlength{\tabcolsep}{6pt}
\renewcommand{\arraystretch}{1.05}
\caption{Full Results on LIBERO-Plus Benchmark.}
\begin{tabular}{lcccccccc}
\toprule
 & Camera & Robot & Language & Light & Background & Noise & Layout & Total \\
\midrule

\rowcolor{gray!8}
\multicolumn{9}{c}{\rule{0pt}{2.6ex}\textbf{DreamVLA}\rule{0pt}{2.2ex}} \\
\midrule
Spatial & 79.3 & 46.0 & 64.4 & 96.9 & 96.0 & 93.1 & 90.1 & 79.7 \\
Object  & 85.4 & 38.4 & 80.2 & 94.3 & 93.5 & 91.9 & 77.9 & 79.0 \\
Goal    & 58.5 & 40.8 & 39.7 & 82.7 & 80.8 & 84.4 & 59.3 & 61.7 \\
Long    & 39.2 & 38.9 & 72.7 & 67.5 & 63.3 & 72.6 & 69.2 & 59.8 \\
Avg     & 65.0 & 40.9 & 63.5 & 85.7 & 82.7 & 85.0 & 74.0 & 69.9 \\
\midrule

\rowcolor{gray!8}
\multicolumn{9}{c}{\rule{0pt}{2.6ex}\textbf{AdaMoE}\rule{0pt}{2.2ex}} \\
\midrule
Spatial & 55.1 & 12.0 & 20.8 & 72.9 & 76.4 & 62.4 & 69.1 & 51.0 \\
Object  & 73.7 & 14.5 & 29.6 & 85.5 & 83.9 & 62.2 & 69.5 & 57.9 \\
Goal    & 65.9 & 25.4 & 12.9 & 81.7 & 81.9 & 58.6 & 64.9 & 53.3 \\
Long    & 22.1 & 17.3 & 20.5 & 53.6 & 55.0 & 52.1 & 58.0 & 38.1 \\
Avg     & 53.8 & 17.5 & 20.6 & 73.7 & 73.8 & 58.6 & 65.8 & 50.1 \\
\midrule

\rowcolor{gray!8}
\multicolumn{9}{c}{\rule{0pt}{2.6ex}\textbf{$\pi_0$}\rule{0pt}{2.2ex}} \\
\midrule
Spatial & 71.3 & 52.3 & 68.2 & 92.8 & 91.9 & 87.2 & 87.3 & 77.7 \\
Object  & 76.3 & 33.2 & 79.1 & 92.9 & 87.9 & 87.7 & 71.2 & 74.1 \\
Goal    & 63.7 & 40.1 & 45.1 & 79.2 & 82.6 & 81.5 & 51.5 & 61.4 \\
Long    & 39.6 & 35.1 & 62.3 & 78.1 & 70.6 & 74.5 & 70.5 & 60.1 \\
Avg     & 62.3 & 39.8 & 63.1 & 86.0 & 82.8 & 82.4 & 69.6 & 68.2 \\
\midrule

\rowcolor{gray!8}
\multicolumn{9}{c}{\rule{0pt}{2.6ex}\textbf{$\pi_0$ w/ object head}\rule{0pt}{2.2ex}} \\
\midrule
Spatial & 81.9 & 54.3 & 67.4 & 95.2 & 96.1 & 87.7 & 88.6 & 80.6 \\
Object  & 89.4 & 46.0 & 83.6 & 97.3 & 98.4 & 94.8 & 77.4 & 82.5 \\
Goal    & 70.6 & 44.5 & 40.5 & 91.0 & 85.1 & 85.8 & 66.8 & 67.1 \\
Long    & 46.8 & 39.4 & 65.5 & 85.4 & 70.6 & 73.4 & 77.9 & 64.0 \\
Avg     & 71.7 & 45.8 & 63.5 & 92.4 & 86.9 & 85.1 & 77.4 & 73.4 \\
\midrule

\rowcolor{gray!8}
\multicolumn{9}{c}{\rule{0pt}{2.6ex}\textbf{$\pi_0$ w/ skill head}\rule{0pt}{2.2ex}} \\
\midrule
Spatial & 79.5 & 54.0 & 64.6 & 93.8 & 92.2 & 91.7 & 89.4 & 79.8 \\
Object  & 86.4 & 40.2 & 76.6 & 94.3 & 87.9 & 95.3 & 77.7 & 78.9 \\
Goal    & 70.3 & 51.8 & 44.9 & 92.8 & 88.6 & 86.1 & 62.8 & 68.9 \\
Long    & 45.8 & 34.9 & 62.9 & 79.2 & 65.1 & 81.4 & 76.9 & 62.7 \\
Avg     & 70.0 & 45.0 & 61.7 & 90.2 & 83.0 & 88.4 & 76.3 & 72.5 \\
\midrule

\rowcolor{gray!8}
\multicolumn{9}{c}{\rule{0pt}{2.6ex}\textbf{$\pi_0$ w/ depth head}\rule{0pt}{2.2ex}} \\
\midrule
Spatial & 81.6 & 54.0 & 69.5 & 96.2 & 92.8 & 88.9 & 92.2 & 81.4 \\
Object  & 84.1 & 44.7 & 79.7 & 92.9 & 92.5 & 92.9 & 73.9 & 79.0 \\
Goal    & 70.6 & 44.0 & 45.1 & 89.2 & 86.4 & 84.7 & 53.6 & 65.4 \\
Long    & 38.4 & 34.1 & 71.3 & 83.9 & 64.1 & 76.9 & 73.4 & 61.8 \\
Avg     & 68.1 & 43.9 & 65.8 & 90.7 & 83.4 & 85.6 & 72.8 & 71.7 \\
\midrule

\rowcolor{gray!8}
\multicolumn{9}{c}{\rule{0pt}{2.6ex}\textbf{\alias\ (Ours)}\rule{0pt}{2.2ex}} \\
\midrule
Spatial & 86.4 & 60.6 & 65.9 & 99.3 & 95.7 & 92.3 & 94.5 & 84.0 \\
Object  & 86.6 & 52.0 & 77.1 & 94.3 & 96.8 & 92.4 & 75.2 & 80.9 \\
Goal    & 75.7 & 50.6 & 42.4 & 96.8 & 92.9 & 85.2 & 68.5 & 70.8 \\
Long    & 48.2 & 43.5 & 67.4 & 87.6 & 72.7 & 72.8 & 83.3 & 66.2 \\
Avg     & 73.7 & 51.4 & 62.6 & 94.6 & 89.0 & 85.2 & 79.9 & 75.4 \\
\bottomrule

\end{tabular}
\label{tab:libero_plus_full_results}
\end{table*}

For the RoboTwin 2.0~\cite{chen2025robotwin20scalabledata} benchmark, we provide the complete results in Table~\ref{tab:robotwin_full_results}.

\begin{table*}[t]
\centering
\small
\setlength{\tabcolsep}{5pt}
\renewcommand{\arraystretch}{1.15}
\caption{\textbf{RoboTwin 2.0 Benchmark Full Results.}}
\begin{tabular}{l!{\color{gray!60}\vrule width 0.6pt}cccc!{\color{gray!60}\vrule width 0.6pt}c}
\toprule
\multirow{2}{*}{Model}
& Adjust Bottle
& Beat Hammer Block
& Click Bell
& Dump Bin BigBin
& \multirow{2}{*}{Avg} \\
& Moving PlayingCard away
& Lift Pot
& Place Burger Fries
& Place Can Basket
& \\
\midrule

\multirow{2}{*}{$\pi_0$}
& 97\% & 78\% & 35\% & 89\% & \multirow{2}{*}{77.38\%} \\
& 79\% & 92\% & 85\% & 64\% & \\
\midrule

\multirow{2}{*}{$\pi_0$ w/ object head}
& \textbf{99\%} & 94\% & 62\% & \textbf{94\%} & \multirow{2}{*}{\underline{86.75\%}} \\
& 91\% & 95\% & 93\% & 66\% & \\
\midrule

\multirow{2}{*}{$\pi_0$ w/ skill head}
& \underline{98\%} & 92\% & 39\% & \textbf{94\%} & \multirow{2}{*}{85.00\%} \\
& 93\% & \underline{96\%} & \underline{95\%} & 73\% & \\
\midrule

\multirow{2}{*}{$\pi_0$ w/ depth head}
& \underline{98\%} & \textbf{96\%} & \underline{63\%} & 87\% & \multirow{2}{*}{86.38\%} \\
& \underline{97\%} & 82\% & 94\% & \underline{74\%} & \\
\midrule

\multirow{2}{*}{$\pi_0$ w/ all heads \textbf{(Ours)}}
& \textbf{99\%} & \underline{95\%} & \textbf{65\%} & \underline{93\%} & \multirow{2}{*}{\textbf{90.63\%}} \\
& \textbf{98\%} & \textbf{99\%} & \textbf{98\%} & \textbf{78\%} & \\
\bottomrule

\end{tabular}
\label{tab:robotwin_full_results}
\end{table*}

\subsection{Implementation Details of Each Head and Adapter}
\label{sec:head_adapter_impl}

\noindent \textbf{Object Head.}
We implement object-level attention supervision by maximizing the attention mass assigned to stage-specific object masks, matching Eq.~\mainref{eq:grounding_loss} in the main paper. Algorithm~\ref{alg:object_attn_baseline} outlines the procedure for computing the object grounding loss. It selects a subset of attention heads, indexes image-view patches, converts stage-aware masks into valid object regions, and penalizes attention mass outside the target region. The supervision is applied across multiple transformer layers and averaged to produce the final loss.

\begin{algorithm*}[htbp]
\caption{Python Pseudocode of Applying Object Grounding Loss on Object Head}
\label{alg:object_attn_baseline}
\begin{lstlisting}[
  language=Python,
  basicstyle=\ttfamily\footnotesize,
  keywordstyle=\color{kwpurple}\bfseries,
  commentstyle=\color{cmteal},
  showstringspaces=false,
  columns=fullflexible,
  keepspaces=true,
  breaklines=true,
  breakatwhitespace=true,
  xleftmargin=0.8em,
  frame=none
]
def object_attn_guidance(
    all_attn_states,
    teacher_attn_maps,          # {"object_maps": ..., "object_masks": ...}
    object_head_indices,
    att_mask,
    view_patch_indices,
    action_query_start_idx,
):
    layer_losses = []
    for layer_idx, attn_data in all_attn_states:
        Q, K = get_qk(attn_data)
        P = attention_probs(Q, K, att_mask)                 # [B, H, S, S]
        P = P[:, object_head_indices, action_query_start_idx:, :]

        P_img = index_select(P, dim=-1, index=view_patch_indices)
        P_img = reshape(P_img, [B, H, A_len, 3, 256])

        L = object_grounding_loss(
            P_img,
            teacher_attn_maps["object_maps"],
            teacher_attn_maps["object_masks"],
        )
        layer_losses.append(L)

    return mean(layer_losses) if len(layer_losses) > 0 else 0.0

def masked_mean(loss, valid, eps=1e-9):
    mask = valid[:, None].float().expand_as(loss) # [B, A_len]
    denom = mask.sum() + eps
    return (loss * mask).sum() / denom

def object_grounding_loss(P_img, object_maps, object_masks, eps=1e-6, delta=1e-6):
    S = mean(P_img, dim=1)                                  # [B, A_len, 3, 256]
    M_obj = (object_maps > eps).float()                     # [B, 3, 256]
    M_view = object_masks.float()                           # [B, 3]
    M = M_obj * broadcast(M_view)                           # [B, 3, 256]

    mass = sum_over_view_patch(S * broadcast(M))            # [B, A_len]
    loss = -log(clamp_min(mass, delta))                     # [B, A_len]

    valid = (sum_over_view_patch(M) > 0).float()            # [B]
    return masked_mean(loss, valid)
\end{lstlisting}
\end{algorithm*}

\vspace{2pt}
\noindent \textbf{Depth Head.} 
To integrate depth information into cross-modal attention, we use a specialized Key-Value (KV) projector that maps depth tokens into compatible representations for selected attention heads. Algorithm~\ref{alg:depth_kv_and_attn} provides the implementation of the \texttt{DepthKVProjector} and how it is used to modify attention computation. Standard heads use key and value states from VLM backbone, while selected heads attend to projected depth tokens, supporting geometry-aware reasoning.

\begin{algorithm*}[htbp]
\caption{Python Pseudocode of Depth KV Projector and Depth Head Attention}
\label{alg:depth_kv_and_attn}
\begin{lstlisting}[
  language=Python,
  basicstyle=\ttfamily\footnotesize,
  keywordstyle=\color{kwpurple}\bfseries,
  commentstyle=\color{cmteal},
  showstringspaces=false,
  columns=fullflexible,
  keepspaces=true,
  breaklines=true,
  breakatwhitespace=true,
  xleftmargin=0.8em,
  frame=none
]
class DepthKVProjector:
    def __init__(self, kv_projector):
        self.kv_projector = kv_projector

    @property
    def heads_to_modify(self):
        return self.kv_projector.heads_to_modify

    def project_group(self, depth_tokens, g, B, T_depth, H, D):
        # depth_tokens: [B, T_depth, hidden]
        depth_tokens = rmsnorm(depth_tokens)  # [B, T_depth, hidden]

        k = self.kv_projector.k_linear[g](depth_tokens)  # [B, T_depth, H*D]
        v = self.kv_projector.v_linear[g](depth_tokens)  # [B, T_depth, H*D]

        k = k.view(B, T_depth, H, D).transpose(1, 2)     # [B, H, T_depth, D]
        v = v.view(B, T_depth, H, D).transpose(1, 2)     # [B, H, T_depth, D]

        return {
            "depth_token_k": k,
            "depth_token_v": v,
            "heads_to_modify": self.heads_to_modify,
        }

    def build(self, depth_tokens_tuple, B, T_depth, H, D):
        # depth_tokens_tuple: tuple of length G, each [B, T_depth, hidden]
        return [
            self.project_group(depth_tokens, g, B, T_depth, H, D)
            for g, depth_tokens in enumerate(depth_tokens_tuple)
        ]

    def get_cfg(self, depth_tokens_tuple, depth_group_idx, B, T_depth, H, D):
        return self.build(depth_tokens_tuple, B, T_depth, H, D)[depth_group_idx]


def depth_modified_attention(
    Q, K, V, att_mask, scaling, dropout_p,
    depth_tokens_tuple=None,
    depth_kv_projector: DepthKVProjector = None,
    depth_cfg=None,
    depth_group_idx=0,
    B=None, T_depth=None, H=None, D=None,
):
    depth_cfg = depth_kv_projector.get_cfg(
        depth_tokens_tuple,
        depth_group_idx=depth_group_idx,
        B=B, T_depth=T_depth, H=H, D=D
    )

    heads_to_modify = depth_cfg["heads_to_modify"]
    Kd = depth_cfg["depth_token_k"]                               # [B, H, T_depth, D]
    Vd = depth_cfg["depth_token_v"]                               # [B, H, T_depth, D]

    std_heads = all_heads_except(H, heads_to_modify)
    mod_heads = heads_to_modify
    out = zeros_like(Q)

    if len(std_heads) > 0:
        out[:, std_heads] = sdpa(
            Q[:, std_heads], K[:, std_heads], V[:, std_heads],
            att_mask_for(K, std_heads), scaling, dropout_p
        )

    if len(mod_heads) > 0:
        out[:, mod_heads] = sdpa(
            Q[:, mod_heads], Kd[:, mod_heads], Vd[:, mod_heads],
            None, scaling, dropout_p
        )

    return out
\end{lstlisting}
\end{algorithm*}

\vspace{2pt}
\noindent \textbf{Skill Head.} 
The Skill Head encourages semantic grounding by matching attention-derived features to a soft skill distribution target. Algorithm~\ref{alg:skill_kl_loss} describes the KL loss computation pipeline. For each transformer layer, we extract the action-query attention output, average features, and apply a classification head. The output is compared to normalized histogram targets, capturing the distribution of skill labels across the trajectory.

\begin{algorithm*}[htbp]
\caption{Python Pseudocode of Applying Skill Head KL Loss}
\label{alg:skill_kl_loss}
\begin{lstlisting}[
  language=Python,
  basicstyle=\ttfamily\small,
  keywordstyle=\color{kwpurple}\bfseries,
  commentstyle=\color{cmteal},
  showstringspaces=false,
  columns=fullflexible,
  keepspaces=true,
  breaklines=true,
  breakatwhitespace=true,
  xleftmargin=0.8em,
  frame=none
]
def skill_guidance_loss(
    all_attn_states,
    observation,               # may contain skill_soft or skill_id
    skill_head,                # linear head: [d] -> [K]
    skill_num_classes,         # K
    action_query_start_idx,
    skill_use_control: bool,
):
    target_prob = build_skill_soft_label(observation, skill_num_classes)
    if target_prob is None:
        return 0.0

    layer_feats = []
    for layer_idx, attn_data in all_attn_states:
        attn_out = select_skill_attn_out(attn_data, skill_use_control)
        skill_attn_out = attn_out[:, :, action_query_start_idx:, :]    # [B, H, A_len, d]
        feat = skill_attn_out.mean(dim=(1, 2))                         # [B, d]
        layer_feats.append(feat)

    if len(layer_feats) == 0:
        return 0.0

    feat = stack(layer_feats, dim=1).mean(dim=1)                 # [B, d]
    logits = skill_head(feat)                                    # [B, K]

    log_prob = log_softmax(logits, dim=-1)
    return kl_div_batchmean(log_prob, target_prob)


def select_skill_attn_out(attn_data, skill_use_control: bool):
    if (not skill_use_control) and ("skill_origin" in attn_data):
        return attn_data["skill_origin"]
    return attn_data["skill"]


def build_skill_soft_label(observation, K):
    if hasattr(observation, "skill_id") and observation.skill_id is not None:
        ids = observation.skill_id.long()
        if ids.ndim >= 2 and ids.shape[-1] == 1:
            ids = ids.squeeze(-1)

        if ids.ndim == 1:
            counts = one_hot(ids, K).float()                  # [B, K]
            T = 1
        else:
            ids_flat = ids.view(ids.shape[0], -1)             # [B, T]
            counts = one_hot(ids_flat, K).float().sum(dim=1)  # [B, K]
            T = ids_flat.shape[1]

        # y_k = count_k / T
        return counts / float(T)

    return None


def kl_div_batchmean(log_prob, target_prob):
    # log_prob:    [B, K]  (log softmax of student logits)
    # target_prob: [B, K]  (teacher/label distribution, sum=1)
    return kl_div(log_prob, target_prob, reduction="batchmean", log_target=False)
\end{lstlisting}
\end{algorithm*}

\vspace{2pt}
\noindent\textbf{ControlNet-style Adapter.}
To enable fine-grained control signal injection, we design a ControlNet-inspired dual-path attention mechanism. Algorithm~\ref{alg:control_attention_zero_conv} shows the implementation of \texttt{ControlAttention}, which splits the attention computation into a main path and a control-specific branch. The outputs from both branches are fused using a zero-initialized linear projection, allowing conditional modulation without disrupting pretrained behavior.

\begin{algorithm*}[htbp]
\caption{Python Pseudocode of ControlNet-style Dual-Path Control Attention with Zero-Conv Fusion}
\label{alg:control_attention_zero_conv}
\begin{lstlisting}[
  language=Python,
  basicstyle=\ttfamily\small,
  keywordstyle=\color{kwpurple}\bfseries,
  commentstyle=\color{cmteal},
  showstringspaces=false,
  columns=fullflexible,
  keepspaces=true,
  breaklines=true,
  breakatwhitespace=true,
  xleftmargin=0.8em,
  frame=none
]
class ControlAttention:
    def __init__(self, original_attn, *, hidden_size, num_control_heads, use_headwise_gate=True):
        self.origin = original_attn
        self.branch = make_control_branch(
            original_attn,
            num_control_heads=num_control_heads,
            use_headwise_gate=use_headwise_gate
        )

        self.num_heads = original_attn.config.num_attention_heads
        self.head_dim  = original_attn.head_dim

        # zero-initialized projection (ControlNet design)
        self.zero_conv = zero_init_linear(hidden_size, hidden_size)

        # optional: expand control Q heads to match origin heads
        self.has_q_expansion, self.q_expand = maybe_build_q_expansion(
            origin_heads=self.num_heads,
            control_heads=num_control_heads,
            head_dim=self.head_dim
        )

    def dual_path(self, hidden_states):
        B, T, _ = hidden_states.shape

        q0 = self.origin.q_proj(hidden_states)
        k0 = self.origin.k_proj(hidden_states)
        v0 = self.origin.v_proj(hidden_states)
        Q0, K0, V0 = reshape_to_heads(q0, k0, v0, H=self.num_heads, D=self.head_dim)  # [B,H,T,D]

        qc = self.branch.q_proj(hidden_states)  # may include extra dims for head-wise gates
        kc = self.branch.k_proj(hidden_states)
        vc = self.branch.v_proj(hidden_states)

        gate_h = None
        qc_query, qc_gate = maybe_split_query_and_gate(qc)  # qc_gate is optional
        if qc_gate is not None:
            gate_h = reshape_gate(qc_gate, H=self.num_heads)         # [B,H,T,1]
            Qc = reshape_query(qc_query, Hc=self.branch.num_heads)   # [B,Hc,T,D]
        else:
            Qc = reshape_query(qc, Hc=self.branch.num_heads)         # [B,Hc,T,D]

        Kc, Vc = reshape_to_heads(kc, vc, H=self.branch.num_heads, D=self.head_dim)  # [B,Hc,T,D]

        if self.has_q_expansion:
            Qc = expand_heads(Qc, q_expand=self.q_expand, target_H=self.num_heads)   # [B,H,T,D]

        return (Q0, K0, V0), (Qc, Kc, Vc), gate_h

    def fuse(self, origin_out, branch_out):
        # Zero Conv Fusion: y = y + ZeroConv(branch_out)
        return origin_out + self.zero_conv(branch_out)
\end{lstlisting}
\end{algorithm*}

\subsection{Implementation Details of Factor Quality Ablation}
\label{sec:factor_quality_ablation_impl}

This part provides detailed explanations corresponding to Section~\mainref{sec:head_specialization} and Section~\mainref{sec:factor_quality_analysis} of the main paper.

\noindent\textbf{Object Head.}
To assess the impact of factor quality, we require precise control over the model's grounding strength. While the standard supervision in Eq.~\mainref{eq:grounding_loss} of the main paper encourages the model to maximize attention on the object, this ablation study necessitates clamping the intensity to specific scalar values $\alpha \in \{0.25, 0.5, 0.75, 1.0\}$.We define the grounding strength $m$ as the cumulative attention mass falling within the ground-truth region mask $\mathcal{M}$:\begin{equation}m = \sum_{j \in \mathcal{M}} A_{j},\end{equation}where $A_{j}$ are the attention weights of the action token. Since the total attention is normalized, $m$ directly represents the concentration of focus on the target object. For this experiment, we replace the standard objective with a regression loss to force $m$ towards the target $\alpha$:
\begin{equation}
\mathcal{L}_{\text{ablation}}=
\begin{cases}
\dfrac{0.5\,(m-\alpha)^2}{\beta}, & \text{if } |m-\alpha|<\beta,\\[4pt]
|m-\alpha|-0.5\,\beta, & \text{otherwise}.
\end{cases}
\end{equation}
where the smoothing parameter $\beta$ is set to $0.05$. This objective allows us to strictly regulate the grounding quality for sensitivity analysis.

For the baseline model $\pi_0$ (which lacks explicit object supervision), we measure its intrinsic grounding capability using the same metric $m$. The reported value ($26.5\%$) is obtained by averaging $m$ over 200 evaluation steps during inference. This result indicates that, in the absence of targeted supervision, the model exhibits a natural tendency to attend to task-irrelevant objects, thereby motivating our design choice to introduce auxiliary guidance.

\vspace{2pt}
\noindent\textbf{Depth Head.}
Unlike the Object and Skill heads, the Depth head is enforced via architectural constraints (attention injection) rather than optimization objectives. Therefore, we cannot regulate its quality through loss scaling. Instead, we control the strength of geometric cues by modulating the signal-to-noise ratio of the input features. Let $\mathbf{f}_{\text{depth}}$ denote the feature extracted from the frozen depth encoder, and $\bm{\epsilon} \sim \mathcal{N}(\mathbf{0}, \mathbf{I})$ be a Gaussian noise vector normalized to match the statistics of the depth features. We introduce a control parameter $\delta \in [0, 1]$ (referred to as the "depth feature ratio" in the analysis) to construct the ablated feature representation $\mathbf{\tilde{f}}$:\begin{equation}\mathbf{\tilde{f}} = \delta \cdot \mathbf{f}_{\text{depth}} + (1 - \delta) \cdot \bm{\epsilon}.
\end{equation}
These corrupted features $\mathbf{\tilde{f}}$ are then projected into keys $K_{\text{Depth}}$ and values $V_{\text{Depth}}$ for the specific attention heads. By varying $\delta$ from $0$ (pure noise) to $1.0$ (clean depth signal), we quantitatively evaluate how the quality of 3D structural information impacts manipulation success.

\vspace{2pt}
\noindent\textbf{Skill Head.}
To examine the causal effect of skill recognition on task success, we regulate the model's intent classification accuracy to specific target levels $\gamma \in \{0.25, 0.5, 0.75, 1.0\}$. We define the \textit{soft accuracy} $S$ as the mean predicted probability assigned to the ground-truth skill class $y_i$ across a batch of size $N$:\begin{equation}S = \frac{1}{N} \sum_{i=1}^{N} \hat{p}_i(y_i),
\end{equation}
where $\hat{p}_i(y_i)$ denotes the probability of the correct label derived from the softmax distribution.
To enforce convergence to the target accuracy $\gamma$, we introduce an auxiliary control loss $\mathcal{L}_{\text{ctrl}}$ derived from the Smooth L1 distance:
\begin{equation}
\mathcal{L}_{\text{ctrl}} =
\begin{cases}
\dfrac{0.5\,(S-\gamma)^2}{\beta}, & \text{if } |S-\gamma|<\beta,\\[8pt]
|S-\gamma|-0.5\,\beta, & \text{otherwise}.
\end{cases}
\end{equation}
with $\beta$ set to $0.02$. The final objective for the skill head during ablation is a weighted sum: $\mathcal{L}_{\text{total}} = \mathcal{L}_{\text{skill}} + \lambda \mathcal{L}_{\text{ctrl}}$. By adjusting the weight $\lambda$, we empirically ensure the model's skill recognition performance converges to the designated target.

To measure the intrinsic skill representation capability of the baseline $\pi_0$ (reported as 48.4\%), we employ the identical skill head architecture and classification objective ($\mathcal{L}_{\text{skill}}$) as described above. Specifically, we attach the projection layer ($\bar{\mathbf{f}} \to \hat{\mathbf{p}}$) to the pre-trained $\pi_0$. Distinct from the controlled ablation models, we freeze the entire backbone and exclusively optimize the projection parameters ($\bm{W}, \mathbf{b}$) using the classification loss. This setup effectively functions as a linear probe on the fixed action features to evaluate their linear separability regarding skill semantics.

Considering the LIBERO dataset consists of 3 task-level skill categories, a purely random guess over the task-level skills would yield an accuracy of $\sim 33.3\%$. In implementation, we use four classifier outputs: the three effective skills plus one null/background class for unannotated or transition frames. Consequently, the baseline's performance of 48.4\% represents only a marginal improvement over chance. This indicates that without explicit temporal logic supervision, the representation of $\pi_0$ captures negligible high-level intent information, failing to effectively disentangle the long-horizon structure of tasks.

\subsection{Ablation on Head and Adapter Design Choices}
\label{sec:head_adapter_design_ablation}

This part provides detailed explanations corresponding to Section~\mainref{sec:design_choices_each_head} of the main paper.

\subsubsection{Object Head}
\label{sec:object-head-ablation}

To improve the interpretability and spatial alignment of attention in action decoding, we supervise a dedicated set of heads $\mathcal{H}_{\text{obj}}$ to focus on semantically meaningful regions—such as the object to be grasped or its intended destination. We investigate two strategies for supervising these attention heads: a binary mask–based region loss, and a Gaussian prior–based KL divergence loss.

\vspace{2pt}
\noindent\textbf{Object Region Supervision.}
To guide a subset of heads $\mathcal{H}_{\text{obj}}$ toward attending semantically meaningful areas such as the grasp object or destination region, we use direct supervision from binary masks $\bm{M}$ annotated by foundation models. These masks indicate which patches are considered object-relevant. Consistent with Eq.~\mainref{eq:grounding_loss}, the supervision loss is defined as a negative log likelihood over the total attention mass inside the valid object region:

\begin{equation}
\mathcal{L}_{\text{object}} = -\frac{1}{|\mathcal{T}_{a}|}\sum_{t\in\mathcal{T}_{a}}\log\left(\max\left(\sum_{p}\bar{P}_{t,p}M_p,\epsilon\right)\right).
\end{equation}

\noindent
Importantly, $\bm{M}$ only specifies which patches are object regions while leaving the distribution inside those regions unconstrained. This formulation does not penalize the model for exactly where it attends inside the object, but it does penalize insufficient attention mass inside the object region. As a result, the model learns to concentrate its attention inside the annotated object boundaries, without requiring precise spatial alignment. Empirically, this encourages more consistent and interpretable object-level grounding in attention maps.

\vspace{2pt}
\noindent\textbf{Gaussian Prior Supervision.}
As an alternative, we evaluate a softer supervision strategy by replacing the binary mask with a 2D Gaussian prior centered at the mass centroid of the annotated object region. This provides a spatial bias encouraging attention to concentrate near the most representative region of the object. Specifically, we generate a normalized Gaussian heatmap $\bm{G} \in \mathbb{R}^{3 \times 256}$ and compute the KL divergence between student attention and this distribution:

\begin{equation}
\mathcal{L}_{\text{KL}} = \frac{1}{|\mathcal{A}|} \sum_{a \in \mathcal{A}} \sum_{v=1}^{3} \sum_{p=1}^{256} \bm{G}_{v,p} \left( \log \bm{G}_{v,p} - \log \bm{A}_{a,v,p} \right).
\end{equation}

\vspace{2pt}
\noindent\textbf{Experiments and Results.} We compare these two supervision strategies on the \textit{RoboTwin 2.0} benchmark. As shown in Table~\ref{tab:robotwin_region_ablations}, using object region supervision significantly outperforms the Gaussian prior approach, achieving an average success rate of 83.33\% compared to 72.00\%. This demonstrates that direct region supervision provides a stronger and more effective learning signal for grounding attention in semantically meaningful areas, leading to better task performance.

\begin{table*}[htbp]
\centering
\caption{\textbf{Ablation Study of Object Head Design on RoboTwin 2.0 tasks.} We compare two strategies for supervising attention heads in the Object Head: region-based supervision using binary masks from foundation models, and Gaussian prior–based KL divergence. The region-based method leads to significantly higher performance (83.33\% average success rate), especially on precision-critical tasks such as \textit{Click Bell}, confirming the advantage of providing explicit spatial constraints over soft priors when grounding object-level attention.}
\normalsize
\setlength{\tabcolsep}{5pt}
\renewcommand{\arraystretch}{1.15}
\begin{tabular}{l!{\color{gray!60}\vrule width 0.6pt}ccc!{\color{gray!60}\vrule width 0.6pt}c}
\toprule
Model & Beat Hammer Block & Click Bell & Dump Bin BigBin & Avg \\
\midrule
$\pi_0$ w/ gaussian & 89\% & 40\% & 87\% & 72.00\% \\
$\pi_0$ w/ object region \textbf{(Ours)} & \textbf{94\%} & \textbf{62\%} & \textbf{94\%} & \textbf{83.33\%} \\
\bottomrule
\end{tabular}
\label{tab:robotwin_region_ablations}
\end{table*}

\subsubsection{Depth Head}
\label{sec:depth-head-ablation}

To better incorporate geometric information for 3D reasoning in manipulation tasks, we introduce a dedicated \textit{Depth Head} that leverages depth embeddings from pretrained models. Specifically, we study how different choices in depth feature extraction and processing affect performance, focusing on two aspects: (1) the backbone capacity of the depth encoder, and (2) whether to apply token downsampling to the depth tokens before fusion.

\vspace{2pt}
\noindent\textbf{Depth Encoder Variants.}
We adopt the Depth Anything 3 model as our depth encoder and evaluate three of its released variants: \texttt{small}, \texttt{base}, and \texttt{large}. These differ in parameter count and representational capacity. All variants produce dense depth tokens which are then processed by our Depth Head module.

\vspace{2pt}
\noindent\textbf{Downsampling Strategy.}
We also explore the impact of applying spatial downsampling to the output tokens of the depth encoder. The downsampling operation reduces the number of tokens before they are fed into the transformer layers of the Depth Head, potentially reducing noise and enhancing the focus on salient regions.

\vspace{2pt}
\noindent\textbf{Experiments and Results.}
As summarized in Table~\ref{tab:robotwin_depth_variants}, using the \texttt{Depth Anything 3–small} encoder achieves the best average success rate (83.00\%) across the RoboTwin 2.0 benchmark. This configuration also performs best or competitively across individual tasks. Notably, while the \texttt{large} variant attains strong performance (82.00\%), it does not surpass the smaller model despite its increased complexity, suggesting diminishing returns from larger encoders. In contrast, the \texttt{base} variant lags behind (69.67\%), indicating that encoder size alone does not guarantee better performance.

Moreover, removing token downsampling results in a notable performance drop (68.00\%), especially on fine-grained tasks like \textit{Click Bell}. This confirms that moderate spatial compression of depth tokens helps reduce redundancy and improve attention allocation in the Depth Head. Overall, these results highlight the importance of choosing a lightweight but expressive depth encoder and applying spatial abstraction to its outputs for robust depth-aware visuomotor learning.

As an orthogonality check, adding our depth head to Spatial Forcing on a cluttered stacking task improves success from 35\% to 50\%, while adding it to $\pi_0$ improves success from 25\% to 45\%, suggesting that the depth-specialized pathway is complementary to stronger spatial VLA backbones.

\begin{table*}[htbp]
\centering
\caption{\textbf{Ablation Study of Depth Head Design on RoboTwin 2.0 tasks.} We evaluate the impact of encoder capacity and token downsampling in our Depth Head. The best results are achieved using the \texttt{Depth Anything 3–small} variant (83.00\% average success rate), which balances compactness and representational power. Larger encoders offer no significant benefit and may introduce redundancy. Removing token downsampling leads to a notable drop in performance (68.00\%), especially on fine-grained tasks like \textit{Click Bell}, supporting the need for moderate spatial abstraction in depth-guided attention.}
\normalsize
\setlength{\tabcolsep}{5pt}
\renewcommand{\arraystretch}{1.15}
\begin{tabular}{l!{\color{gray!60}\vrule width 0.6pt}ccc!{\color{gray!60}\vrule width 0.6pt}c}
\toprule
Model & Beat Hammer Block & Click Bell & Dump Bin BigBin & Avg \\
\midrule
$\pi_0$ w/ small encoder \textbf{(Ours)} & \textbf{96\%} & \underline{63\%} & \textbf{87\%} & \textbf{83.00\%} \\
\midrule
$\pi_0$ w/ base encoder & 77\% & 49\% & \underline{83\%} & 69.67\% \\
$\pi_0$ w/ large encoder & \underline{92\%} & \textbf{67\%} & \textbf{87\%} & \underline{82.00\%} \\
\midrule
$\pi_0$ w/ small encoder w/o downsample & 78\% & 39\% & \textbf{87\%} & 68.00\% \\
\bottomrule
\end{tabular}
\label{tab:robotwin_depth_variants}
\end{table*}

\subsubsection{Skill Head}
\label{sec:skill-head-ablation}

To enhance the policy's ability to capture nuanced behaviors in manipulation tasks, we design dedicated \textit{skill heads} that condition the action generation on explicit skill representations. This module serves as a semantic interface between high-level task understanding and low-level action prediction. In this section, we study the effects of different label supervision strategies in training the Skill Head, focusing on: (1) the use of discrete hard labels, and (2) the alternative soft label formulation which allows richer skill supervision.

\vspace{2pt}
\noindent\textbf{Hard Label Supervision.}
A straightforward approach is to assign a one-hot \textit{hard label} to each skill instance based on expert annotation. These discrete identifiers are treated as categorical embeddings, enabling the Skill Head to specialize behavior per skill class. While this method is simple and interpretable, it may be limited in representing task ambiguity or overlapping skill boundaries.

\vspace{2pt}
\noindent\textbf{Soft Label Supervision.}
To provide a more flexible representation, we experiment with using \textit{soft labels}, where each training segment is associated with a probability distribution over multiple skill classes. These distributions are derived from stage-wise skill annotations by accumulating the skill ids that occur within the action segment, reflecting ambiguous transitions and mixed-intent windows. The Skill Head is trained to align with these soft distributions, encouraging smoother generalization and richer grounding.
For the details of constructing soft labels, please refer to Eq.~\ref{eq:soft_skill_target}.

\vspace{2pt}
\noindent\textbf{Experiments and Results.}
As shown in Table~\ref{tab:robotwin_skill_variants}, the Skill Head trained with soft label supervision achieves the best overall performance, with an average success rate of 75.00\% across three RoboTwin 2.0 tasks. It consistently outperforms the hard label variant, including on challenging tasks such as \textit{Click Bell} (39\% vs. 36\%) and \textit{Dump Bin BigBin} (94\% vs. 82\%), highlighting the benefit of using probabilistic skill distributions to capture diverse manipulation strategies. While the hard label variant performs well on more structured tasks like \textit{Beat Hammer Block} (90\%), its limited flexibility leads to lower generalization in tasks with more ambiguity.

\begin{table*}[htbp]
\centering
\caption{\textbf{Ablation Study on Skill Head Supervision.}
We compare Skill Head variants trained with hard vs. soft label supervision on RoboTwin 2.0 tasks. The soft label variant (ours) achieves the highest average performance (75.00\%), showing improved generalization on less structured tasks such as \textit{Click Bell} and \textit{Dump Bin BigBin}.}

\normalsize
\setlength{\tabcolsep}{5pt}
\renewcommand{\arraystretch}{1.15}
\begin{tabular}{l!{\color{gray!60}\vrule width 0.6pt}ccc!{\color{gray!60}\vrule width 0.6pt}c}
\toprule
Model & Beat Hammer Block & Click Bell & Dump Bin BigBin & Avg \\
\midrule
$\pi_0$ w/ hard labels & 90\% & 36\% & 82\% & 69.33\% \\
$\pi_0$ w/ soft labels \textbf{(Ours)} & \textbf{92\%} & \textbf{39\%} & \textbf{94\%} & \textbf{75.00\%} \\
\bottomrule
\end{tabular}
\label{tab:robotwin_skill_variants}
\end{table*}

\subsubsection{Feature Fusion via ControlNet-style Adapter}
\label{sec:adapter-ablation}

To integrate auxiliary features into the \textit{action decoder} without disrupting its core attention flow, we explore three fusion strategies: direct addition, gated modulation, and zero-initialized convolution (ours). These strategies modulate the main attention stream with external signals in different ways, balancing simplicity, control, and stability.

\vspace{2pt}
\noindent\textbf{Direct Addition.}
A straightforward approach is to directly add the auxiliary attention features $\mathrm{Attn}_{\text{specified}}(\mathbf{x})$ to the main attention stream $\mathrm{Attn}_{\text{main}}(\mathbf{x})$:

\begin{equation}
\mathrm{Attn}(\mathbf{x}) = \mathrm{Attn}_{\text{main}}(\mathbf{x}) + \mathrm{Attn}_{\text{specified}}(\mathbf{x}).
\end{equation}

\noindent
While simple and computationally efficient, this method lacks any adaptive control over the fused signal. It may lead to feature interference or training instability, particularly in tasks requiring fine-grained control.

\vspace{2pt}
\noindent\textbf{Gated Addition.}
To enable learnable modulation, we apply a nonlinear transformation to the auxiliary features using a $\tanh$ gate before fusion. This produces the final attention as:

\begin{equation}
\mathrm{Attn}(\mathbf{x}) = \mathrm{Attn}_{\text{main}}(\mathbf{x}) + \tanh\left(\mathrm{Attn}_{\text{specified}}(\mathbf{x})\right).
\end{equation}

\noindent
Unlike scalar gating, this formulation allows each element of the auxiliary feature map to be adaptively scaled within $(-1, 1)$, enabling smoother gradients and more expressive control. However, the unbounded nature of the fused signal can still introduce training instability in tasks requiring fine spatial precision.

\vspace{2pt}
\noindent\textbf{Zero-initialized Convolution.}
Inspired by residual modulation in ControlNet, we propose applying a zero-initialized convolutional layer to the auxiliary features before fusion. This yields the final output as:

\begin{equation}
\mathrm{Attn}(\mathbf{x}) = \mathrm{Attn}_{\text{main}}(\mathbf{x}) + \mathrm{ZeroConv}\left(\mathrm{Attn}_{\text{specified}}(\mathbf{x})\right).
\end{equation}

\noindent
The zero initialization ensures that the fused path initially behaves as an identity function, preventing early training collapse. The network can then gradually learn to utilize auxiliary cues in a stable and interpretable manner. Empirically, this design leads to more consistent performance gains across a range of tasks.

\vspace{2pt}
\noindent\textbf{Experiments and Results.}
We evaluate all three fusion methods on the \textit{RoboTwin 2.0} benchmark. As summarized in Table~\ref{tab:robotwin_feature_ablations}, our zero-initialized convolution strategy achieves the best average success rate of 83.33\%, significantly outperforming both direct addition (64.00\%) and gated addition (73.33\%). Notably, the performance on the \textit{Click Bell} task improves the most, demonstrating the benefit of stable and learnable modulation when precise spatial control is required.

\begin{table*}[htbp]
\centering
\caption{\textbf{Ablation Study of Feature Fusion Strategies on RoboTwin 2.0.}
We compare three strategies for incorporating auxiliary features into the action decoder: direct addition, elementwise $\tanh$-gated addition, and zero-initialized convolution. The zero-initialized convolution achieves the highest average success rate (83.33\%), particularly excelling in precision-sensitive tasks such as \textit{Click Bell}, highlighting the benefit of stable and learnable feature modulation.}
\normalsize
\setlength{\tabcolsep}{5pt}
\renewcommand{\arraystretch}{1.15}
\begin{tabular}{l!{\color{gray!60}\vrule width 0.6pt}ccc!{\color{gray!60}\vrule width 0.6pt}c}
\toprule
Model & Beat Hammer Block & Click Bell & Dump Bin BigBin & Avg \\
\midrule
$\pi_0$ w/ direct add & 67\% & 37\% & 88\% & 64.00\% \\
$\pi_0$ w/ gate & 87\% & 42\% & 91\% & 73.33\% \\
$\pi_0$ w/ zero conv \textbf{(Ours)} & \textbf{94\%} & \textbf{62\%} & \textbf{94\%} & \textbf{83.33\%} \\
\bottomrule
\end{tabular}
\label{tab:robotwin_feature_ablations}
\end{table*}

\subsection{Ablation on Overall Architecture}
\label{sec:overall_architecture_ablation}

This part provides detailed explanations corresponding to Section~\mainref{sec:design_choices_each_head} of the main paper.

\noindent\textbf{Guidance Layers.}
We study how the choice of guidance layers affects robustness by applying guidance to different subsets of transformer layers in $\pi_0$. Specifically, we compare guiding all layers with guiding only one of four layer quartiles, where layers are evenly divided from bottom to top. All settings use the same training protocol and evaluation benchmarks.
Table~\ref{tab:libero_plus_full_results_layer_quartiles} reports performance on LIBERO-Plus benchmark only the third quartile of layers achieves the best overall performance, with a total score of 75.4, outperforming guidance on all layers (74.1) as well as the first, second, and fourth quartiles (74.4, 74.3, and 73.8 respectively). This trend is consistent across multiple task categories and variation types.
These results suggest that guidance is most effective when applied to mid-to-upper layers, which likely capture higher-level semantic and task-relevant representations. In contrast, guiding all layers or very early/late layers may dilute the effect of guidance or interfere with low-level feature learning.

\begin{table*}[htbp]
\centering
\normalsize
\setlength{\tabcolsep}{6pt}
\renewcommand{\arraystretch}{1.12}
\caption{\textbf{Ablation Study of Guidance Layer Subsets on LIBERO-Plus.}
We evaluate guidance on all layers and on four layer quartiles. The third quartile achieves the highest total score (75.4), higher than guiding on all layers (74.1) and the other quartiles (74.4/74.3/73.8), indicating that focusing guidance on a specific layer range improves robustness.}
\begin{tabular}{lcccccccc}
\toprule
 & Camera & Robot & Language & Light & Background & Noise & Layout & Total \\
\midrule

\rowcolor{gray!8}
\multicolumn{9}{c}{\rule{0pt}{2.6ex}\textbf{$\pi_0$ guided on all layers}\rule{0pt}{2.2ex}} \\
\midrule
Spatial & 81.6 & 61.7 & 66.2 & 96.2 & 95.7 & 91.7 & 92.7 & 82.7 \\
Object  & 79.5 & 43.0 & 81.6 & 96.0 & 96.0 & 91.9 & 74.7 & 78.9 \\
Goal    & 71.6 & 45.2 & 42.2 & 95.0 & 85.8 & 83.4 & 66.1 & 67.7 \\
Long    & 51.6 & 43.5 & 65.3 & 83.9 & 76.8 & 80.8 & 79.8 & 67.5 \\
Avg     & 70.7 & 47.9 & 63.1 & 92.9 & 88.1 & 86.7 & 77.9 & 74.1 \\
\midrule

\rowcolor{gray!8}
\multicolumn{9}{c}{\rule{0pt}{2.6ex}\textbf{$\pi_0$ guided on first quartile of layers}\rule{0pt}{2.2ex}} \\
\midrule
Spatial & 83.5 & 59.1 & 63.1 & 97.3 & 95.0 & 92.6 & 91.4 & 82.1 \\
Object  & 85.1 & 47.0 & 85.6 & 97.0 & 91.5 & 96.2 & 73.2 & 81.1 \\
Goal    & 72.1 & 44.7 & 40.0 & 92.1 & 87.9 & 83.4 & 61.6 & 66.5 \\
Long    & 50.4 & 45.5 & 64.2 & 88.7 & 74.4 & 82.2 & 83.7 & 68.5 \\
Avg     & 72.3 & 48.8 & 62.4 & 93.9 & 86.8 & 88.5 & 76.7 & 74.4 \\
\midrule

\rowcolor{gray!8}
\multicolumn{9}{c}{\rule{0pt}{2.6ex}\textbf{$\pi_0$ guided on second quartile of layers}\rule{0pt}{2.2ex}} \\
\midrule
Spatial & 83.0 & 58.9 & 64.9 & 96.6 & 95.3 & 91.2 & 93.5 & 82.4 \\
Object  & 80.3 & 45.5 & 83.3 & 94.3 & 94.0 & 94.8 & 74.4 & 79.7 \\
Goal    & 72.3 & 44.0 & 39.0 & 93.9 & 89.7 & 87.1 & 65.4 & 67.8 \\
Long    & 55.8 & 41.2 & 64.8 & 86.9 & 78.9 & 73.1 & 86.2 & 67.8 \\
Avg     & 72.5 & 47.0 & 62.2 & 93.0 & 89.1 & 86.1 & 79.1 & 74.3 \\
\midrule

\rowcolor{gray!8}
\multicolumn{9}{c}{\rule{0pt}{2.6ex}\textbf{$\pi_0$ guided on third quartile of layers}\rule{0pt}{2.2ex}} \\
\midrule
Spatial & 86.4 & 60.6 & 65.9 & 99.3 & 95.7 & 92.3 & 94.5 & 84.0 \\
Object  & 86.6 & 52.0 & 77.1 & 94.3 & 96.8 & 92.4 & 75.2 & 80.9 \\
Goal    & 75.7 & 50.6 & 42.4 & 96.8 & 92.9 & 85.2 & 68.5 & 70.8 \\
Long    & 48.2 & 43.5 & 67.4 & 87.6 & 72.7 & 72.8 & 83.3 & 66.2 \\
Avg     & 73.7 & 51.4 & 62.6 & 94.6 & 89.0 & 85.2 & 79.9 & 75.4 \\
\midrule

\rowcolor{gray!8}
\multicolumn{9}{c}{\rule{0pt}{2.6ex}\textbf{$\pi_0$ guided on fourth quartile of layers}\rule{0pt}{2.2ex}} \\
\midrule
Spatial & 83.8 & 54.3 & 65.4 & 97.6 & 94.6 & 89.2 & 93.2 & 81.7 \\
Object  & 83.8 & 44.5 & 79.1 & 93.9 & 92.7 & 90.5 & 75.9 & 78.9 \\
Goal    & 71.8 & 47.7 & 43.9 & 92.5 & 86.5 & 83.6 & 66.1 & 68.2 \\
Long    & 45.8 & 44.0 & 64.5 & 86.1 & 76.1 & 78.1 & 86.2 & 67.0 \\
Avg     & 70.8 & 47.4 & 62.6 & 92.6 & 87.1 & 85.1 & 79.6 & 73.8 \\

\bottomrule
\end{tabular}
\label{tab:libero_plus_full_results_layer_quartiles}
\end{table*}


\subsection{Complete Model Architecture of \alias}
\label{sec:complete_model_architecture}

We provide the complete model architecture of \alias in Table~\ref{tab:pi0_model2_arch}, detailing every module used in both perception and control pathways. The model integrates a SigLIP vision tower for multi-view visual encoding, a PaliGemma language backbone for multimodal grounding, and a lightweight Gemma-based expert head for action prediction. Optional branches such as the Depth Head, Skill Head, and ControlAttention modules can be toggled depending on the task setup, enabling flexible scaling and specialization. This architecture supports the strong performance of \alias on diverse visuomotor benchmarks, including RoboTwin 2.0, by unifying visual, linguistic, and temporal modalities within a compact yet expressive framework.

\begin{table*}[t]
\centering
\small
\setlength{\tabcolsep}{4pt}
\caption{\textbf{Model Architecture of \alias.} This table lists the detailed layer-wise composition of our visuomotor agent, including the vision encoder, language backbone, action decoder, and optional modules such as the Depth Head, Skill Head, and ControlAttention layers. Our design uses a multi-view SigLIP transformer for image encoding, a PaliGemma (Gemma-2B) backbone for multimodal reasoning, and a compact Gemma-300M expert for action prediction. The modular architecture allows for easy integration of spatial and semantic grounding signals, contributing to the strong results achieved by \alias across manipulation tasks.}
\label{tab:pi0_model2_arch}
\begin{tabular}{l l c l l}
\toprule
Module & Layer Type & Layer Num & Input Shape & Output Shape \\
\midrule
\multicolumn{5}{c}{SigLIP Vision Tower (per view, V=3)} \\
\midrule
SiglipVisionTransformer & SiglipVisionEmbeddings & 1 & $(B,3,224,224)$ & $(B,256,768)$ \\
 & SiglipEncoderLayer & 12 & $(B,256,768)$ & $(B,256,768)$ \\
 & LayerNorm & 1 & $(B,256,768)$ & $(B,256,768)$ \\
\midrule
\multicolumn{5}{c}{PaliGemma Multi-Modal Projector (per view)} \\
MultiModalProjector & Linear & 1 & $(B,256,768)$ & $(B,256,2048)$ \\
\midrule
\multicolumn{5}{c}{PaliGemma Language Model (Gemma-2B)} \\
\midrule
Embed Tokens & Embedding & 1 & $(B,48)$ & $(B,48,2048)$ \\
Language Transformer & GemmaDecoderLayer & 18 & $(B,816,2048)$ & $(B,816,2048)$ \\
Norm & RMSNorm & 1 & $(B,816,2048)$ & $(B,816,2048)$ \\
\midrule
\multicolumn{5}{c}{Action/State/Time Embedding} \\
\midrule
State Projection & Linear & 1 & $(B,32)$ & $(B,1,1024)$ \\
Action In Projection & Linear & 1 & $(B,50,32)$ & $(B,50,1024)$ \\
Time Embedding & Sin/Cos & 1 & $(B)$ & $(B,1024)$ \\
Action-Time MLP & Linear + SiLU + Linear & 1 & $(B,50,2048)$ & $(B,50,1024)$ \\
\midrule
\multicolumn{5}{c}{Gemma Action Expert (Gemma-300M)} \\
\midrule
Expert Transformer & GemmaDecoderLayer & 18 & $(B,51,1024)$ & $(B,51,1024)$ \\
Norm & RMSNorm & 1 & $(B,51,1024)$ & $(B,51,1024)$ \\
Action Out Projection & Linear & 1 & $(B,50,1024)$ & $(B,50,32)$ \\
\midrule
\multicolumn{5}{c}{Depth Branch (Optional)} \\
\midrule
DepthEncoder (primary view) & DepthAnything + TokenMerging2D & 1 & $(B,3,224,224)$ & $4\times(B,16,1024)$ \\
DepthTokenKVProjector & Linear (K/V) & 4 & $(B,16,1024)$ & $(B,H,16,d_h)$ (K/V) \\
\midrule
\multicolumn{5}{c}{Skill Head (Optional)} \\
\midrule
Skill Head & Linear & 1 & $(B,256)$ & $(B,K)$ \\
\midrule
\multicolumn{5}{c}{ControlAttention (Optional, ControlNet-style Adapter)} \\
\midrule
Expert Self-Attn & ControlAwareAttention & 18 & $(B,51,1024)$ & $(B,51,1024)$ \\
\midrule
\end{tabular}
\end{table*}

\subsection{Code Details}

Our code and dataset will be open-sourced after acceptance.

We developed \alias based on the codebases of \textit{openpi} (the official release of the $\pi_0$ model) and \textit{RoboTwin 2.0}.

During development, we identified several limitations in these two codebases:

\vspace{2pt}
\noindent\textbf{Data Format Conversion.}
Both the official LIBERO dataset from \textit{openpi} and our custom-collected RoboTwin 2.0 dataset were originally stored in the LeRobot 2.0 format, which suffers from a critical data loading bottleneck. LeRobot 3.0 resolves this issue with improved I/O efficiency. To enable faster training and evaluation, we therefore converted both public and private datasets into the LeRobot 3.0 format.

\vspace{2pt}
\noindent\textbf{Training Speed Optimization.}
The default PyTorch training pipeline provided by \textit{openpi} is significantly slower than its JAX counterpart. To address this, we applied \texttt{torch.compile} to wrap the model, which led to a noticeable speedup in training efficiency without impacting performance.

\vspace{2pt}
\noindent\textbf{Framework and Precision Sensitivity.}
When training with full \texttt{float32} precision, we observe that the model $\pi_0$ achieves equivalent performance (90) across both JAX and PyTorch implementations, suggesting that the choice of train/test framework is not a limiting factor. However, when switching to full \texttt{bfloat16} training precision, performance degrades significantly (e.g., down to 10) in our setting. This issue is eliminated by using either full \texttt{float32} or mixed precision training. We therefore adopt mixed precision by default, which provides a good balance between speed and stability while matching full \texttt{float32} performance. Details are reported in Table~\ref{tab:pi0_precision_framework_ablation}.

\vspace{2pt}
\noindent\textbf{Batch Size and Gradient Accumulation.}
To increase the effective batch size, we implemented gradient accumulation in the training loop. However, this modification did not lead to meaningful performance improvements and in some cases slightly degraded the results. As such, gradient accumulation is disabled by default in our final setup.

\vspace{2pt}
\noindent\textbf{Validation Strategy.}
The official codebase does not include a validation set or evaluation pipeline during training. However, we find that monitoring the convergence of auxiliary objectives—such as object grounding loss and skill prediction loss—is critical to ensuring effective learning. We thus split each dataset into training and validation subsets using a 93:7 ratio, and incorporated open-loop validation loss tracking throughout training. This allows us to verify that auxiliary heads are making meaningful progress, even in the absence of closed-loop task rollouts.

\begin{table*}[t]
\centering
\normalsize
\caption{\textbf{Precision and framework ablation for $\pi_0$.} Performance remains stable (90) across training/testing frameworks (JAX vs.\ Torch) when using full \texttt{float32} precision. In contrast, full \texttt{bfloat16} training leads to a significant drop (10), consistent with LIBERO-Plus reproducibility issues. Mixed-precision training serves as an efficient alternative, achieving the same performance as full \texttt{float32}.}
\label{tab:pi0_precision_framework_ablation}
\begin{tabular}{lllllr}
\toprule
Model & Pretrain Ckpt Precision & Train Framework & Test Framework & Training Precision Policy & Performance \\
\midrule
$\pi_0$ & float32  & JAX   & JAX   & float32 (full)          & 90 \\
$\pi_0$ & float32  & JAX   & Torch & float32 (full)          & 90 \\
$\pi_0$ & bfloat16 & Torch & Torch & bfloat16 (full)         & 10 \\
$\pi_0$ & float32  & Torch & Torch & float32 (full)          & 90 \\
\midrule
$\pi_0$ & float32  & Torch & Torch & mixed precision         & 90 \\
\bottomrule
\end{tabular}
\end{table*}

\subsection{Dataset Construction}
\label{sec:dataset}

This section provides implementation details for the factor annotation pipeline summarized in Fig.~\mainref{fig:scale_pipeline} and Section~\mainref{sec:guidance_dataset_main_paper} of the main paper.

\subsubsection{\textbf{Object Masks}}
\label{sec:object_mask_construction}

To provide the spatial targets required by Eq.~\mainref{eq:grounding_loss}, we construct \emph{stage-aware} object masks via a semi-automatic, human-in-the-loop pipeline.
Each episode is first partitioned into a sequence of temporal stages, where each stage corresponds to a specific task-relevant object.

We automate the initialization process using Qwen3-VL~\cite{bai2025qwen3vltechnicalreport}.
For the start frame of each stage, we query Qwen3-VL with the stage description to detect the target object and generate candidate foreground point prompts.
Given these VLM-proposed points, we invoke the video tracking capability of SAM2~\cite{ravi2025sam} to propagate the object mask across frames within the stage interval.To ensure high-quality supervision, we implement a final human verification step.
This hybrid workflow combines the efficiency of VLM-based auto-labeling with the precision of human oversight, yielding supervision that is both \emph{temporally localized} and \emph{object specific}.

For training, we convert each per-frame binary mask to patch-level targets aligned with the $16\times16$ image-token grid.
Specifically, for each patch $p\in\mathcal{P}$ we average pool the mask pixels inside the patch to obtain a foreground-coverage score $s_p\in[0,1]$, and then threshold it to obtain a binary patch indicator:

\begin{equation}
m_p \;=\; \mathbb{I}[\,s_p \ge \tau\,],\qquad p\in\mathcal{P}.
\end{equation}

Frames outside any annotated stage interval are treated as unlabeled for object supervision.
In addition, if the propagated mask is empty for a given view/frame (equivalently, $\sum_{p\in\mathcal{P}} m_p = 0$, typically because the stage-specific object is not visible), we also mark that view/frame as unlabeled and exclude it from Eq.~\mainref{eq:grounding_loss}.

\subsubsection{\textbf{Skill Labels}}
\label{sec:skill_label_construction}
To support the semantic intent objective in the Skill Head (Eq.~\mainref{eq:skill_loss}), we derive a \emph{soft} target distribution from the stage-wise skill annotations.
Each stage is assigned a discrete skill identifier, and the stage label is applied to all timesteps within its interval.
Given a segment of $T$ timesteps with skill ids $\{s_t\}_{t=1}^{T}$, we compute a histogram over $K$ classes and normalize it into a probability vector $\mathbf{y}\in\mathbb{R}^K$:

\begin{equation}
\label{eq:soft_skill_target}
y_k \;=\; \frac{\sum_{t=1}^{T}\mathbb{I}[s_t = k]}{\sum_{j=0}^{K-1}\sum_{t=1}^{T}\mathbb{I}[s_t = j]},
\qquad k=0,\ldots,K-1,
\end{equation}

When only one class appears in the segment, $\mathbf{y}$ reduces to a one-hot target; when multiple skills occur, $\mathbf{y}$ reflects their relative prevalence.
For LIBERO, $K=4$: three task-level skill categories plus one null/background class for unannotated or transition frames.
This construction is directly matched to the KL-divergence loss in Eq.~\mainref{eq:skill_loss}, providing stable supervision that encourages each designated skill head to encode trajectory-level intent rather than purely step-wise cues.


\subsection{Experiment Details}\label{sec:experiment_details}

\vspace{2pt}
\noindent\textbf{LIBERO-Plus.} 
To evaluate robustness, we train all models solely on the official LIBERO dataset, annotated via the pipeline described in Section~\ref{sec:dataset}. Evaluation is performed zero-shot on the full \textit{LIBERO-Plus} benchmark to assess generalization. We adopt a two-stage training strategy: (1) we first fine-tune the pretrained $\pi_0$ on the \textit{LIBERO} dataset without any guidance; (2) we then continue training with auxiliary guidance using object and skill losses, with \texttt{object\_loss\_weight}=0.001 and \texttt{skill\_loss\_weight}=0.001.

For both stages, we use the AdamW optimizer with a cosine learning rate schedule: 1,000 warmup steps, peak learning rate $2.5\times10^{-5}$, decaying to $2.5\times10^{-6}$. Training uses a global batch size of 64 and an action chunk size of 50 across 8 NVIDIA H200 GPUs, with mixed precision enabled. Evaluation is conducted on a single NVIDIA RTX 4090 GPU, with the VLM backbone in \texttt{bfloat16} precision and the action decoder in \texttt{float32}. All models, including the $\pi_0$ baseline, are trained and evaluated under identical settings for fair comparison. For ablations on training hyperparameters, refer to Section~\ref{sec:training_hyperparameter_ablation}.

\vspace{2pt}
\noindent\textbf{RoboTwin 2.0.} 
We evaluate on 8 representative tasks from \textit{RoboTwin 2.0}: \textit{Adjust Bottle}, \textit{Beat Hammer Block}, \textit{Click Bell}, \textit{Dump Bin BigBin}, \textit{Moving PlayingCard away}, \textit{Lift Pot}, \textit{Place Burger Fries}, and \textit{Place Can Basket}. For each task, we collect 1,000 demonstration trajectories in randomized environments. Training follows the same optimizer and precision setup as above, with \texttt{object\_loss\_weight}=0.01 and \texttt{skill\_loss\_weight}=0.01 for auxiliary supervision. Each model is trained for 30k steps using 4 NVIDIA H200 GPUs, with a global batch size of 16 and an action chunk size of 50. Evaluation is done on a single RTX 4090 GPU using the same precision settings. Success rates are computed over 100 rollouts per task.

\vspace{2pt}
\noindent\textbf{Real-World.} 
We consider six real-world tasks, each with approximately 50 human demonstration episodes. Each episode is automatically annotated with object, skill, and geometry signals using our developed labeling tool. Models are trained on two NVIDIA H200 GPUs and evaluated on an RTX 4090. We follow standard training and inference procedures to ensure a fair and reproducible comparison.


\subsection{Training Hyperparameter Ablation.}
\label{sec:training_hyperparameter_ablation}
We conduct an ablation study on auxiliary supervision weights while keeping the training budget fixed at 30k steps and a global batch size of 64. As shown in Table \ref{tab:hparam_ablation_groups_nosetting_nodepth}, our final configuration uses balanced low weights, with $w_{\mathrm{obj}} = 0.001$ and $w_{\mathrm{skill}} = 0.001$, and achieves the best three-track average of 87.83.

Using stronger auxiliary weights, such as $(w_{\mathrm{obj}}, w_{\mathrm{skill}}) = (0.01, 0.01)$, lowers the three-track average to 85.77, suggesting that overly strong auxiliary objectives can interfere with the action-generation objective. Asymmetric settings are also less effective overall: lowering only the skill weight or only the object weight reaches 86.12 and 85.22, respectively. Further reducing the skill weight to 0.0003 improves the Background track but still underperforms the balanced low-weight setting overall.

\begin{table*}[htbp]
\centering
\small
\setlength{\tabcolsep}{5pt}
\renewcommand{\arraystretch}{1.15}
\caption{
\textbf{Ablation study on auxiliary loss weights.} All runs use 30k training steps and a global batch size of 64; only the object and skill supervision weights vary. When a configuration does not explicitly vary one of the weights, that weight is set to 0.01. We report the Light, Background, and Layout tracks, and Avg denotes their arithmetic mean. Our final setting uses balanced low weights $(w_{\mathrm{obj}}, w_{\mathrm{skill}}) = (0.001, 0.001)$ and achieves the best average score.
}
\label{tab:hparam_ablation_groups_nosetting_nodepth}
\begin{tabular}{lcc!{\color{gray!60}\vrule width 0.6pt}rrrr}
\toprule
Setting &
$w_{\mathrm{obj}}$ &
$w_{\mathrm{skill}}$ &
Light &
Background &
Layout &
Avg \\
\midrule
\rowcolor{gray!8}
Final (Ours) & 0.001 & 0.001 & \textbf{94.60} & 89.00 & \textbf{79.90} & \textbf{87.83} \\
High obj. / high skill & 0.01 & 0.01 & 92.05 & 88.07 & 77.19 & 85.77 \\
High obj. / low skill & 0.01 & 0.001 & 92.46 & 89.42 & 76.48 & 86.12 \\
Low obj. / high skill & 0.001 & 0.01 & 91.51 & 88.56 & 75.59 & 85.22 \\
Low obj. / lower skill & 0.001 & 0.0003 & 91.17 & \textbf{89.96} & 76.00 & 85.71 \\
\bottomrule
\end{tabular}
\end{table*}

\subsection{Real-World Experiments: Deployment \& Evaluation}
\label{sec:real_world_deployment_eval}

\subsubsection{Real-world Deployment Setup}
\label{sec:real_deploy}
We deploy GuidedVLA for real-world inference on a single NVIDIA RTX 4090 GPU.
At each inference cycle, given RGB observations from the robot-mounted cameras, GuidedVLA outputs a \textbf{50-step} action chunk.
The chunk is parameterized at an effective rate of \textbf{20\,Hz}.
On the client side, an executor upsamples the 20\,Hz keyframes via linear interpolation to produce smooth trajectories, and streams commands to the low-level controller at a \textbf{50\,Hz} control rate.
To ensure stable execution, after publishing each chunk we wait for joint-position convergence with a \textbf{4.0\,s} timeout; if the timeout is reached, we proceed to the next inference cycle.
Camera extrinsics are set to match the training distribution.
The third-person camera is mounted at an elevation angle of approximately \textbf{45$^\circ$}, at roughly \textbf{60\,cm} above the workspace center (Figure~\mainref{fig:real_setting}).
All test objects are placed within a \textbf{50\,cm $\times$ 60\,cm $\times$ 30\,cm} workspace in front of the robot.
During deployment, depth-aware inference is performed by a \textbf{frozen Depth Anything V3} encoder (small variant) integrated into the model.
Its depth features are spatially downsampled to match the token resolution, and then injected into a dedicated \textbf{Depth Head} within the attention pathway.

\subsubsection{Detailed Evaluation Setup}
\label{sec:real_eval}

\textbf{Task Definitions \& Success Criteria:}
\label{sec:real_tasks}
We evaluate three household manipulation tasks on the ALOHA AgileX dual-arm platform (T1--T3) and three laboratory manipulation tasks on the PSI-Bot dual-arm platform (T4--T6).
Each evaluation trial lasts for at most \textbf{120\,s} and terminates early once the success condition is met.
Unless otherwise specified, we require a \textbf{1\,s dwell time}: the relevant objects must remain stable in the target configuration for at least 1\,s without human intervention.

\textbf{ALOHA household tasks:}

\textbf{(T1) Pick up fruits and vegetables.}
The robot must place the green pepper and carrot onto the plate, and place the strawberry into the bowl.
A trial is successful if the pepper and carrot are both inside the plate region and the strawberry is inside the bowl region.

\textbf{(T2) Stack bowls and place on the first shelf.}
The robot stacks two bowls and places the stacked bowls onto the first shelf.
A trial is successful if the bowls form a stable stacked configuration and the stack is placed within the designated shelf region.

\textbf{(T3) Clean the tabletop (sweep $\rightarrow$ dustpan $\rightarrow$ pour $\rightarrow$ return).}
The robot sweeps trash into the dustpan with a broom, pours the trash from the dustpan into the tray, and returns both the broom and dustpan back to the table.
A trial is successful if the robot completes the pouring action over the tray and returns the tools to the table.

\textbf{PSI-Bot laboratory tasks:}

\textbf{(T4) Place beaker in heating mantle.}
The robot grasps a beaker and inserts it into the heating mantle.
A trial is successful if the beaker bottom is seated inside the mantle opening (i.e., inserted into the cavity).

\textbf{(T5) Stack small beakers inside a large beaker.}
The robot places small beakers into a large beaker.
A trial is successful if the small beakers are contained within the large beaker.

\textbf{(T6) Heat the beaker (place the asbestos mesh, then place the beaker on it).}
The robot first places the asbestos mesh on the lower level of the iron stand, and then places the beaker on top of the mesh.
A trial is successful if the mesh is placed on the designated lower support ring and the beaker is stably placed on the mesh.

\subsection{Real-World Generalization Settings}
\label{sec:real_generalization}

We evaluate three real-robot generalization regimes: \textbf{in-domain (positional)}, \textbf{scene}, and \textbf{lighting}.
Each trial is first reset to a canonical task layout and then randomized according to \emph{exactly one} regime; we do not combine multiple shifts within a single trial.

\paragraph{In-domain (positional) generalization.}
We perturb the initial object placement within the training distribution by sampling from a $3\times3$ grid of 9 discrete anchors centered at the nominal pose.
Adjacent anchors are spaced by 1--2\,cm (approximately within $\pm$2\,cm per axis relative to the nominal position), while keeping task semantics unchanged.

\paragraph{Scene generalization.}
We introduce clutter by adding 3--5 distractor objects per trial, sampled from the same domain as the task (household items for ALOHA tasks, lab items for PSI-Bot tasks).
Distractors are placed to avoid occluding target objects and to keep the nominal manipulation corridor feasible, thereby inducing appearance/context shifts without altering the intended task.

\paragraph{Lighting generalization.}
We change illumination using colored decorative lighting with three color settings.
Lighting is kept constant within each trial and constrained not to render target objects visually ambiguous, inducing appearance shifts while preserving task observability.

\begin{figure}
  \centering
  \includegraphics[width=\linewidth]{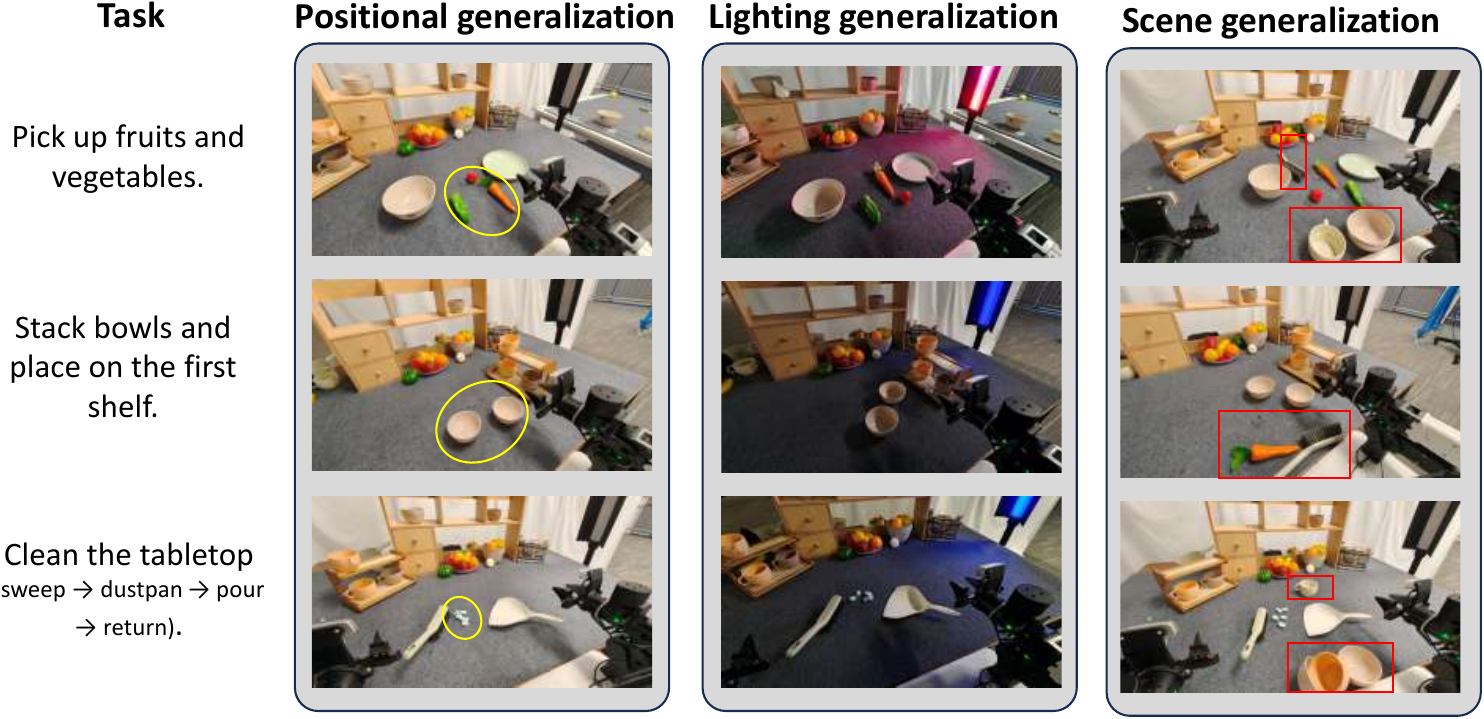}
  \caption{\textbf{ALOHA real-world generalization settings (T1--T3).}
  From left to right: \textbf{in-domain (positional)} perturbations using a $3\times3$ anchor grid, \textbf{lighting} shifts with colored illumination, and \textbf{scene} shifts by adding distractor objects.}
  \label{fig:real_gen_aloha}
\end{figure}

\begin{figure}
  \centering
  \includegraphics[width=\linewidth]{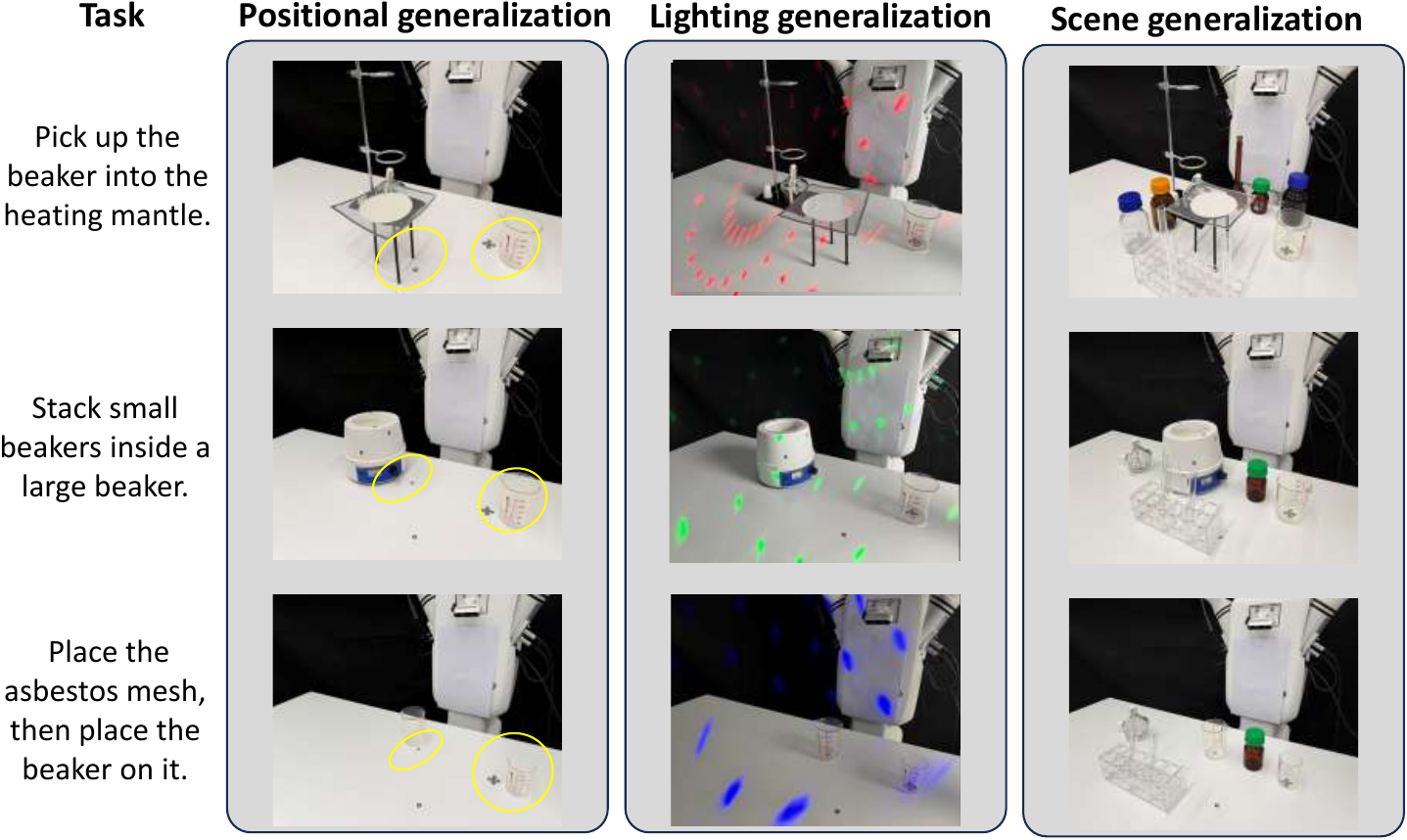}
  \caption{\textbf{PSI-Bot real-world generalization settings (T4--T6).}
  From left to right: \textbf{in-domain (positional)} perturbations using a $3\times3$ anchor grid, \textbf{lighting} shifts with colored illumination, and \textbf{scene} shifts by adding distractor objects.}
  \label{fig:real_gen_psibot}
\end{figure}

\subsection{Visualization of LIBERO-Plus Results}
\label{sec:libero_plus_visualization}

To complement the quantitative success rates on simulation-based benchmarks in Table~\ref{tab:libero_plus_full_results}, we visualize representative \emph{successful} policy rollouts from the LIBERO-Plus benchmark, covering its four task suites: \textbf{spatial}, \textbf{object}, \textbf{goal}, and \textbf{long}.
Each visualization shows a sequence of 7 keyframes sampled from a successful episode, covering the stages of approach, interaction, and completion. From top to bottom, the rows visualize: RGB image, object head attention, predicted depth map, and depth head attention.

\subsubsection{Object attention is stage-aware}
The object-specialized head dynamically shifts its attention across the episode.
In early stages, the attention is concentrated on the object of interest (e.g., a bowl or a book), enabling precise targeting for grasping. As the robot transitions toward goal states, the attention gradually shifts to the target container or placement area, clearly demonstrating a \textit{stage-wise} semantic understanding of task progression.

\subsubsection{Depth attention captures geometric awareness}
The depth head consistently highlights meaningful spatial regions, including both the robot arm and task-relevant objects.
This behavior enables \textit{geometry-aware reasoning}, particularly useful for tasks involving occlusion, stacking, or motion planning.
For example, in stacking tasks or when reaching into a container, the depth attention map exhibits strong focus on object contours and their relative positioning, helping the policy plan precise and feasible motions.

\begin{figure*}[htbp]
    \centering
    \includegraphics[width=\textwidth]{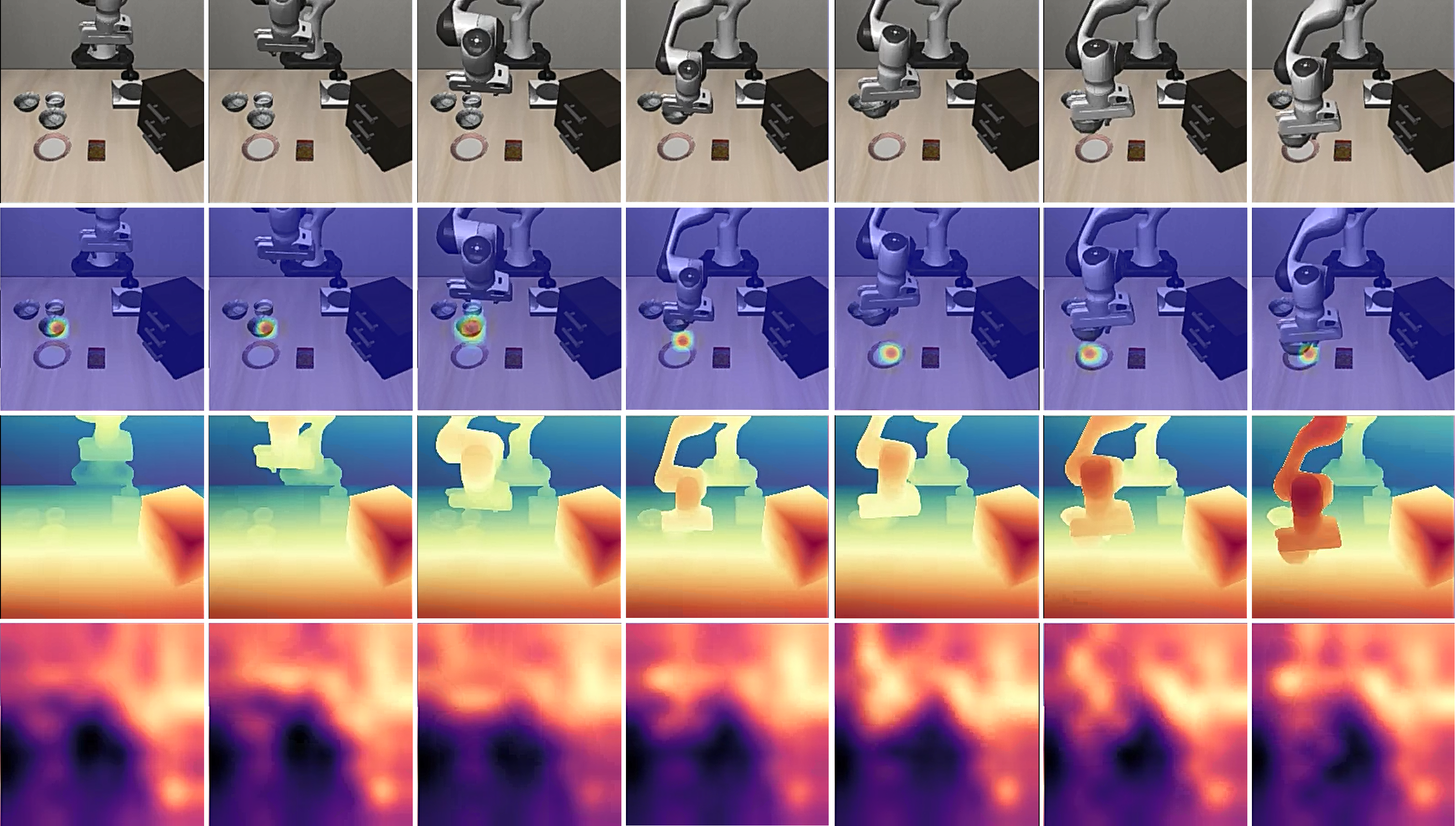}
    \caption{
        \textbf{LIBERO-Plus rollout visualization (spatial task suite of LIBERO-Plus).}
        Each column corresponds to one stage in the whole episode, with 7 stages in total. First row shows the original RGB observations during the rollout. Second row visualizes the attention maps from \alias’s object head. Third row presents the depth information encoded by the depth encoder, and fourth row illustrates the corresponding attention maps produced by \alias’s depth head based on the depth features in the third row.
    }
    \label{fig:attention_visualization_spatial}
\end{figure*}

\begin{figure*}[htbp]
    \centering
    \includegraphics[width=\textwidth]{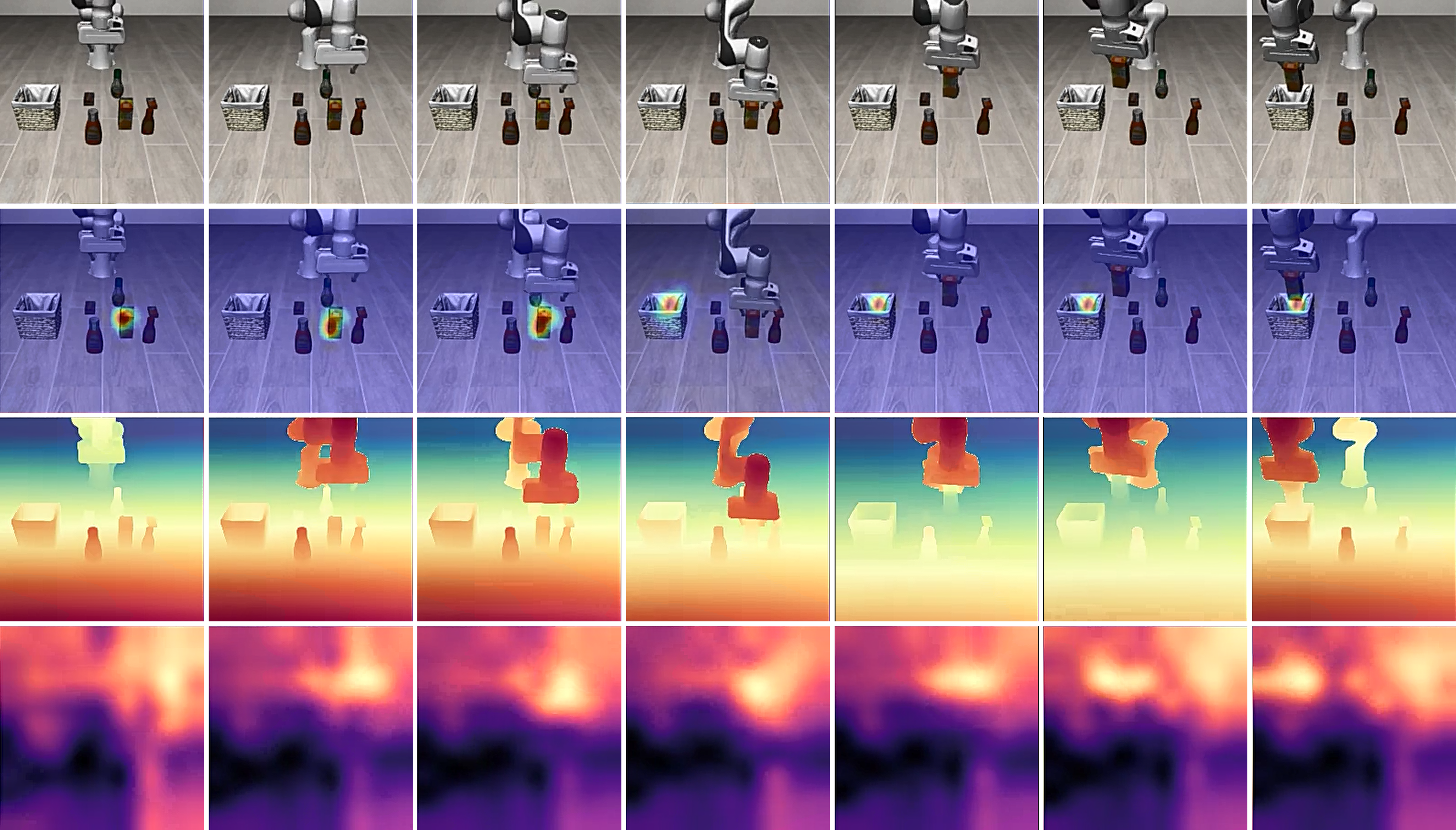}
    \caption{
        \textbf{LIBERO-Plus rollout visualization (object task suite of LIBERO-Plus).}
        Each column corresponds to one stage in the whole episode, with 7 stages in total.
    }
    \label{fig:attention_visualization}
\end{figure*}

\begin{figure*}[htbp]
    \centering
    \includegraphics[width=\textwidth]{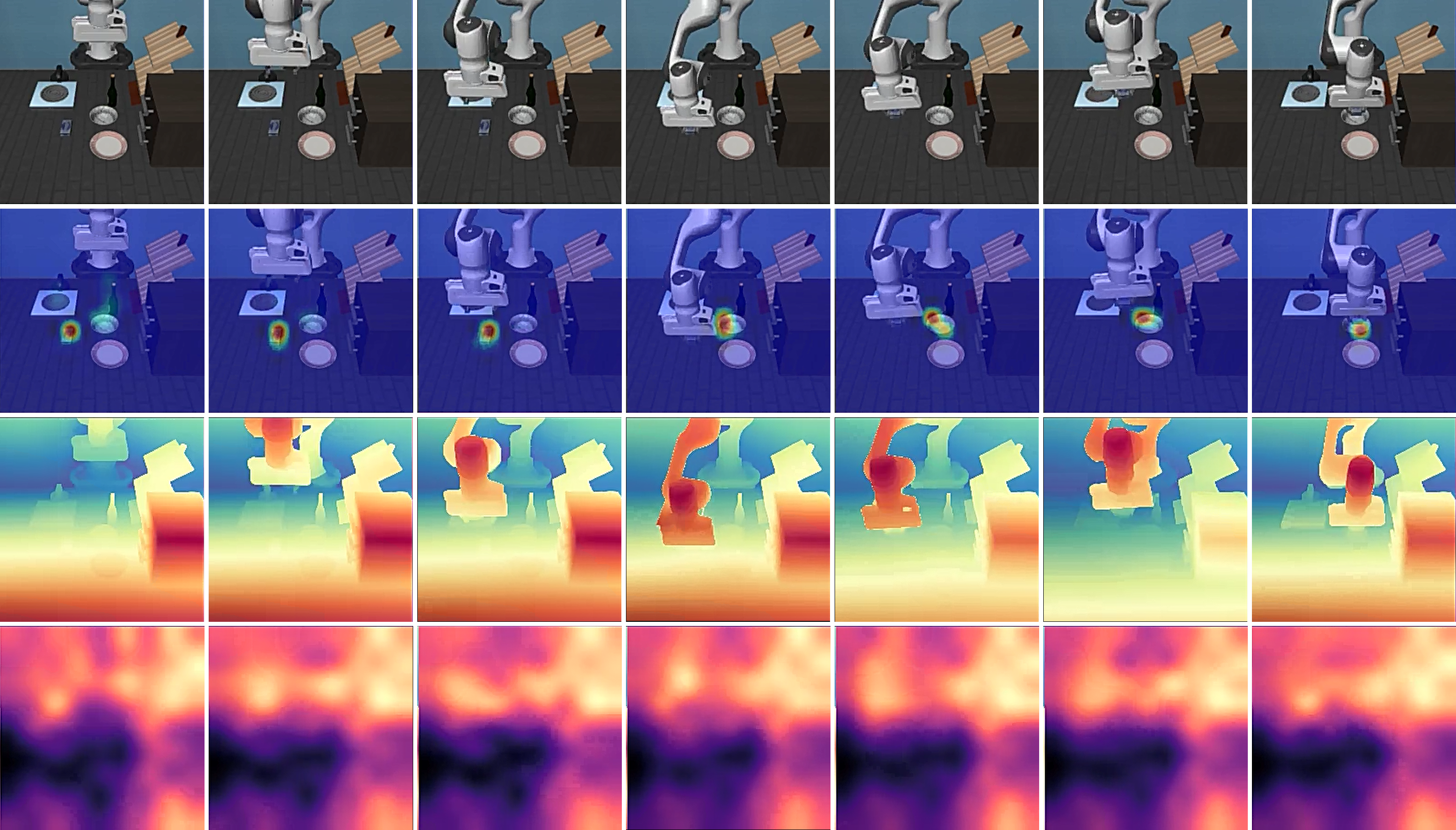}
    \caption{
        \textbf{LIBERO-Plus rollout visualization (goal task suite of LIBERO-Plus).}
        Each column corresponds to one stage in the whole episode, with 7 stages in total.
    }
    \label{fig:attention_visualization_goal}
\end{figure*}

\begin{figure*}[htbp]
    \centering
    \includegraphics[width=\textwidth]{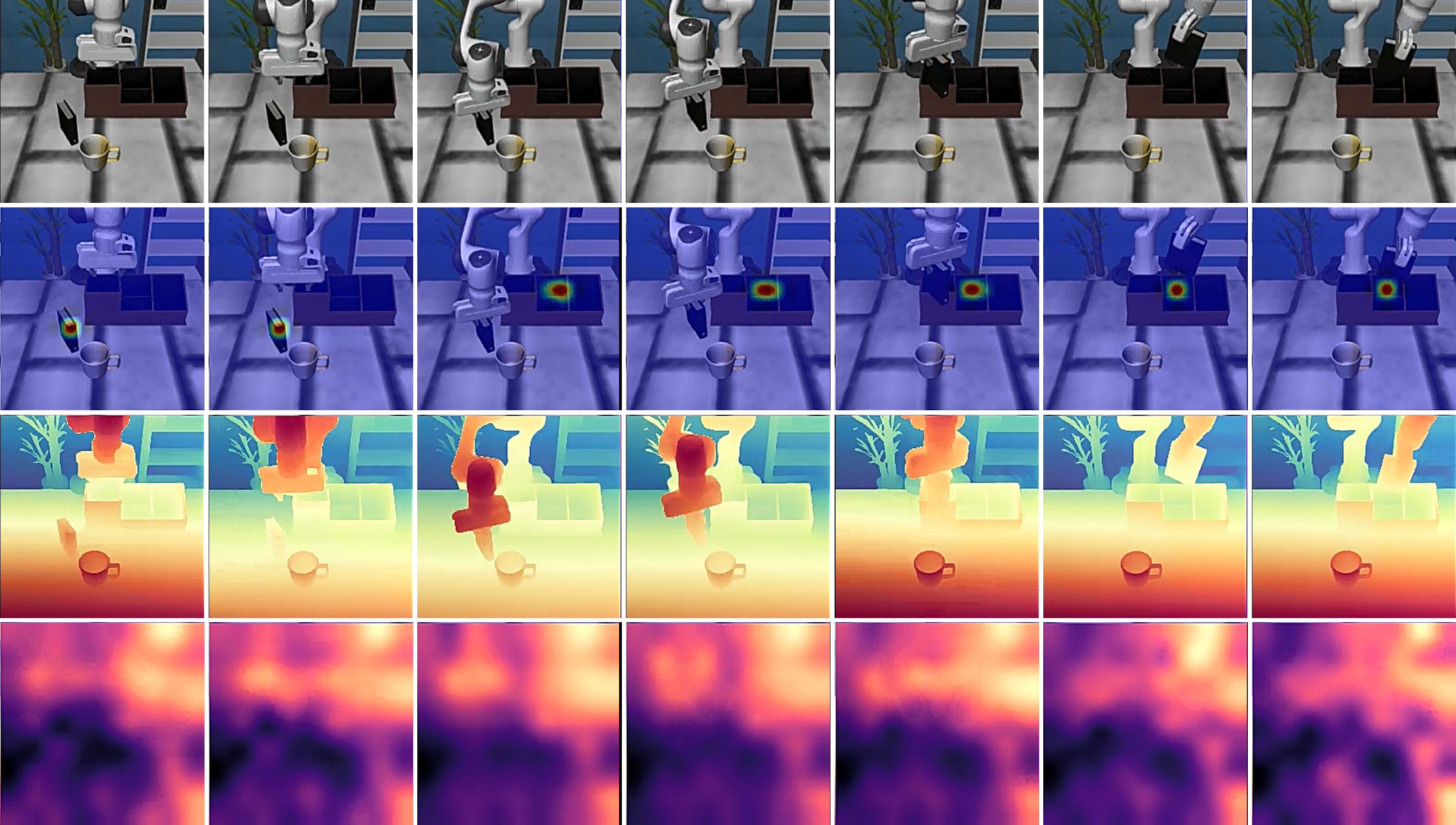}
    \caption{
        \textbf{LIBERO-Plus rollout visualization (long task suite of LIBERO-Plus).}
        Each column corresponds to one stage in the whole episode, with 7 stages in total.
    }
    \label{fig:attention_visualization_long}
\end{figure*}

\begin{figure*}[htbp]
    \centering
    \includegraphics[width=0.9\textwidth]{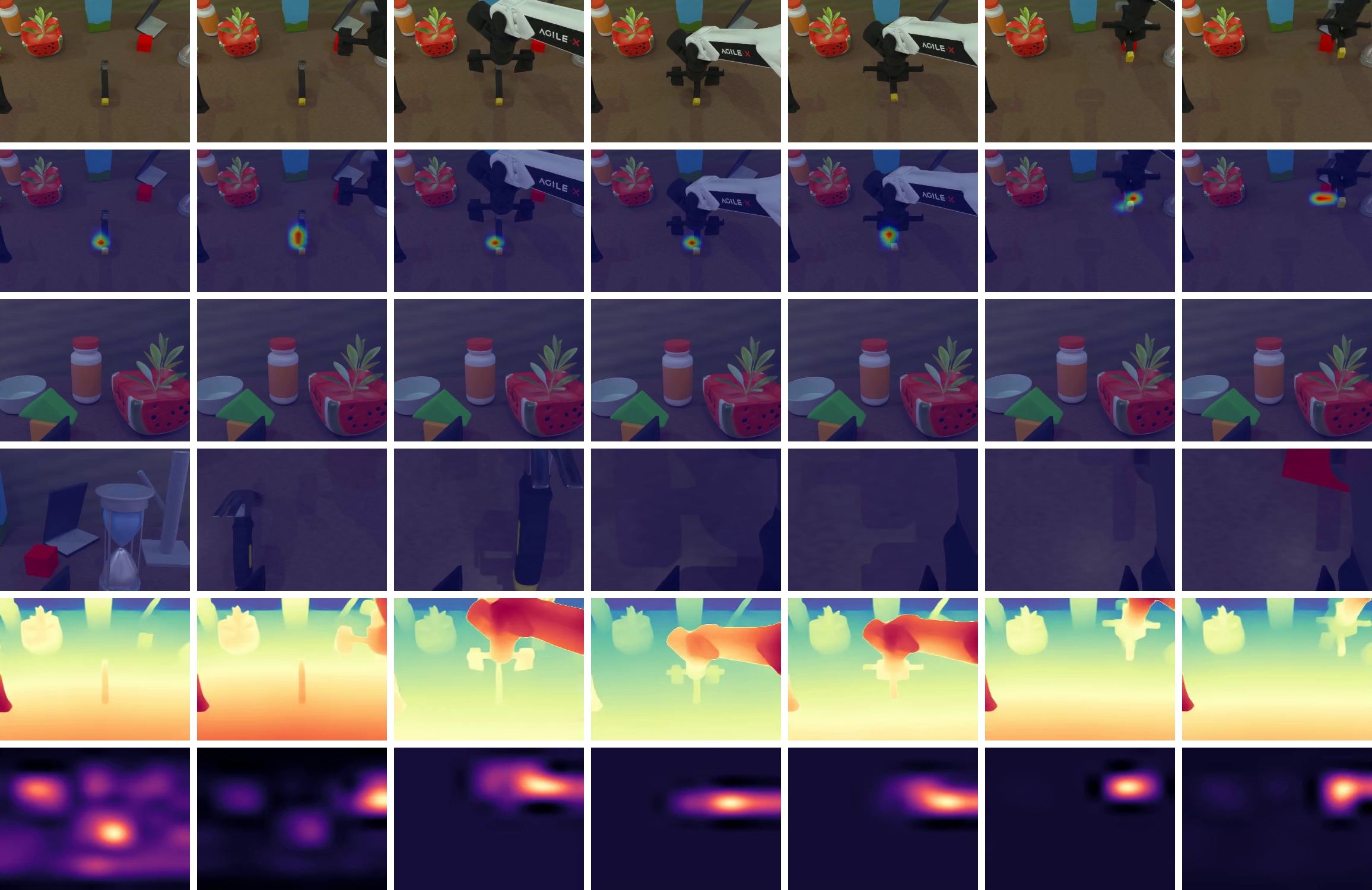}
    \caption{
        \textbf{RoboTwin 2.0 rollout visualization (beat hammer block).}
        Each column corresponds to one stage in the whole episode, with 7 stages in total. The first row shows the original RGB observations during the rollout. The second, third, and fourth rows visualize the attention maps from \alias’s object head for the main camera, left wrist camera, and right wrist camera, respectively. The fifth row presents the depth information encoded by the depth encoder from the main camera view, while the sixth row illustrates the corresponding attention maps generated by \alias’s depth head based on the depth features shown in the fifth row.
    }
    \label{fig:robotwin_vis1}
\end{figure*}

\begin{figure*}[htbp]
    \centering
    \includegraphics[width=0.9\textwidth]{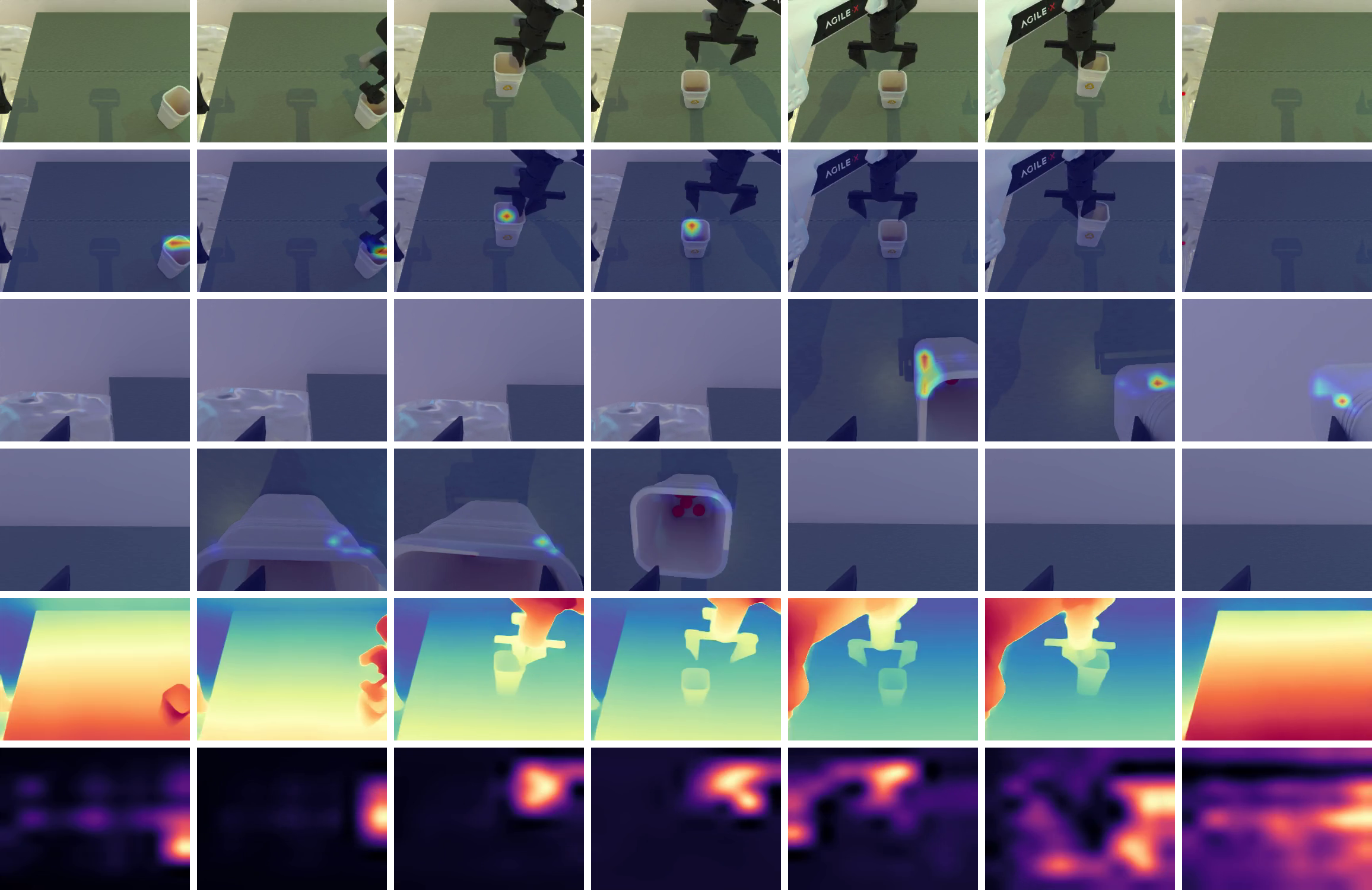}
    \caption{
        \textbf{RoboTwin 2.0 rollout visualization (dump bin bigbin).}
        Each column corresponds to one stage in the whole episode, with 7 stages in total.
    }
    \label{fig:robotwin_vis2}
\end{figure*}

\begin{figure*}[htbp]
    \centering
    \includegraphics[width=0.9\textwidth]{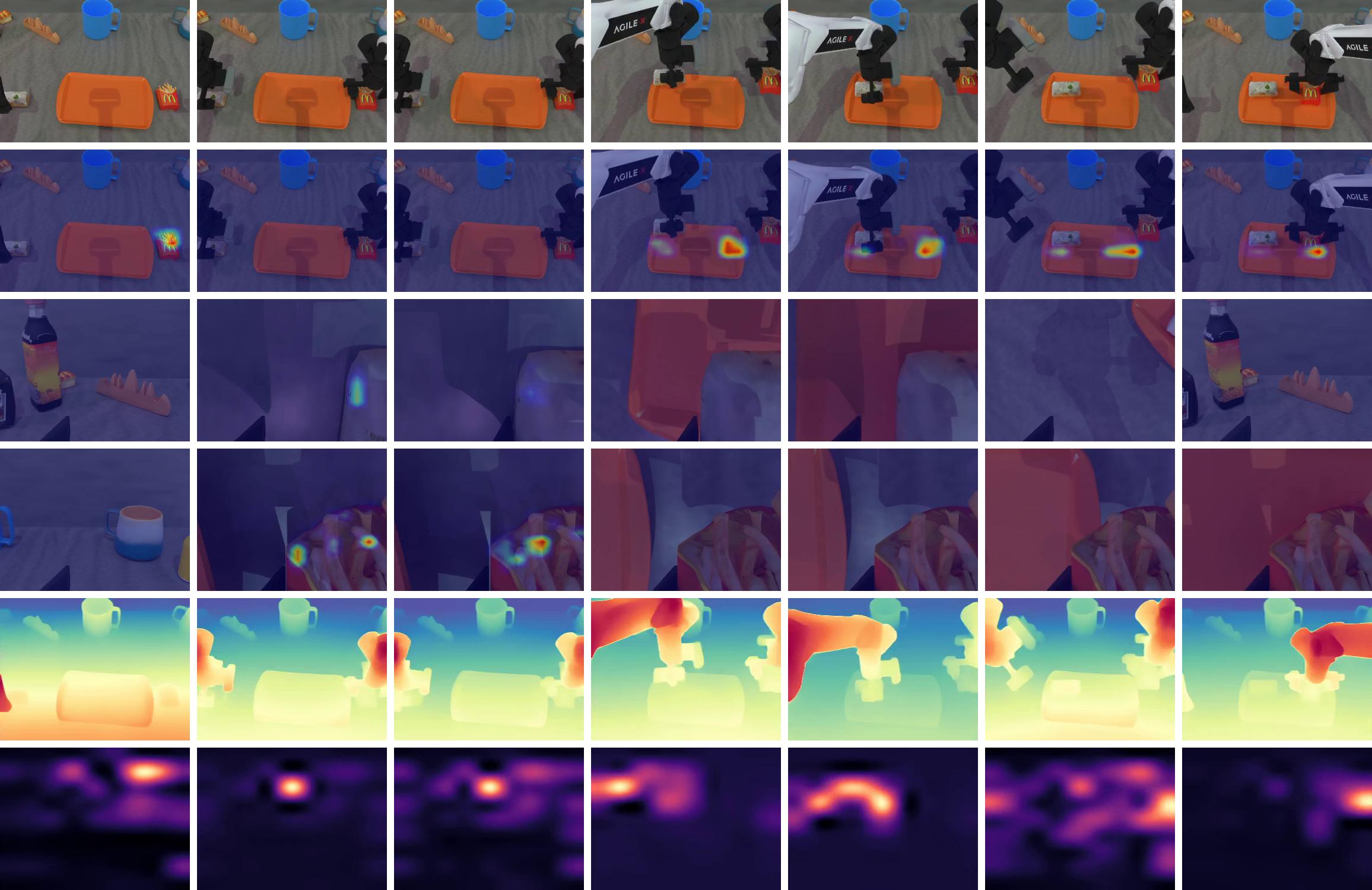}
    \caption{
        \textbf{RoboTwin 2.0 rollout visualization (place burger fries).}
        Each column corresponds to one stage in the whole episode, with 7 stages in total.
    }
    \label{fig:robotwin_vis3}
\end{figure*}

\begin{figure*}[htbp]
    \centering
    \includegraphics[width=0.9\textwidth]{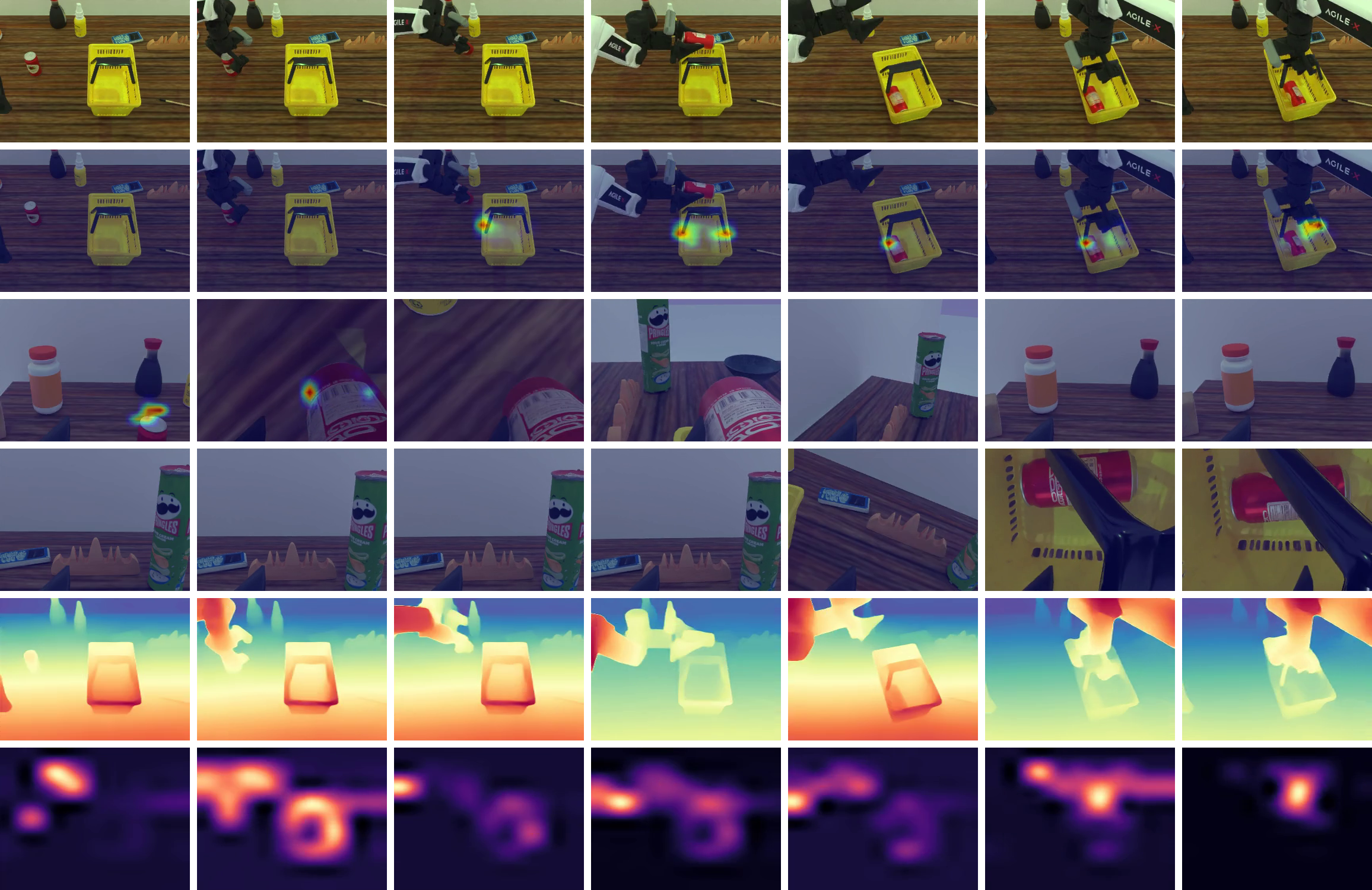}
    \caption{
        \textbf{RoboTwin 2.0 rollout visualization (place can basket).}
        Each column corresponds to one stage in the whole episode, with 7 stages in total.
    }
    \label{fig:robotwin_vis4}
\end{figure*}

\subsection{Visualization of Real-World Result}
\label{sec:real_world_visualization}

\subsubsection{Visualization of Real-World Tasks Rollouts}
\label{sec:real_vis}

To complement the quantitative success rates in Table~\mainref{tab:combined_real_world}, we visualize representative \emph{successful} real-robot rollouts under the three distribution shifts defined in Sec.~\ref{sec:real_generalization}, following the evaluation protocol and success criteria in Sec.~\ref{sec:real_tasks}.
For each task, we show a 3-row keyframe grid with \textbf{7} manually selected stages (approach, grasp, transport, placement, and completion).
Rows correspond to \textbf{in-domain (positional)}, \textbf{lighting}, and \textbf{scene} shifts (top to bottom).

\begin{figure*}[t]
  \centering
  \includegraphics[width=0.98\linewidth]{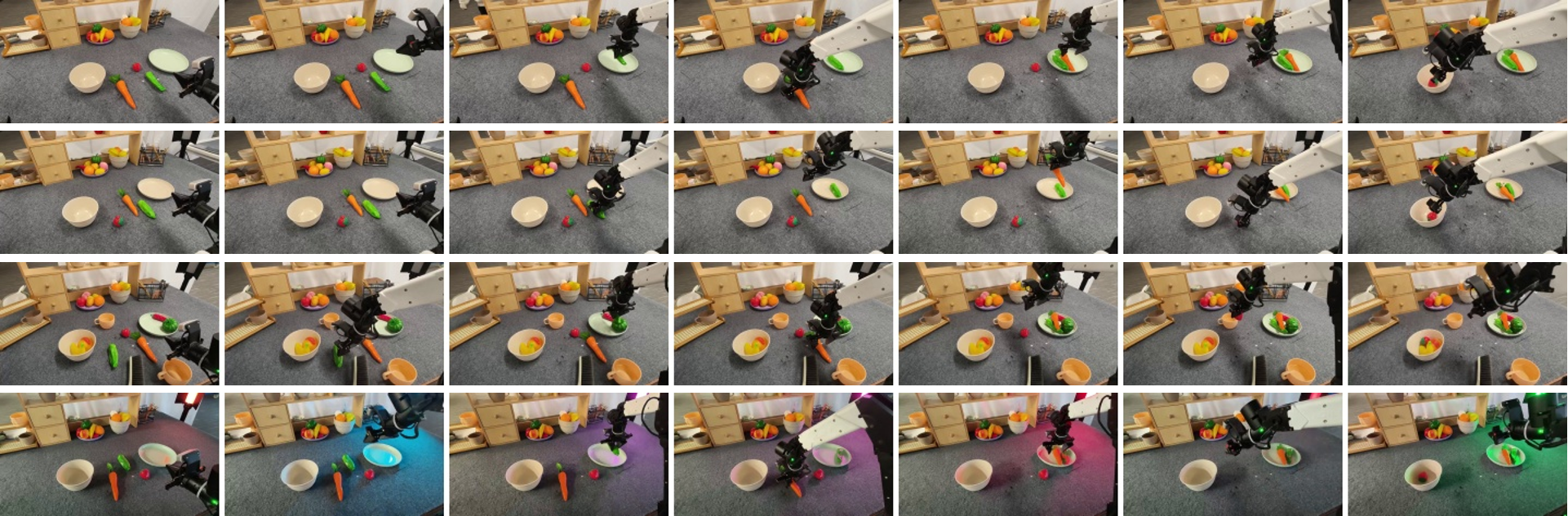}
\caption{\textbf{Real-robot rollout visualization (ALOHA, T1) under distribution shifts.}
Rows: in-domain (positional) / lighting / scene (top to bottom). Columns show 7 key stages of a representative successful trajectory.}
  \label{fig:rollout_t1_aloha}
\end{figure*}

\begin{figure*}[t]
  \centering
  \includegraphics[width=0.98\linewidth]{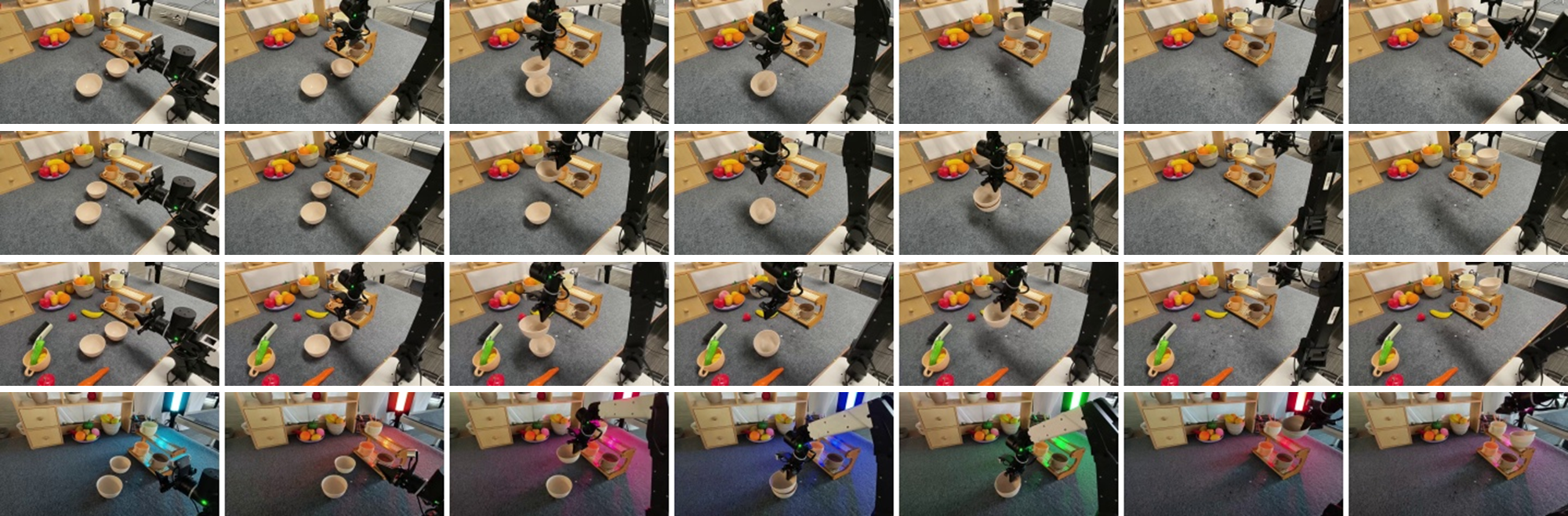}
 \caption{\textbf{Real-robot rollout visualization (ALOHA, T2) under distribution shifts.}Rows: in-domain (positional) / lighting / scene (top to bottom). Columns show 7 key stages of a representative successful trajectory.}
  \label{fig:rollout_t2_aloha}
\end{figure*}

\begin{figure*}[t]
  \centering
  \includegraphics[width=0.98\linewidth]{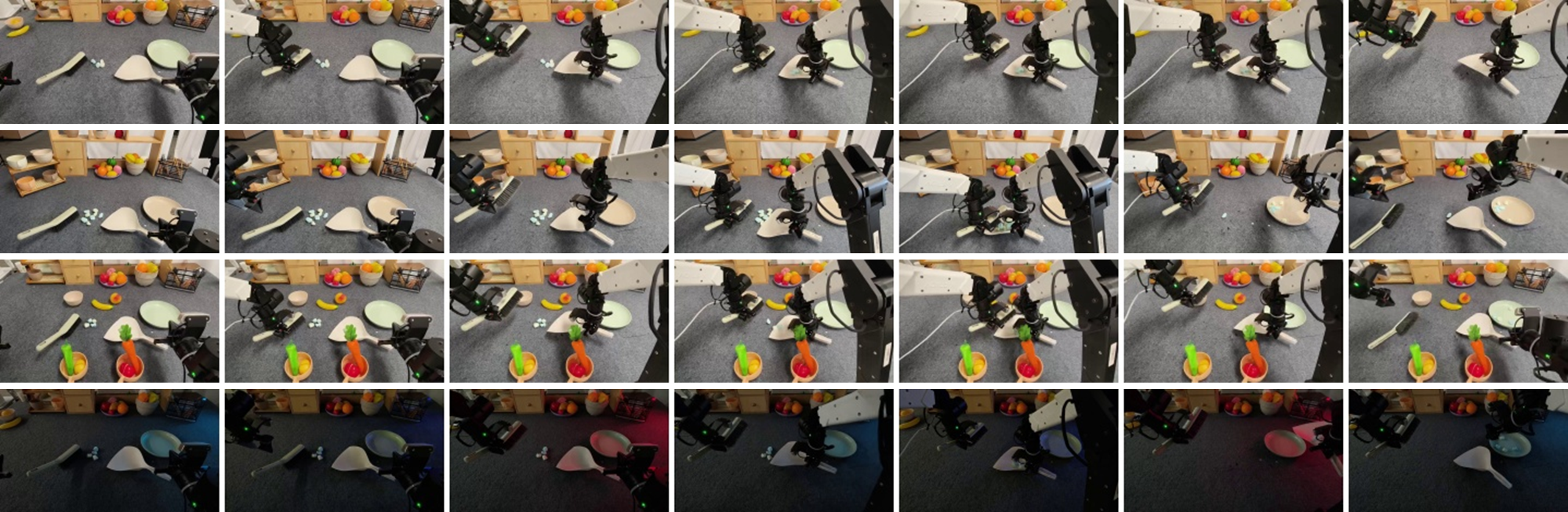}
\caption{\textbf{Real-robot rollout visualization (ALOHA, T3) under distribution shifts.}
Rows: in-domain (positional) / lighting / scene (top to bottom). Columns show 7 key stages of a representative successful trajectory.}
\label{fig:rollout_t3_aloha}
\end{figure*}

\begin{figure*}[t]
  \centering
  \includegraphics[width=0.98\linewidth]{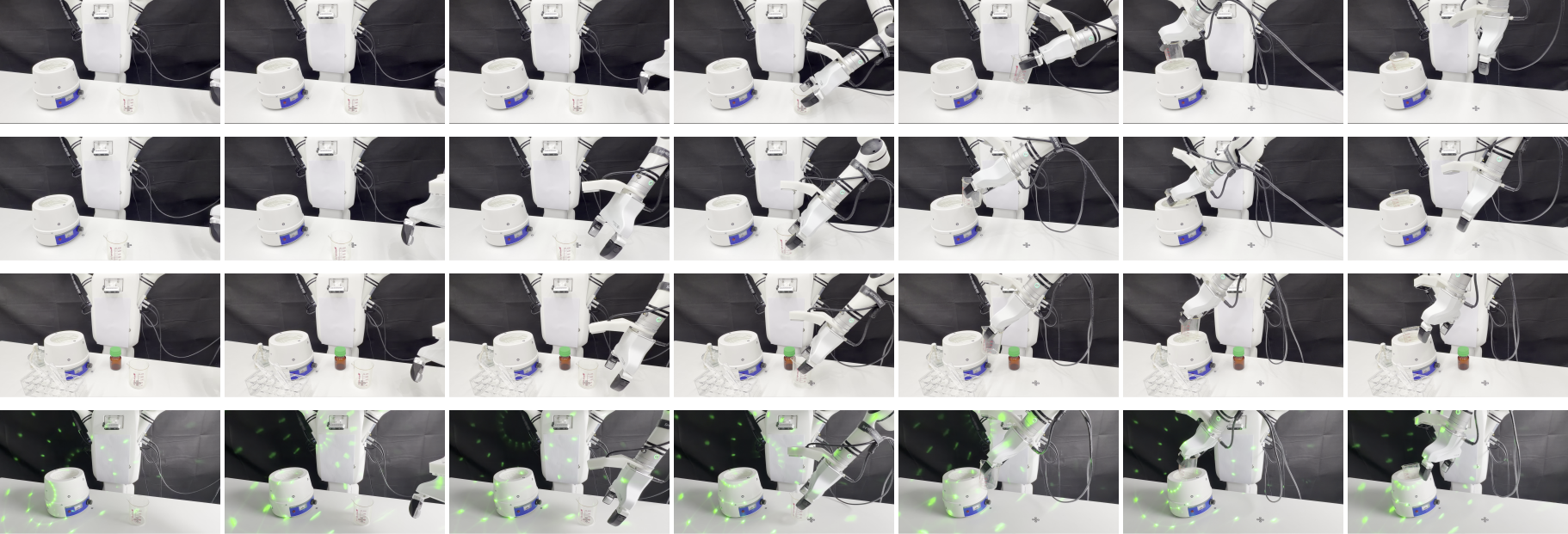}
 \caption{\textbf{Real-robot rollout visualization (PSI-Bot, T4) under distribution shifts.}
Rows: in-domain (positional) / lighting / scene (top to bottom). Columns show 7 key stages of a representative successful trajectory.}
  \label{fig:rollout_t4_psibot}
\end{figure*}

\begin{figure*}[t]
  \centering
  \includegraphics[width=0.98\linewidth]{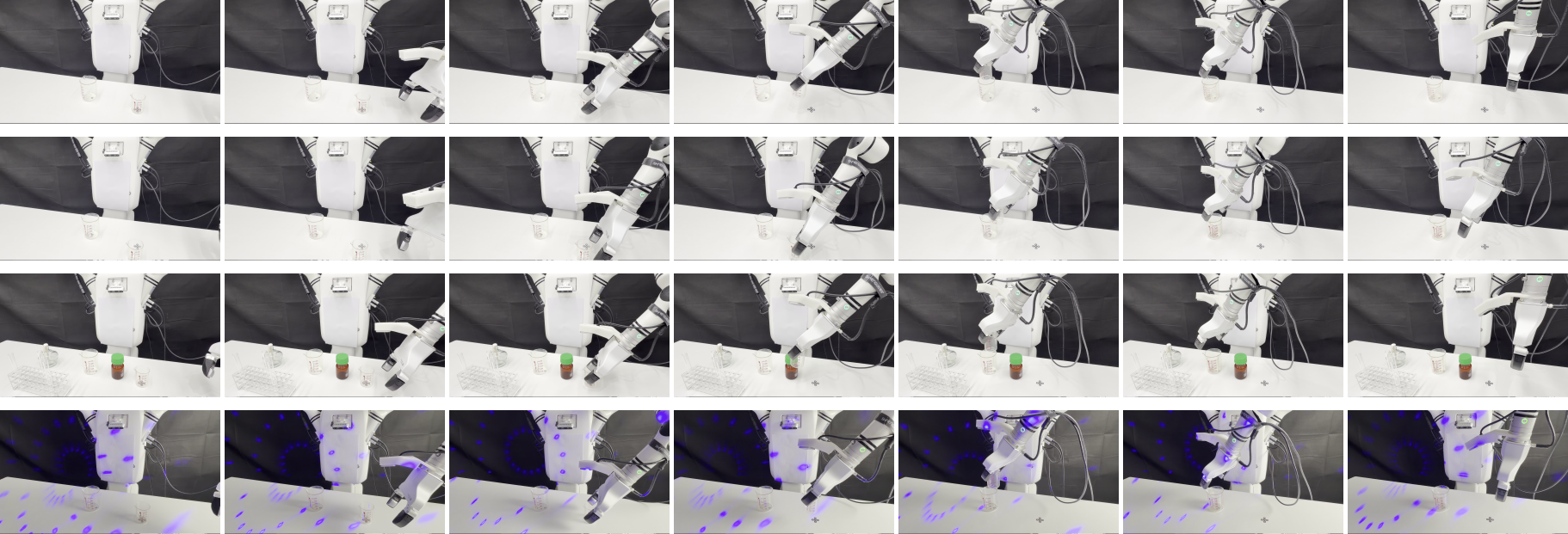}
\caption{\textbf{Real-robot rollout visualization (PSI-Bot, T5) under distribution shifts.}
Rows: in-domain (positional) / lighting / scene (top to bottom). Columns show 7 key stages of a representative successful trajectory.}
  \label{fig:rollout_t5_psibot}
\end{figure*}

\begin{figure*}[t]
  \centering
  \includegraphics[width=0.98\linewidth]{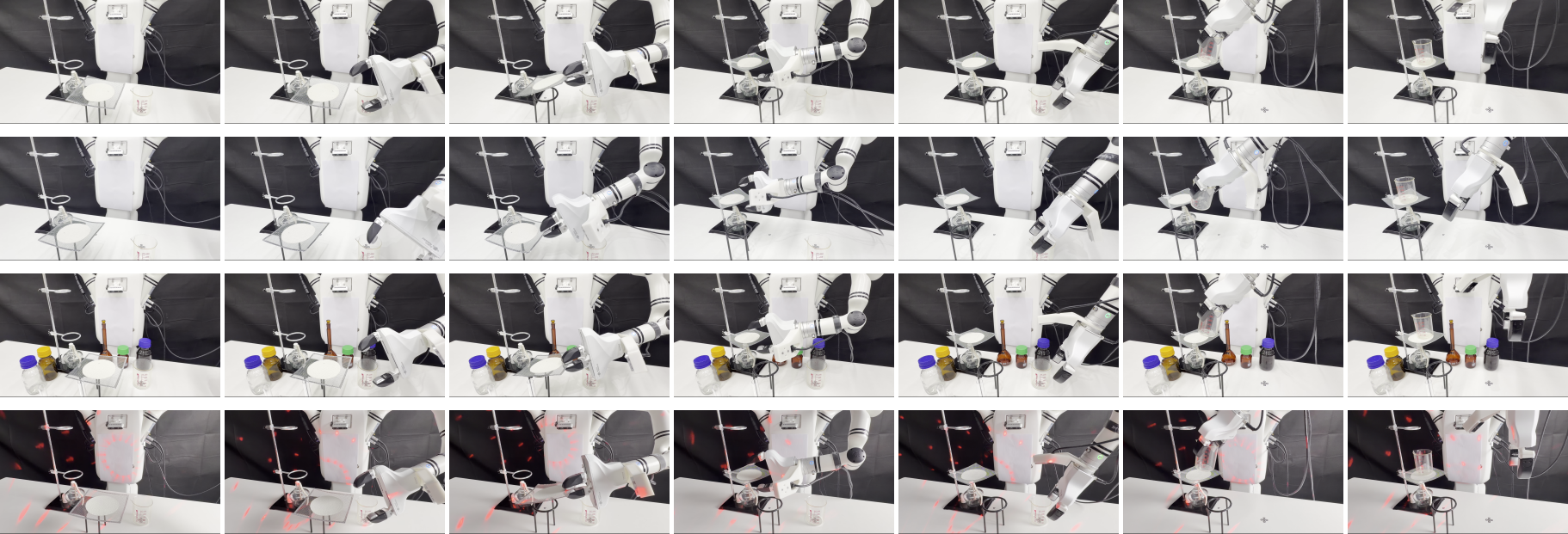}
\caption{\textbf{Real-robot rollout visualization (PSI-Bot, T6) under distribution shifts.}
Rows: in-domain (positional) / lighting / scene (top to bottom). Columns show 7 key stages of a representative successful trajectory.}

  \label{fig:rollout_t6_psibot}
\end{figure*}

\subsubsection{Head-wise Mechanism Visualization on Real-World Robots}
\label{sec:real_head_diagnostics}
To further substantiate that our decoupled supervision indeed induces factor-specific behaviors, we provide head-wise diagnostics on real robots from both quantitative and qualitative perspectives.
We align each specialized head with the tasks where its factor is most critical:
\textbf{Object} head $\rightarrow$ \textbf{T1} (pick up fruits and vegetables) and \textbf{T4} (place beaker in heating mantle),
\textbf{Depth/Geometry} head $\rightarrow$ \textbf{T2} (stack bowls) and \textbf{T5} (stack beakers),
and \textbf{Skill/Temporal} head $\rightarrow$ \textbf{T3} (clean the tabletop) and \textbf{T6} (Heat the beaker).
This alignment allows us to isolate each head's contribution without conflating unrelated failure modes.

\paragraph{Quantitative attribution}
Table~\ref{tab:real_headwise} reports success rates for the base $\pi_0$ policy, three single-head variants, and the full GuidedVLA under the same three distribution shifts (positional/in-domain, scene, and lighting).
Single-head variants are evaluated \emph{only} on their aligned tasks (other entries are marked as ``--'') and serve as diagnostic ablations rather than standalone general-purpose policies.

\paragraph{Qualitative mechanism evidence}
To complement the head-wise success rates in Table~\ref{tab:real_headwise}, we also provide qualitative evidence that each specialized head exhibits the intended factor-specific behavior on its aligned tasks.

\begin{figure*}[t]
  \centering
  \includegraphics[width=0.98\linewidth]{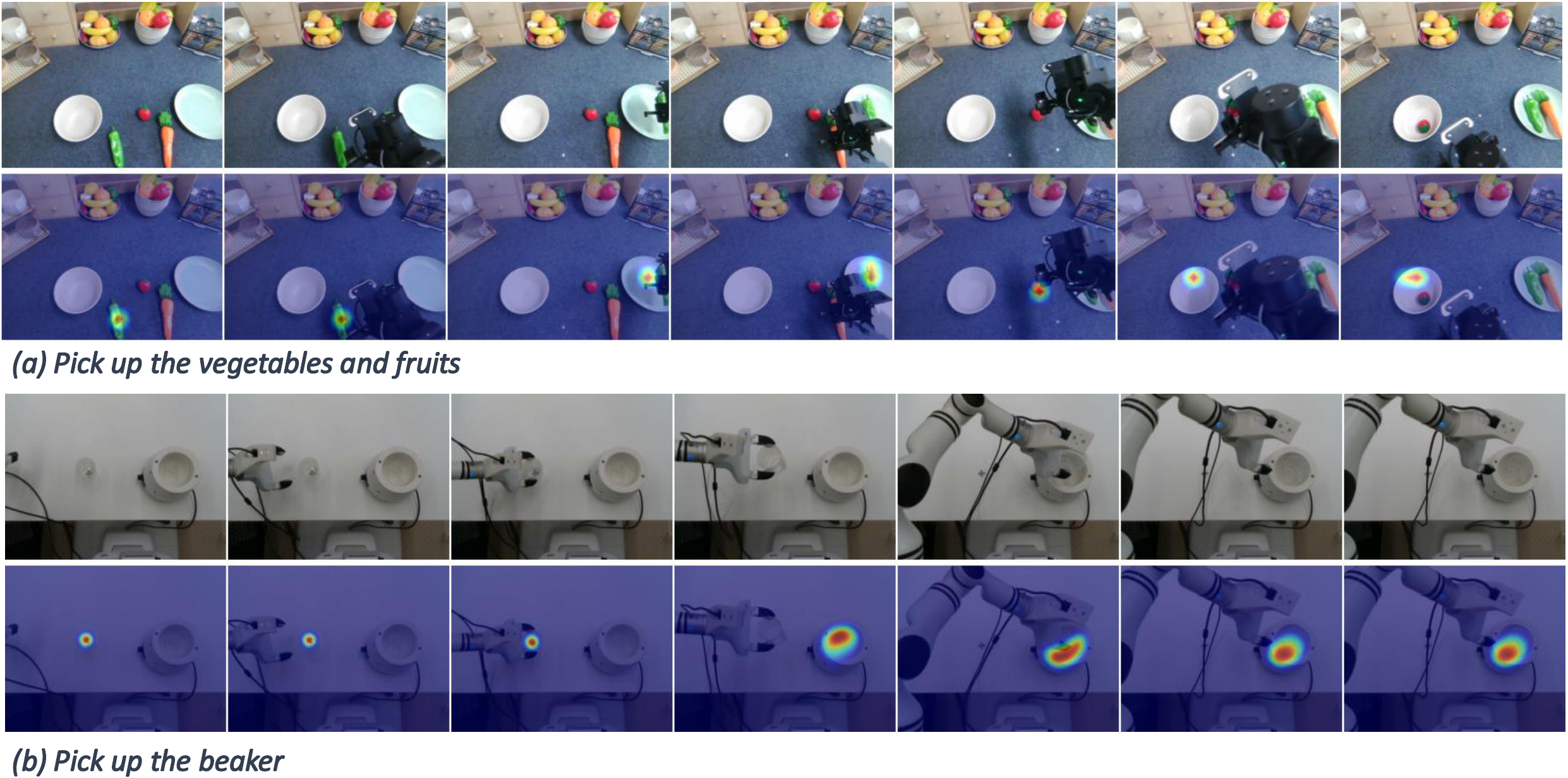}
 \caption{\textbf{Object-head attention on real robots (aligned tasks: T1/T4).}
For each task, columns show 7 matched key stages of a representative successful rollout (left to right).
Top: raw RGB observations. Bottom: normalized attention heatmaps from the \textbf{object-specialized head} overlaid on RGB (warmer colors indicate higher attention).}
\label{fig:real_obj_attn}
  \vspace{-6pt}
\end{figure*}

\begin{figure*}[t]
  \centering
  \includegraphics[width=0.98\linewidth]{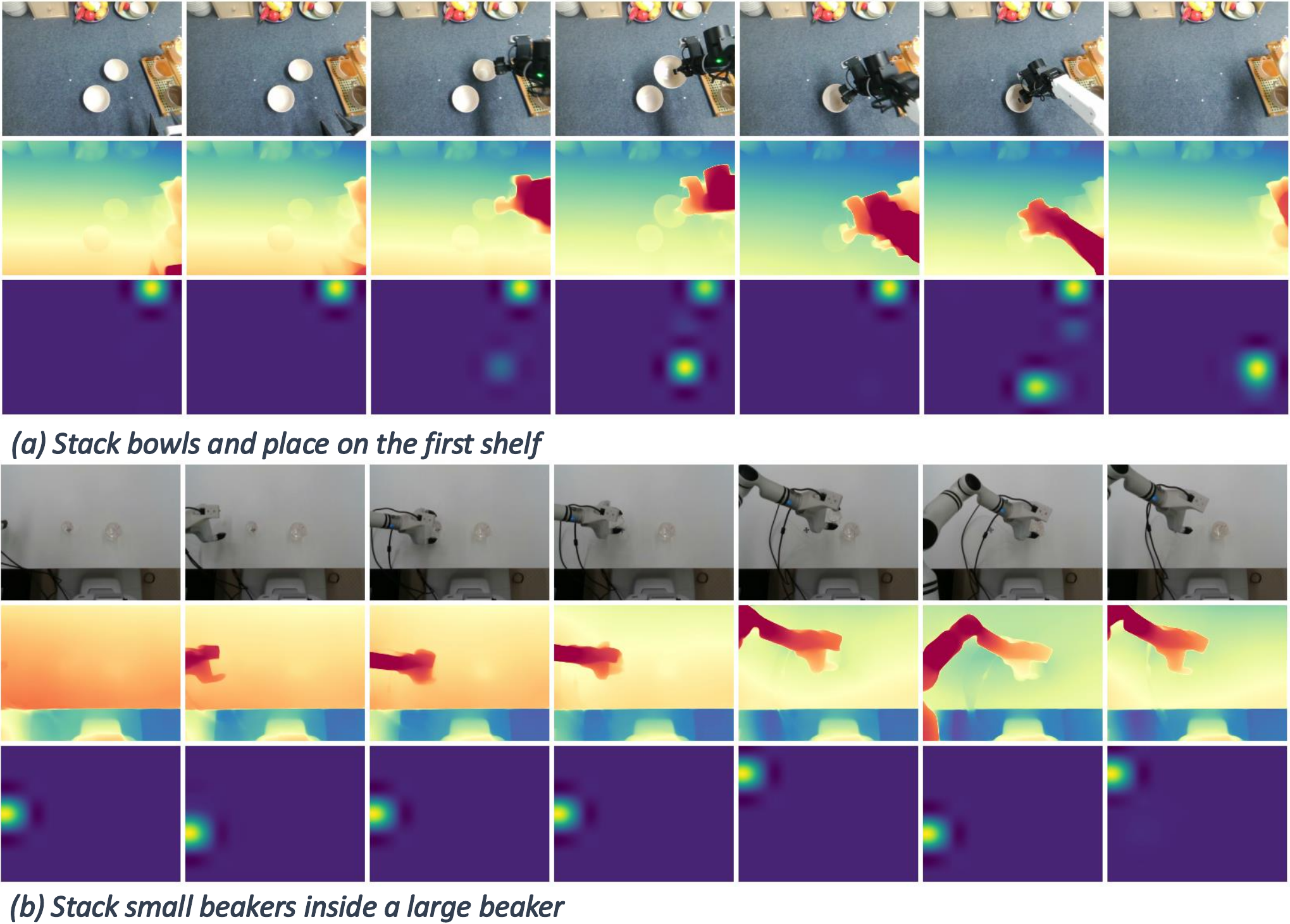}
  \caption{\textbf{Depth/geometry-head diagnostics on real robots (aligned tasks: T2/T5).}
Columns show 7 matched key stages of a representative successful rollout (left to right).
Top: RGB observations. Middle: depth predictions (Depth Anything V3, small variant). Bottom: normalized attention heatmaps from the \textbf{depth/geometry-specialized head} (warmer colors indicate higher attention).}
\label{fig:real_depth_geom}

  \vspace{-6pt}
\end{figure*}

\begin{figure*}[t]
  \centering
\includegraphics[width=0.98\linewidth]{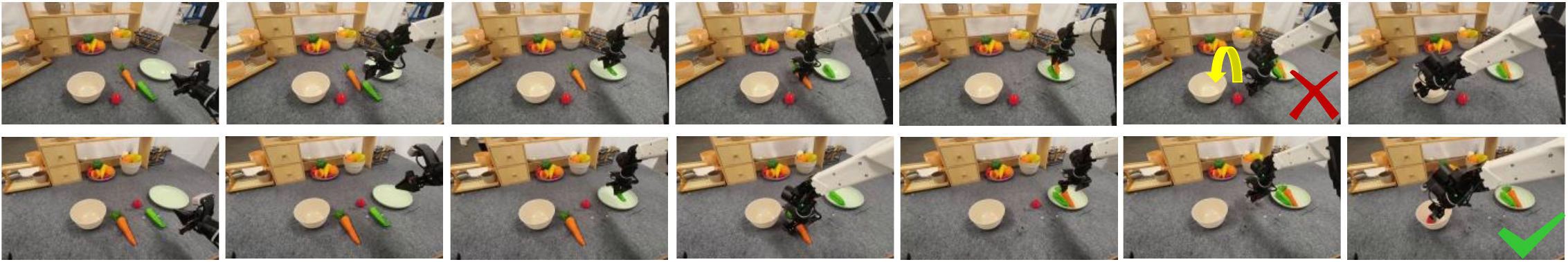}
  \caption{\textbf{Skill/temporal diagnostics on a multi-stage real-robot task.}
Columns show key stages of the tabletop-cleaning sequence.
Top: $\pi_0$ exhibits incorrect temporal progression (e.g., premature termination or missing required sub-steps; marked with red x).
Bottom: GuidedVLA completes the required sub-task order, consistent with skill/temporal supervision.}
\label{fig:real_skill_rollout}
  \vspace{-6pt}
\end{figure*}

Specifically, Fig.~\ref{fig:real_obj_attn} visualizes \textbf{object grounding} by overlaying the attention from the object-specialized head on RGB frames at matched key stages.
Fig.~\ref{fig:real_depth_geom} visualizes \textbf{depth/geometry reasoning} by showing depth cues together with attention overlays from the depth/geometry-specialized head.
Fig.~\ref{fig:real_skill_rollout} visualizes \textbf{skill progression} on a multi-stage task, where $\pi_0$ may skip required sub-steps while GuidedVLA completes the intended sequence.
All examples follow Sec.~\ref{sec:real_tasks} and Sec.~\ref{sec:real_generalization}.


\subsection{Failure Case Analysis (Tasks 1--6)}
\label{sec:failure_cases_all}

We analyze representative failure modes of the baseline $\pi_0$ on two real-robot platforms: 
(i) \textbf{ALOHA AgileX} for household tasks (T1--T3) and 
(ii) \textbf{PSI-Bot} (\emph{Realman RM63 + DexHand2 Pro}, dual Intel D435) for chemical-lab tasks (T4--T6).
Across both domains, failures consistently cluster into three manipulation-critical factors---\textbf{object grounding}, \textbf{metric geometry/clearance}, and \textbf{temporal skill progression}. 
Figs.~\ref{fig:failure_cases_tasks123} and~\ref{fig:failure_cases_chem_tasks456} visualize representative failures (panels (a)--(c) correspond to Tasks 1--3 and 4--6, respectively). Unless noted otherwise, examples are under in-domain conditions with nominal object placement.

\subsubsection{Object grounding failures}
The policy executes \emph{phantom grasps} by approaching empty space near the target, or grasps with an \emph{offset} that causes slippage at lift-off.
This is most evident for small objects in household scenes (Fig.~\ref{fig:failure_cases_tasks123}a) and becomes more severe for transparent glassware in the lab due to refraction/specularities (Fig.~\ref{fig:failure_cases_chem_tasks456}a, top).

 $\pi_0$ relies on incidental appearance cues (contrast/highlights) rather than invariant target identity and precise spatial alignment, making grounding brittle under appearance changes.

\subsubsection{Metric geometry and clearance failures}
The policy fails when millimeter-level depth and clearance are required: 
\emph{half-grasp} on nested bowls due to insufficient insertion depth (Fig.~\ref{fig:failure_cases_tasks123}b), 
rim collisions during heating-mantle insertion (Fig.~\ref{fig:failure_cases_chem_tasks456}a, bottom), 
beaker--beaker collisions during nesting under clutter (Fig.~\ref{fig:failure_cases_chem_tasks456}b, bottom),
and collisions with the ring structure from inaccurate stand geometry localization (Fig.~\ref{fig:failure_cases_chem_tasks456}c, top).
Implicit geometric cues from RGB are insufficient for precise insertion/stacking with tight clearances, especially under clutter and reflective materials.

\subsubsection{Temporal skill collapse in multi-stage execution (T3/T6)}
\begin{figure*}[t]
\centering
\includegraphics[width=0.98\linewidth]{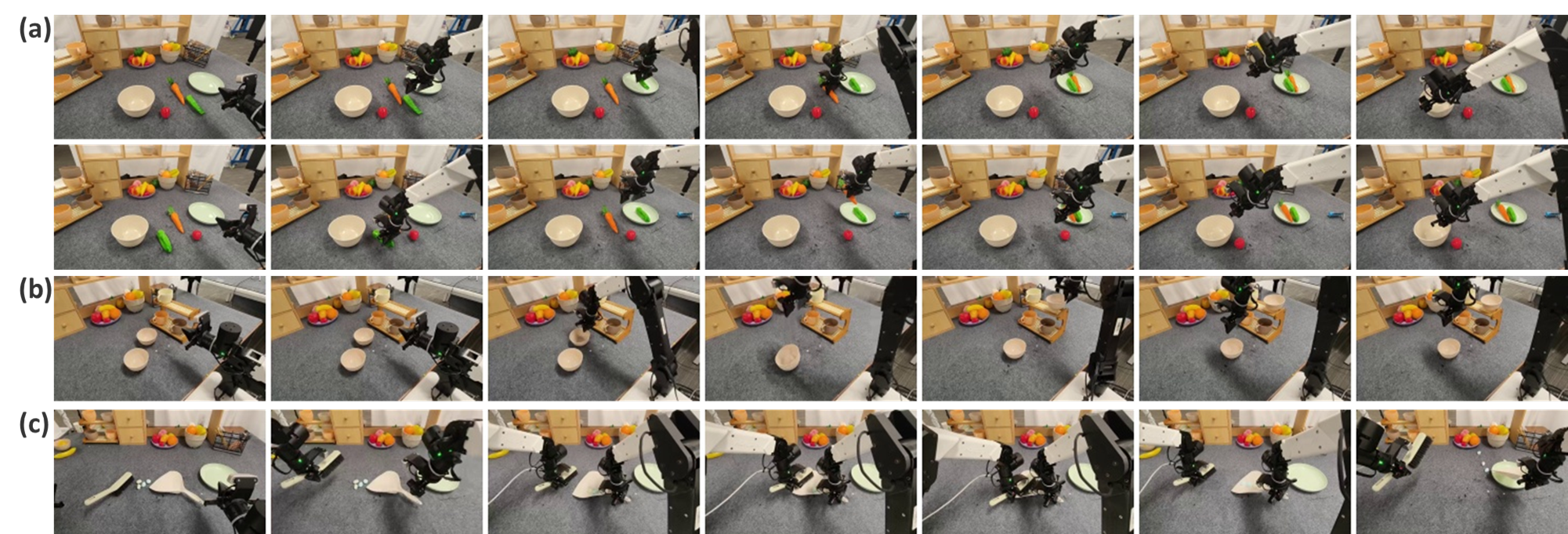}
\caption{\textbf{Representative failure cases of baseline $\pi_0$ on household manipulation tasks (T1--T3, ALOHA).}
(a) T1: \emph{phantom grasp} (top) and \emph{grasp offset/slip} (bottom) when grasping the small strawberry. 
(b) T2: \emph{half-grasp} on nested bowls due to insufficient insertion depth, failing to lift both bowls together. 
(c) T3: \emph{stage-skipping}---pouring succeeds but the required tool-return stage is omitted. 
Examples are under in-domain conditions with nominal object placement.}

\label{fig:failure_cases_tasks123}
\end{figure*}

\begin{figure*}[t]
\centering
\includegraphics[width=0.98\linewidth]{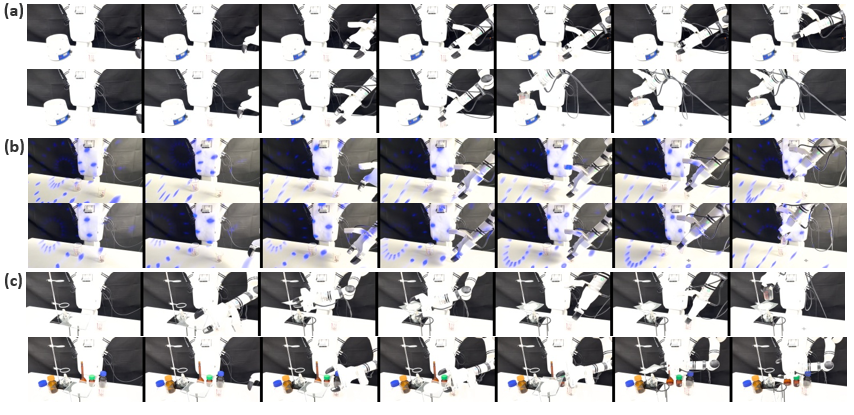}
\caption{\textbf{Representative failure cases of baseline $\pi_0$ on chemical-lab manipulation tasks (T4--T6, PSI-Bot).}
(a) T4: transparent beaker induces \emph{phantom grasp} (top) and \emph{rim collision} during mantle insertion from clearance misestimation (bottom).
(b) T5: \emph{miss-grasp} under lighting/specular highlights (top) and \emph{beaker--beaker collision} during nesting under clutter (bottom).
(c) T6: collision with the ring structure from geometry mislocalization (top) and premature release before stabilization causing gauze roll-off (bottom).
Lab conditions amplify grounding/geometry/temporal weaknesses due to transparent materials and millimeter-level tolerances.}

\label{fig:failure_cases_chem_tasks456}
\end{figure*}

The policy completes a visually salient subgoal but skips required subsequent stages, e.g., pouring succeeds but the tool-return phase is omitted in tabletop cleaning (Fig.~\ref{fig:failure_cases_tasks123}c), and premature release before stabilization causes roll-off in ring-stand assembly (Fig.~\ref{fig:failure_cases_chem_tasks456}c, bottom).
Without explicit supervision for stage awareness, the action decoder can collapse to a short-horizon mode and fail to maintain long-horizon intent.

\subsection{Limitations and Future Work}
Our method requires manual selection of task-relevant factors, which can be domain-dependent. Automating factor discovery, exploring additional factors (e.g., force/torque reasoning), and investigating more general head specialization strategies are promising directions for future research.

\end{document}